%% file: main.tex
\input{preamble}

\begin{document}
\raggedright
\thispagestyle{plain}
\begin{center}
{\Large\bfseries\revtitle{Simulation-Based Evaluation of Residual DDPG-Augmented Computed Torque Control for a Cable-Driven Lower-Limb Rehabilitation Robot Model under Disturbances and Parametric Uncertainties}}\\[7pt]
{Mohammad-Hossein Fakouri\textsuperscript{1} and Ali Keymasi-Khalaji\textsuperscript{1,*}}\\[5pt]
{\small\textsuperscript{1}Department of Mechanical Engineering, Faculty of Engineering, Kharazmi University, Tehran, Iran}\\[3pt]
{\small\textsuperscript{*}Correspondence: \texttt{keymasi@khu.ac.ir}}
\end{center}
\vspace{4pt}
\begin{abstract}
\revtitle{Nominal computed torque control (CTC) relies on accurate inverse dynamics; parameter mismatch and disturbances can leave tracking error, while joint limits and unilateral cable forces constrain motion. This numerical study evaluates a bounded Deep Deterministic Policy Gradient (DDPG) residual correction around nominal CTC for a planar three-degree-of-freedom, three-cable lower-limb rehabilitation robot \emph{model}. In four simulation cases, we examine tracking, joint limits, workspace, cable-Jacobian conditioning and inferred cable demands. In a representative combined uncertainty/disturbance case, Cartesian RMS error was 0.0221419~m with CTC and 0.0128462~m with residual DDPG--CTC (approximately 42\% lower). On a different circular reference, the residual branch nevertheless increased joint-limit-violating channel samples from 44 to 101. Separate supplemental experiments analyze training variability and checkpoints, alternative compensators, reward/action sensitivity, independently varied parameters and disturbances, trajectory transfer, constrained cable-force allocation and force lag. These studies show both improvements and failures, including early terminations and force-capacity-dependent degradation. The verified original algebraic force-backprojection path and later bounded-force simulations are distinguished; numerical allocations do not establish physical pretension or motor performance. Simscape and MSC ADAMS checks are numerical cross-model consistency assessments. No hardware, hardware-in-the-loop or patient tests were performed, and no clinical efficacy or physical safety guarantee is claimed.}
\end{abstract}
\hypersetup{
 pdftitle={Simulation-Based Evaluation of Residual DDPG-Augmented Computed Torque Control for a Cable-Driven Lower-Limb Rehabilitation Robot Model under Disturbances and Parametric Uncertainties},
 pdfauthor={Mohammad-Hossein Fakouri; Ali Keymasi-Khalaji},
 pdfsubject={Revised numerical comparison of residual DDPG and computed torque control}}
\input{sections/01_introduction}
\input{sections/02_robot_kinematics}
\input{sections/03_dynamics_and_allocation}
\input{sections/04_control_design}

\input{sections/05_simulation_method}
\input{sections/05b_revision_methods}
\input{sections/06_results_and_discussion}
\input{sections/06b_revision_results}
\input{sections/06c_revision_limitations}
\input{sections/07_conclusion}
\section*{Author contributions}
Mohammad-Hossein Fakouri: Conceptualization, Methodology, Software, Validation, Formal analysis, Investigation, Original draft and Visualization. Ali Keymasi-Khalaji: Supervision, Conceptualization, Methodology, Writing---review and editing.
\input{sections/08_declarations_competing}
\input{sections/09_declarations_funding}
\input{sections/10_data_code_availability}
\input{sections/11_declaration_ai}
\FloatBarrier
\renewcommand{\bibfont}{\small\linespread{0.93}\selectfont}
\setlength{\bibsep}{-2pt}
\bibliographystyle{unsrtnat}
\bibliography{bibliography/references}
\end{document}

%% file: preamble.tex
\documentclass[11pt,a4paper]{article}
\usepackage[margin=26mm]{geometry}
\usepackage[numbers,sort&compress]{natbib}
\usepackage{amsmath,amssymb,amsfonts,bm,graphicx,booktabs,array,tabularx,multirow,siunitx,subcaption,float,placeins,enumitem,url,hyperref}
\usepackage[most]{tcolorbox}
\usepackage{etoolbox}
\tcbuselibrary{breakable}
\definecolor{revisiongreen}{RGB}{214,249,205}
\graphicspath{{figures/current/}{figures/revision/}}
\newenvironment{revisionblock}{%
  \ifdefined\SHOWREVISIONS
  \begin{tcolorbox}[enhanced,breakable,boxrule=0pt,frame hidden,
    colback=revisiongreen,left=1.3mm,right=1.3mm,top=0.6mm,bottom=0.6mm,
    before skip=2pt,after skip=2pt]
  \fi
}{%
  \ifdefined\SHOWREVISIONS\end{tcolorbox}\fi
}
\newcommand{\revtitle}[1]{\ifdefined\SHOWREVISIONS\textcolor{green!40!black}{#1}\else #1\fi}
\newcommand{\revcaption}[1]{\ifdefined\SHOWREVISIONS\textcolor{green!40!black}{#1}\else #1\fi}
\newcommand{\ZenodoDOI}{10.5281/zenodo.24218424}

\newcommand{\q}{\bm{q}}
\newcommand{\qd}{\dot{\bm{q}}}
\newcommand{\qdd}{\ddot{\bm{q}}}
\newcommand{\x}{\bm{x}}
\newcommand{\xd}{\bm{x}_d}
\newcommand{\e}{\bm{e}}
\newcommand{\ed}{\dot{\bm{e}}}
\newcommand{\F}{\bm{F}}
\newcommand{\Jc}{\bm{J}_c}
\newcommand{\D}{\bm{D}}
\newcommand{\C}{\bm{C}}
\newcommand{\G}{\bm{G}}
\newcommand{\tauctc}{\bm{\tau}_{\mathrm{ctc}}}
\newcommand{\taurl}{\bm{\tau}_{\mathrm{RL}}}
\newcommand{\taucmd}{\bm{\tau}_{\mathrm{cmd}}}
\newcommand{\taud}{\bm{\tau}_{d}}

%% file: sections/01_introduction.tex
\section{Introduction}

Neurological and musculoskeletal impairments can reduce a person's ability to execute repeatable lower-limb movements, motivating robotic devices that deliver programmable and measurable assistance during rehabilitation exercises. Robotic platforms can reproduce prescribed motions, record kinematic and actuation variables, and support systematic controller development under repeatable conditions. These capabilities motivate engineering research on accurate tracking, disturbance rejection, actuation feasibility, and progression toward experimental validation. For lower-limb systems, the commanded movement must remain compatible with joint limits, while patient-dependent mass, segment length, voluntary motion, and attachment conditions can differ from the nominal model \cite{alrahmani2026lower,wang2023bedside,wu2011calt,jin2015calex}.

Cable-driven rehabilitation robots (CDRRs) are attractive in this setting because motors can be located away from the assisted limb while lightweight cables transmit force to the moving structure. Remote actuation can reduce moving inertia and mechanical bulk, improve configurability, and permit relatively large workspaces compared with rigid transmissions of similar moving mass \cite{shoaib2021cdr,qian2018review}. Cable-driven locomotor-training systems, active leg exoskeletons, articulated gait-training mechanisms, bedside lower-limb robots, and movable rehabilitation platforms demonstrate the breadth of this architecture \cite{wu2011calt,jin2015calex,alamdari2016gait,wang2023bedside,zou2019movable,barbosa2018lower}. Related upper-limb and waist devices have also employed impedance, admittance, hybrid, and task-dependent assistance strategies to adapt robot behavior to rehabilitation objectives \cite{chen2019waist,oyman2022upperlower,oyman2018impedance,yang2017admittance,li2021hybrid}.

The mechanical advantages of cable actuation are accompanied by a fundamental unilateral constraint: a cable can pull but cannot push. Consequently, a generalized torque command is meaningful only when it can be produced by nonnegative cable tensions. The feasible set depends on cable routing, anchor and attachment locations, posture, and the rank and conditioning of the cable Jacobian. A point may be reachable by serial-link kinematics yet infeasible for the selected cable arrangement, and a full-rank mapping can still be numerically sensitive when its smallest singular value is small. Positive-tension control, wrench feasibility, redundancy resolution, and force distribution are therefore central in cable-driven robotics \cite{oh2005positive,khosravi2014robust,babaghasabha2015adaptive,jabbari2017robust,jabbari2017adaptiveNN,picard2021payload}. In a rehabilitation mechanism, these cable-specific constraints are coupled with anatomical or prescribed joint limits; controller assessment should therefore include more than Cartesian error alone.

Model-based nonlinear control provides a transparent starting point for such systems. Computed torque control (CTC) uses a nominal dynamic model to compensate inertia, Coriolis/centrifugal, and gravity terms, leaving a designer-selected second-order tracking-error model when the nominal dynamics match the plant. CTC and related inverse-dynamics methods are attractive because their components have clear physical interpretations and because the nominal feedback gains can be selected using conventional control reasoning. Cable-driven rehabilitation and parallel-robot studies have combined model-based control with positive-tension checks, robust terms, adaptive estimation, impedance/admittance behavior, or assist-as-needed objectives \cite{seyfi2021robust,gharatappeh2016assist,rezazadeh2008impedance,li2021hybrid}. When masses, link lengths, inertias, friction, attachments, or external loads differ from the nominal model, exact cancellation no longer holds and an additional compensation mechanism becomes valuable.

Robust, adaptive, disturbance-observer, and sliding-mode controllers can reduce sensitivity to these uncertainties \cite{khosravi2014robust,babaghasabha2015adaptive,jabbari2017robust,mohammed2016ndobsmc,madani2014finite,seyfi2021robust}. Their effectiveness depends on uncertainty bounds, observer assumptions, and tuning choices, and strongly discontinuous or high-gain compensation can be undesirable in human-involved applications. These considerations motivate a complementary approach in which the model-based controller remains responsible for the nominal dynamics while a learned component is assigned only the unresolved mismatch. Such a division preserves the interpretability and baseline competence of CTC and reduces the authority demanded from the learning policy.

Deep reinforcement learning (DRL) provides methods for learning continuous control policies from repeated interaction with a simulated or physical environment. Deterministic policy gradient algorithms optimize a parameterized policy for continuous actions, and Deep Deterministic Policy Gradient (DDPG) combines an actor, a critic, replay memory, target networks, and exploration noise to implement this idea with neural networks \cite{silver2014dpg,lillicrap2015ddpg}. Learning-based tracking and compensation have been studied for robotic manipulators, uncertain nonlinear plants, gait-rehabilitation interaction, variable admittance control, and cable-driven parallel robots \cite{hu2020rltracking,cao2021fixedtime,guo2019gaitrl,dimeas2015admittance,shah2009uncertain,chu2020ddpg,xiong2019hybrid,joyo2025drl}. These studies show the flexibility of learned continuous-action policies, but a fully model-free low-level controller can require substantial training data, obscure the role of known dynamics, and complicate the interpretation of commanded actions and constraint violations.

Residual reinforcement learning addresses this issue by learning a correction around a conventional controller rather than replacing it \cite{johannink2019residual}. In this architecture, the nominal controller supplies the main command and the learned policy compensates for effects that remain after nominal control. For the present robot, the residual action is bounded joint by joint. This structure limits policy authority and makes residual-action utilization directly auditable; whether the learned correction remains subordinate to the nominal control layer is assessed from the torque logs. Constraint monitoring and empirical robustness tests then provide traceable evidence about the evaluated operating envelope \cite{brunke2021safe}.

Several directly overlapping studies sharpen the distinction required here. Sancak et al.~\cite{sancak2022planar} investigated reinforcement-learning control for point-to-point and dynamic-reference tracking of a planar cable-driven parallel robot; therefore, using DDPG on a cable robot is not novel by itself. Seyfi and Keymasi-Khalaji~\cite{seyfi2021robust} analyzed a robust cable-driven rehabilitation controller with computed-torque-related compensation, tension and joint-limit considerations. Luo et al.~\cite{luo2026lyapunov} combined impedance control and Lyapunov-constrained TD3 compensation for lower-limb cable rehabilitation and reported physical experiments. The evaluated architecture differs in retaining CTC as its nominal layer, restricting an offline-trained DDPG policy to bounded joint-torque corrections, and reporting separate tracking and cable-demand diagnostics for a planar three-cable mechanism. The comparison separates Cartesian tracking behavior from joint-limit and algebraic cable-demand diagnostics.

\begin{revisionblock}
Patient-specific movement estimation and prescribed-trajectory tracking are related but distinct problems. Xu and colleagues extracted lower-limb muscle-synergy patterns from surface electromyography and estimated lower-limb joint movement with an adaptive neuro-fuzzy inference system \cite{xu2023musclesynergy}. Such work motivates future patient-dependent reference generation, whereas the present simulation neither measures electromyography nor infers voluntary movement intention.

Recent model--data integration studies in hydraulic actuation illustrate alternative balances among learned control, physical models and disturbance observers \cite{yao2023modelbased,yao2024datadriven,yao2025modeldata,yao2026eso}. The hydraulic plants, actuator nonlinearities, experimental settings and stability claims are not directly interchangeable with the three-cable lower-limb model. Likewise, residual learning, DDPG and cable-driven reinforcement learning have existing precedents \cite{johannink2019residual,xiong2019hybrid,sancak2022planar}. The present contribution is therefore an integrated, limitations-aware \emph{numerical comparison}, not a new DRL algorithm, a formal constraint guarantee, or a validated physical actuator architecture.

Perturbation-based approaches including smoothed-functional and simultaneous perturbation stochastic approximation (SPSA) can optimize controller or policy parameters. They differ from the online feedback map implemented by an actor: a norm limit on an optimization update is not a torque-output bound. DDPG was selected because its deterministic continuous-action actor supplies observation-dependent, componentwise bounded corrective torques around CTC; this choice does not establish superior convergence or sample efficiency over alternative policy optimizers. A properly matched head-to-head comparison with smoothed-functional or SPSA training remains outside the present study.
\end{revisionblock}

The literature provides strong foundations in cable-driven rehabilitation mechanisms, positive-tension control, nonlinear model-based control, and learning-based compensation. The present study brings these elements into one traceable evaluation for a planar three-degree-of-freedom (3-DOF), three-cable lower-limb mechanism: nominal CTC, bounded residual DDPG torque, time-localized simultaneous parameter uncertainty, external disturbances, constrained inverse kinematics, nonnegative-tension checks, joint-limit accounting, sampled workspace feasibility, and cable-Jacobian conditioning. The focus is the joint consideration of learning-assisted tracking, mechanism-specific actuation constraints, and numerical model consistency; neither DDPG nor residual learning is claimed as a new algorithm. 

The study addresses three engineering questions. First, under matched simulation conditions, does a bounded residual DDPG term reduce Cartesian tracking error relative to nominal CTC when model mismatch and disturbance are active? Second, does the resulting trajectory remain compatible with the implemented joint-limit and algebraic cable-demand checks? Third, how sensitive are the observed improvements to disturbance seeds, time intervals, nearby initial conditions, and an alternate trajectory? These questions define the evidential scope of the paper and separate empirical robustness from formal robust-control or safety guarantees.

Accordingly, this paper develops a simulation-based residual DDPG augmentation of CTC for a planar 3-DOF cable-driven lower-limb rehabilitation robot. The learned policy does not replace CTC. It supplies a bounded residual torque vector that is added to the nominal model-based command. The main technical elements investigated are:
\begin{enumerate}[leftmargin=*,itemsep=2pt]
    \item a documented kinematic and dynamic model of a three-link lower-limb mechanism with three pull-only cables, constrained reference generation, explicit cable-length Jacobian mapping, and defined joint and cable-feasibility checks;
    \item integration of nominal CTC with a three-action residual DDPG policy bounded by $\pm[5,3,2]^T$ N m, together with a control-oriented description of observations, reward terms, actor--critic networks, training, and deterministic evaluation;
    \item four-scenario comparisons under nominal conditions, 10\% simultaneous parameter uncertainty, external disturbance, and their combination, evaluated using tracking and constraint indicators; and
    \item trajectory-specific feasibility and robustness diagnostics, including sampled geometric and nonnegative-tension workspaces, cable-Jacobian rank and conditioning, ten disturbance seeds, interval-wise errors, four nearby initial configurations, and an alternate circular-path transfer evaluation with explicit joint-limit accounting.
\end{enumerate}

Table~\ref{tab:positioning} summarizes how the principal literature groups support the integrated model-based, learning-based, and cable-feasibility formulation developed in this study.

\begin{table}[!htbp]
\centering
\caption{Technical positioning of the present study relative to the principal literature groups.}
\label{tab:positioning}
\small
\setlength{\tabcolsep}{4pt}
\renewcommand{\arraystretch}{1.10}
\begin{tabularx}{\textwidth}{@{}>{\raggedright\arraybackslash}p{0.20\textwidth}>{\raggedright\arraybackslash}X>{\raggedright\arraybackslash}X@{}}
\toprule
Literature group & Main emphasis & Role in the present study \\
\midrule
Cable-driven rehabilitation systems & Mechanism design, low moving inertia, task-specific assistance, and human-oriented control & Defines the application domain and motivates reproducible lower-limb motion generation \\
Cable feasibility and CDPR control & Positive-tension feasibility, wrench distribution, force allocation, robustness, and Jacobian properties & Motivates nonnegative-tension feasibility, Jacobian-rank, minimum-singular-value, and conditioning analyses \\
Nonlinear rehabilitation control & Computed-torque, robust/adaptive, impedance, admittance, and assist-as-needed control & Establishes the interpretable model-based control layer and comparative baseline \\
DDPG and residual RL & Continuous-action learning and learned correction around an existing controller & Provides the bounded compensation mechanism used here \\
Present study & Computed-torque control with bounded residual DDPG for a planar 3-DOF, three-cable lower-limb model & Integrates trajectory-tracking evaluation with joint-limit, cable-tension, workspace-feasibility, and Jacobian-conditioning analyses \\
\bottomrule
\end{tabularx}
\end{table}

The remainder of the paper is organized as follows. Section~2 defines the coordinate convention, forward and inverse kinematics, cable geometry, and pull-only actuation constraint. Section~3 presents the dynamics, uncertainty and disturbance model, force allocation, and Jacobian diagnostics. Section~4 describes CTC and the residual DDPG architecture. Section~5 specifies model-consistency validation, simulation settings, scenarios, and metrics. Section~6 reports and interprets the tracking, feasibility, robustness, and diagnostic results. Section~7 concludes the paper and outlines the next stages of constraint-aware and experimental validation.

\FloatBarrier

%% file: sections/02_robot_kinematics.tex
\section{Robot kinematics and cable geometry}

This section defines the coordinate convention, serial-link kinematics, cable routing, and reference generation used consistently by the dynamic model, controller, and post-processing routines. Particular attention is given to the sign convention of the cable Jacobian because cable-length derivatives and generalized cable-force mappings differ by a minus sign when positive tension acts from the moving attachment toward the fixed anchor.

\subsection{Coordinate frame and generalized coordinates}

The sagittal-plane inertial frame has its origin at the proximal joint of the first link. The $x$-axis is vertical and positive upward, the $y$-axis is horizontal, and gravity acts along the negative $x$ direction. Some result figures retain the original plotting convention with $Y$ on the horizontal axis and $X$ on the vertical axis, which is stated explicitly in their captions. Joint rotations are positive counterclockwise, and $q_2$ and $q_3$ are relative joint angles. The generalized coordinate, velocity, and acceleration vectors are
\begin{equation}
    \q=[q_1,q_2,q_3]^T,\qquad \qd=\frac{d\q}{dt},\qquad \qdd=\frac{d^2\q}{dt^2}.
\end{equation}
The joints represent hip-, knee-, and ankle-related rotations in the planar model. The link lengths are $b_1=0.45$ m, $b_2=0.35$ m, and $b_3=0.21$ m. Cumulative link orientations are
\begin{equation}
    \theta_1=q_1,\qquad \theta_2=q_1+q_2,\qquad \theta_3=q_1+q_2+q_3.
\label{eq:cumulative_angles}
\end{equation}

The notation used repeatedly in the manuscript is summarized in Table~\ref{tab:notation}.
\begin{table}[!htbp]
\centering
\caption{Main symbols in the kinematic, dynamic, and control formulation.}
\label{tab:notation}
\begin{tabular}{ll}
\toprule
Symbol & Meaning \\
\midrule
$\q,\qd,\qdd$ & Joint position, velocity, and acceleration vectors \\
$\x,\xd$ & Actual and desired Cartesian end-effector positions \\
$\bm{p}_i,\bm{p}_{ci}$ & Joint and link-center-of-mass positions \\
$\bm{a}_i,\bm{r}_i$ & Fixed anchor and moving attachment of cable $i$ \\
$l_i,\bm{u}_i$ & Cable length and anchor-to-attachment unit vector \\
$\Jc$ & Implemented cable actuation Jacobian, $-\partial\bm{l}/\partial\q$ \\
$\D,\C,\G$ & Inertia, Coriolis/centrifugal, and gravity terms \\
$\F$ & Nonnegative cable-tension vector \\
$\tauctc,\taurl,\taucmd$ & Nominal, residual, and combined commanded joint torques \\
$\e,\ed$ & Joint tracking error and error-rate vectors \\
$\sigma_{\min},\kappa$ & Minimum singular value and condition number of $\Jc$ \\
\bottomrule
\end{tabular}
\end{table}

\subsection{Forward kinematics}

\begin{figure}[!htbp]
    \centering
    \begin{subfigure}[b]{0.43\textwidth}
        \centering
        \includegraphics[width=\textwidth]{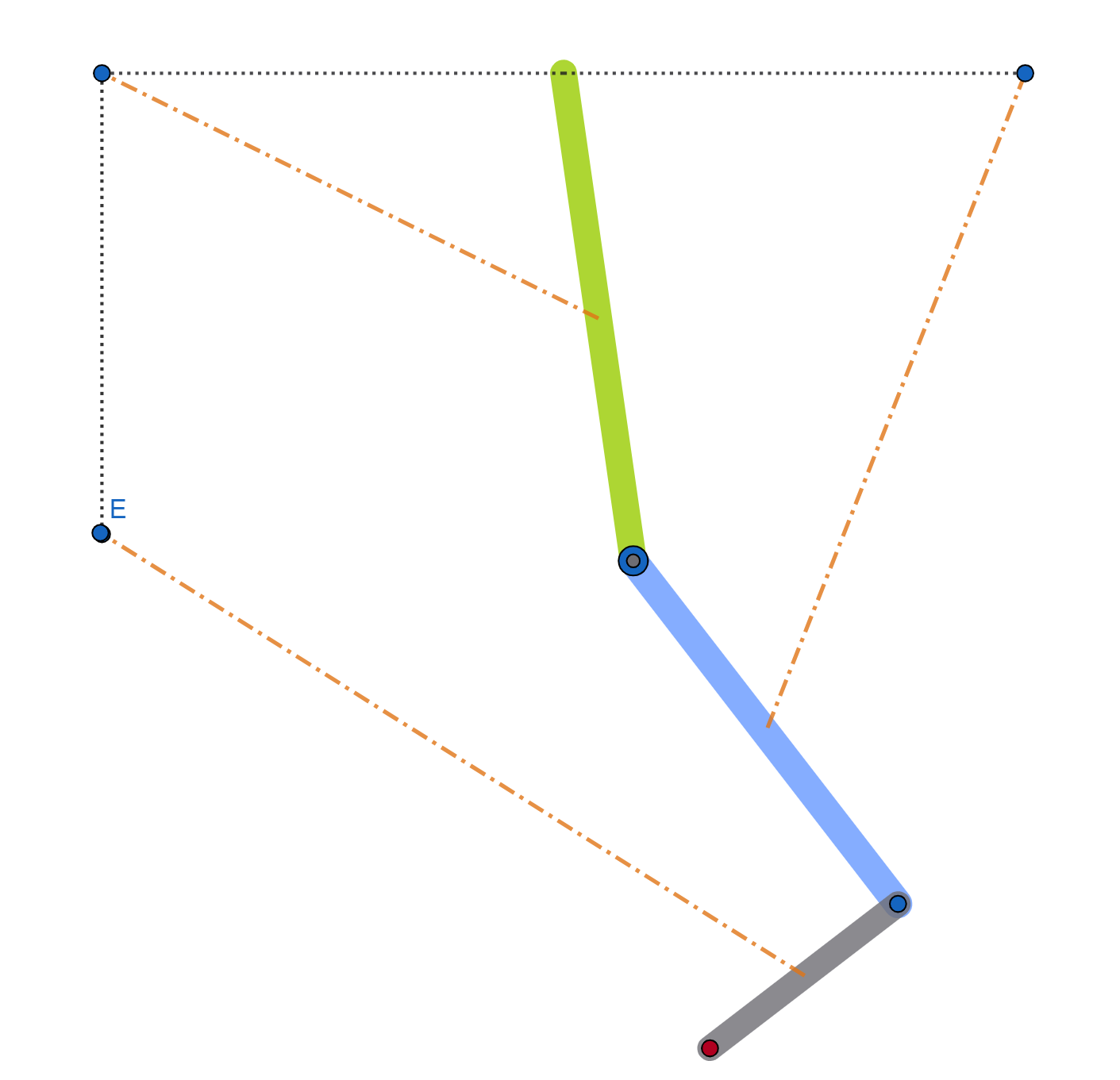}
        \caption{Planar three-link model.}
    \end{subfigure}
    \hfill
    \begin{subfigure}[b]{0.53\textwidth}
        \centering
        \includegraphics[width=\textwidth]{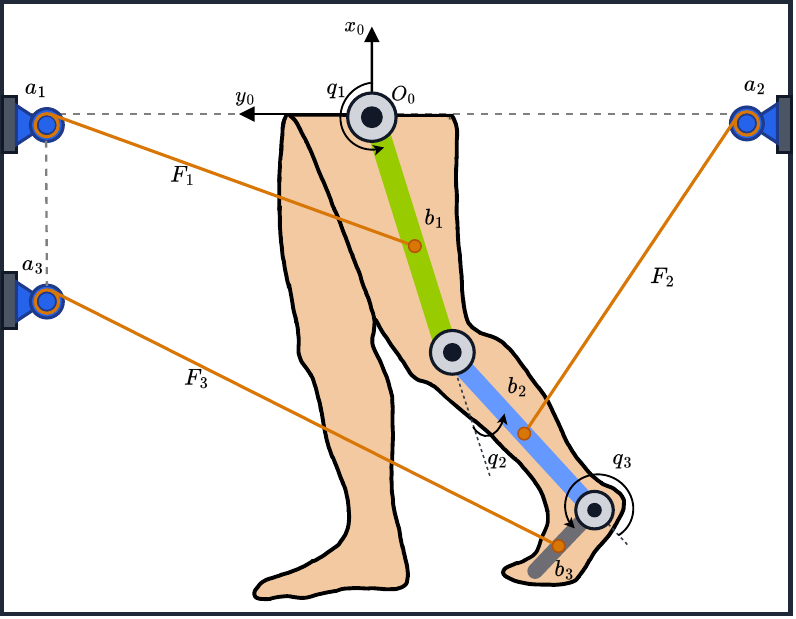}
        \caption{Three-cable routing.}
    \end{subfigure}
    \caption{Planar lower-limb CDRR geometry used in the simulation model. Panel (a) defines the three-link serial mechanism and joint coordinates; panel (b) shows the fixed cable anchors, link-side attachment points, and pull directions that determine the three cable lengths.}
    \label{fig:geometry}
\end{figure}

The successive joint and end-effector positions are
\begin{equation}
\begin{split}
    \bm{p}_0 &= [0,0]^T,\\
    \bm{p}_1 &= b_1[\cos\theta_1,\sin\theta_1]^T,\\
    \bm{p}_2 &= \bm{p}_1+b_2[\cos\theta_2,\sin\theta_2]^T,\\
    \bm{p}_3 &= \bm{p}_2+b_3[\cos\theta_3,\sin\theta_3]^T.
\end{split}
\label{eq:joint_positions}
\end{equation}
Thus the end-effector position $\x=\bm{f}(\q)=[x,y]^T$ is
\begin{equation}
\begin{split}
    x &= b_1\cos q_1+b_2\cos(q_1+q_2)+b_3\cos(q_1+q_2+q_3),\\
    y &= b_1\sin q_1+b_2\sin(q_1+q_2)+b_3\sin(q_1+q_2+q_3).
\end{split}
\label{eq:fk}
\end{equation}
The end-effector Jacobian $\bm{J}_e=\partial\bm{f}/\partial\q\in\mathbb{R}^{2\times3}$ maps joint velocity to Cartesian velocity, $\dot{\x}=\bm{J}_e\qd$. For example,
\begin{equation}
\begin{split}
\frac{\partial x}{\partial q_1}&=-b_1\sin\theta_1-b_2\sin\theta_2-b_3\sin\theta_3,\\
\frac{\partial x}{\partial q_2}&=-b_2\sin\theta_2-b_3\sin\theta_3,\qquad
\frac{\partial x}{\partial q_3}=-b_3\sin\theta_3,
\end{split}
\end{equation}
with analogous cosine terms in the $y$ row. Figure~\ref{fig:geometry} shows the serial-link and cable layouts. The wearable-style illustration is conceptual; all reported results use the planar 3-link/3-cable simulation model.

\subsection{Cable lengths and actuation Jacobian}

The fixed anchor locations are
\begin{equation}
    \bm{a}_1=[0,1.5]^T,\quad
    \bm{a}_2=[0,-1.5]^T,\quad
    \bm{a}_3=[-0.4,1.5]^T\quad \mathrm{m}.
\label{eq:anchors}
\end{equation}
The moving attachment points are located at the centers of the three links:
\begin{equation}
\begin{split}
    \bm{r}_1(\q)&=\frac{b_1}{2}[\cos\theta_1,\sin\theta_1]^T,\\
    \bm{r}_2(\q)&=b_1[\cos\theta_1,\sin\theta_1]^T
    +\frac{b_2}{2}[\cos\theta_2,\sin\theta_2]^T,\\
    \bm{r}_3(\q)&=b_1[\cos\theta_1,\sin\theta_1]^T
    +b_2[\cos\theta_2,\sin\theta_2]^T
    +\frac{b_3}{2}[\cos\theta_3,\sin\theta_3]^T.
\end{split}
\label{eq:attachment}
\end{equation}
For cable $i$, define the anchor-to-attachment vector, length, and corresponding unit vector as
\begin{equation}
    \bm{s}_i(\q)=\bm{r}_i(\q)-\bm{a}_i,\qquad
    l_i(\q)=\|\bm{s}_i(\q)\|_2,\qquad
    \bm{u}_i(\q)=\frac{\bm{s}_i(\q)}{l_i(\q)}.
\label{eq:lengths}
\end{equation}
A positive tension pulls the moving point toward the anchor, so the cable force at the attachment is $-F_i\bm{u}_i$. To make the generalized-force relation take the convenient positive form used in the implementation, the cable actuation Jacobian is defined as
\begin{equation}
    \Jc(\q)=-\frac{\partial\bm{l}(\q)}{\partial\q}\in\mathbb{R}^{3\times3},
\label{eq:Jc_def}
\end{equation}
with elements
\begin{equation}
    J_{c,ij}(\q)=-\bm{u}_i^T(\q)\frac{\partial\bm{r}_i(\q)}{\partial q_j}.
\label{eq:Jc_element}
\end{equation}
Accordingly, cable-length rate and generalized cable torque satisfy
\begin{equation}
    \dot{\bm{l}}=-\Jc(\q)\qd,
\label{eq:ldot}
\end{equation}
\begin{equation}
    \bm{\tau}_c=\Jc^T(\q)\F,
\label{eq:tau_cable}
\end{equation}
where $\F=[F_1,F_2,F_3]^T\in\mathbb{R}^3$. This sign convention matches the MATLAB/Simulink cable-Jacobian and torque-allocation functions used for all reported data. Because cables cannot transmit compression, the basic unilateral constraint is
\begin{equation}
    F_i\geq0,\qquad i=1,2,3.
\label{eq:positive_tension}
\end{equation}
A negative component obtained from Eq.~\eqref{eq:tau_cable} would indicate that the requested generalized torque cannot be produced by the selected cable routing at that posture without changing pretension, routing, or the force-allocation strategy.

\subsection{Reference trajectory and constrained inverse kinematics}

The primary Cartesian reference is a smooth closed rehabilitation-like path over $T=10$ s:
\begin{equation}
\begin{split}
    x_d(t)&=0.015\left(5+\cos\frac{18t}{T}\right)
    \cos\left(\frac{6t}{T}-\frac{\pi}{6}\right)-0.9,\\
    y_d(t)&=0.015\left(5+\cos\frac{18t}{T}\right)
    \sin\left(\frac{6t}{T}-\frac{\pi}{6}\right).
\end{split}
\label{eq:main_path}
\end{equation}

At each sample $t_k$, the desired joint coordinates are obtained by solving the bounded inverse-kinematics problem
\begin{equation}
\begin{split}
    \q_d(t_k)=\arg\min_{\q}\;&
    \left\|\bm{f}(\q)-\xd(t_k)\right\|_2^2,\\
    \mathrm{s.t.}\;&
    \q_{\min}\leq\q\leq\q_{\max}.
\end{split}
\label{eq:ik}
\end{equation}

The active embedded Simulink implementation solves
Eq.~\eqref{eq:ik} using \texttt{fmincon} with the sequential
quadratic-programming algorithm. A fixed initial guess,
\begin{equation}
    \q_0=[3.7,\;0.2,\;4.7]^T\ \mathrm{rad},
\end{equation}
is supplied at each evaluation. No explicit intersample continuity
penalty or previous-solution warm start is used in the implemented
inverse-kinematics objective.

For the archived primary trajectory, the resulting desired joint
sequence is smooth, with a maximum absolute sample-to-sample
joint-coordinate change of approximately $0.268^\circ$ at the
$0.01$-s sampling interval. This observed smoothness is specific to
the tested trajectory and fixed initialization and should not be
interpreted as a formal guarantee against inverse-kinematic branch
switching for arbitrary trajectories.

The active limits used by the embedded Simulink implementation and
the reported final analyses are
\begin{equation}
    \q_{\min}=\frac{\pi}{180}[80,2,250]^T,\qquad
    \q_{\max}=\frac{\pi}{180}[250,160,330]^T.
\label{eq:limits}
\end{equation}
Angles are stored and evaluated in radians, while degrees are used
only for reporting the bounds. Desired velocities and accelerations
are obtained by filtered numerical differentiation.

The same joint limits are used by the active embedded
inverse-kinematics implementation and the maintained standalone
inverse-kinematics dependency. Historical source files retained in
the archive are not used for the reported simulations or analyses.

It is important to distinguish the feasible sets used later. The
\emph{geometric workspace} contains end-effector points reachable
under the joint limits. The
\emph{sampled nonnegative-tension-feasible workspace} is the subset
for which the implemented static torque demand produces a
nonnegative tension vector. \emph{Trajectory feasibility} concerns
only the tested path. Full dynamic and hardware feasibility
additionally requires cable elasticity, actuator dynamics,
pretension, and physical lower and upper force bounds, which define
the next hardware-oriented validation stage.

\FloatBarrier

%% file: sections/03_dynamics_and_allocation.tex
\section{Dynamic modeling and cable-force allocation}

\begin{revisionblock}
The equations below distinguish the nominal inverse-dynamics torque request from its algebraic cable-force mapping in the archived Simulink study. The authors confirmed the original model's signal path for the reported results: requested generalized torque passes through numerical force reconstruction and torque backprojection before the plant-input tag; model-specific feasibility or fallback handling can modify the applied generalized torque. The later force-bounded allocation studies use separate numerical model configurations. None of these paths is a physically validated force servo, and cable elasticity, identified motor or pulley dynamics and force sensing were not modeled in the original experiments.
\end{revisionblock}

\subsection{Implemented joint-space dynamic model}

For the planar three-link mechanism, the equations of motion are written as
\begin{equation}
    \D(\q)\qdd+\C(\q,\qd)+\G(\q)=\bm{\tau}_{a}+\taud,
\label{eq:dynamics}
\end{equation}
where $\D(\q)\in\mathbb{R}^{3\times3}$ denotes the inertia matrix used by the numerical implementation, $\C(\q,\qd)\in\mathbb{R}^{3}$ collects Coriolis and centrifugal terms, $\G(\q)\in\mathbb{R}^{3}$ is the gravity vector, $\bm{\tau}_{a}$ is the generalized actuation command, and $\taud$ is an additive external disturbance torque. The nominal physical parameters are listed in Table~\ref{tab:parameters}.

\begin{table}[!htbp]
\centering
\caption{Nominal robot parameters, cable anchors, and active joint limits.}
\label{tab:parameters}
\begin{tabular}{lll}
\toprule
Quantity & Value & Description \\
\midrule
$b_1,b_2,b_3$ & $0.45,0.35,0.21$ m & Calibrated nominal link lengths \\
$m_1,m_2,m_3$ & $11.125,5.05,1.38$ kg & Link masses \\
$I_{cg,1},I_{cg,2},I_{cg,3}$ & $0.149187,0.0522254,0.004784$ kg m$^2$ & COM inertias \\
$g$ & $9.81$ m s$^{-2}$ & Gravitational acceleration \\
$\q_{\min}$ & $[80,2,250]^T$ deg & Lower joint limits \\
$\q_{\max}$ & $[250,160,330]^T$ deg & Upper joint limits \\
$\bm{a}_1,\bm{a}_2,\bm{a}_3$ & $[0,1.5]^T,[0,-1.5]^T,[-0.4,1.5]^T$ m & Fixed cable anchors \\
\bottomrule
\end{tabular}
\end{table}

The simulated controller and plant use the same joint-space equation structure, with nominal and perturbed coefficients, respectively. Define
\begin{equation}
\begin{gathered}
    I_{o1}=I_{cg,1}+m_1\left(\frac{b_1}{2}\right)^2,\quad
    I_{o2}=I_{cg,2}+m_2\left(\frac{b_2}{2}\right)^2,\quad
    I_{o3}=I_{cg,3}+m_3\left(\frac{b_3}{2}\right)^2,\\
    c_2=\cos q_2,\qquad c_3=\cos q_3,\qquad c_{23}=\cos(q_2+q_3).
\end{gathered}
\label{eq:implemented_terms}
\end{equation}
The inertia matrix implemented in the reported simulations is
\begin{equation}
    \D(\q)=
    \begin{bmatrix}
        D_{11} & D_{12} & D_{13}\\
        D_{12} & D_{22} & D_{23}\\
        D_{13} & D_{23} & D_{33}
    \end{bmatrix},
\label{eq:D_matrix}
\end{equation}
with
\begin{align}
D_{11}={}&I_{o1}+m_2\left(b_1^2+b_1b_2c_2+\frac{b_2^2}{4}\right)\nonumber\\
&+m_3\left(b_1^2+b_2^2+\frac{b_3^2}{4}+2b_1b_2c_2+b_1b_3c_{23}+b_2b_3c_3\right),\\
D_{12}={}&m_2\left(\frac{b_1b_2c_2}{2}+\frac{b_2^2}{4}\right)\nonumber\\
&+m_3\left(b_2^2+\frac{b_3^2}{4}+b_1b_2c_2+\frac{b_1b_3c_{23}}{2}+b_2b_3c_3\right),\\
D_{13}={}&m_3\left(\frac{b_3^2}{4}+\frac{b_1b_3c_{23}}{2}+\frac{b_2b_3c_3}{2}\right),\\
D_{22}={}&I_{o2}+\frac{m_2b_2^2}{4}+m_3\left(b_2^2+b_2b_3c_3+\frac{b_3^2}{4}\right),\\
D_{23}={}&m_3\left(\frac{b_2b_3c_3}{2}+\frac{b_3^2}{4}\right),\\
D_{33}={}&I_{o3}+\frac{m_3b_3^2}{4}.
\label{eq:D_entries}
\end{align}

The Coriolis/centrifugal vector in the archived model is represented using the Christoffel coefficients,
\begin{equation}
    C_i(\q,\qd)=\sum_{j=1}^{3}\sum_{k=1}^{3}c_{ijk}(\q)\dot q_j\dot q_k,
\end{equation}
\begin{equation}
    c_{ijk}=\frac{1}{2}\left(
    \frac{\partial D_{ij}}{\partial q_k}+
    \frac{\partial D_{ik}}{\partial q_j}-
    \frac{\partial D_{jk}}{\partial q_i}\right).
\label{eq:christoffel}
\end{equation}

With the $x$-axis vertical and positive upward, the potential energy used by the implemented model is
\begin{equation}
\begin{split}
V(\q)=g\Bigg[&b_1\left(\frac{m_1}{2}+m_2+m_3\right)\cos q_1
+b_2\left(\frac{m_2}{2}+m_3\right)\cos(q_1+q_2)\\
&+\frac{b_3m_3}{2}\cos(q_1+q_2+q_3)\Bigg],
\end{split}
\label{eq:potential}
\end{equation}
and therefore
\begin{equation}
\G(\q)=-g
\begin{bmatrix}
 b_1(\frac{m_1}{2}+m_2+m_3)\sin q_1+b_2(\frac{m_2}{2}+m_3)\sin(q_1+q_2)+\frac{b_3m_3}{2}\sin(q_1+q_2+q_3)\\
 b_2(\frac{m_2}{2}+m_3)\sin(q_1+q_2)+\frac{b_3m_3}{2}\sin(q_1+q_2+q_3)\\
 \frac{b_3m_3}{2}\sin(q_1+q_2+q_3)
\end{bmatrix}.
\label{eq:G_vector}
\end{equation}
Equations~\eqref{eq:D_matrix}--\eqref{eq:G_vector} document the dynamics implemented in the reported numerical evaluation.

\subsection{Nominal controller model, uncertain plant, and disturbance}

The CTC branch uses fixed nominal quantities $\D_n(\q)$, $\C_n(\q,\qd)$, and $\G_n(\q)$. The geometric parameters used for inverse and forward kinematics are assumed to be measured and calibrated before operation and remain fixed at their nominal values:
\begin{equation}
    b_i^{\mathrm{kin}}(t)=b_{i,n}.
\label{eq:fixed_kinematic_geometry}
\end{equation}

In the archived runs, selected parameters used only inside the simulated plant dynamics were simultaneously increased by 10\% during $1<t<5$ s. The implemented perturbation is
\begin{equation}
\begin{split}
 b_{i,p}^{\mathrm{dyn}}(t)
 &=b_{i,n}\left[1+0.1\chi_{(1,5)}(t)\right],\\
 m_{i,p}(t)
 &=m_{i,n}\left[1+0.1\chi_{(1,5)}(t)\right],\\
 I_{cg,i,p}(t)
 &=I_{cg,i,n}\left[1+0.1\chi_{(1,5)}(t)\right],
\end{split}
\label{eq:uncertainty}
\end{equation}
where $\chi_{(1,5)}$ is one inside the indicated interval and zero otherwise. Here, $b_{i,p}^{\mathrm{dyn}}$ denotes the effective length parameters appearing inside the plant expressions for $\D_p$, $\C_p$, and $\G_p$. These perturbed quantities are not used to redefine the robot geometry in the inverse- or forward-kinematic calculations.

Accordingly, the uncertainty case represents a structured dynamic-model mismatch rather than an unknown physical change in the calibrated robot geometry. It is intended to approximate patient-dependent and attachment-dependent deviations in the effective plant dynamics, including changes in inertial loading and model coefficients that are not represented by the nominal controller model. This separation between fixed calibrated kinematics and perturbed dynamic coefficients is a modeling abstraction used to evaluate controller sensitivity to dynamic mismatch.

The tested perturbation is time-localized, simultaneous, and deterministic. It is not an independently sampled uncertainty distribution and does not represent a statistical population model of patient variability.

When the disturbance case is enabled, three band-limited white-noise sources are gated into the joint-torque channels during $5<t<9$ s. The maintained Simulink model specifies noise power 0.1, correlation/sample parameter 0.5 s, and nominal seeds 1001, 2001, and 3001 for joints 1--3; these seed values are documented for the case-matrix workflow. Seed-based diagnostics overwrite the random-source seeds while retaining the same plant, trajectory, and controller settings. Each paired CTC/residual comparison within a saved run uses the same disturbance realization. The complete random-source state of the separately archived representative run was not retained. The resulting uncertain plant can be written as
\begin{equation}
    \D_p(\q,t)\qdd+\C_p(\q,\qd,t)+\G_p(\q,t)=\taucmd+\taud(t),
\label{eq:plant}
\end{equation}
\begin{revisionblock}
In Eq.~\eqref{eq:plant}, $\taucmd$ represents the \emph{requested} controller torque under the ideal torque-equivalent relation. The authors' archived Simulink model routes this request through algebraic cable-force reconstruction and torque backprojection to the numerical plant, so the applied generalized torque can depart from the request when an allocation fallback is active. The disturbance $\taud$ is a separate modeled input. The later explicitly force-constrained follow-up instead uses $\bm\tau_{\rm applied}=\Jc^T\F^*$, with separately assumed force limits and numerical lag (Section~\ref{sec:revision_methods}). The subscript $p$ denotes plant dynamics.
\end{revisionblock}

\subsection{Torque-to-tension calculation and pull-only feasibility}
\label{sec:allocation_scope}

For a requested generalized torque $\bm{\tau}^{\star}$, the cable mapping requires
\begin{equation}
    \Jc^T(\q)\F=\bm{\tau}^{\star},\qquad \F\geq\bm{0}.
\label{eq:allocation}
\end{equation}
In the present three-cable/three-DOF model, $\Jc^T\in\mathbb{R}^{3\times3}$. When it is full rank, the torque-equivalent vector is unique and can be written as
\begin{equation}
    \F_p=(\Jc^T)^{\dagger}\bm{\tau}^{\star}=(\Jc^T)^{-1}\bm{\tau}^{\star}.
\label{eq:pinv}
\end{equation}
Unlike a redundant cable robot, the square system has no nontrivial null-space direction with which to redistribute tension while preserving the same generalized torque. Therefore, a negative component of $\F_p$ cannot be repaired by a null-space pretension adjustment in this model.

\begin{revisionblock}
The author-confirmed archived model uses an algebraic torque-to-force computation and a reconstructed-torque connection to the numerical plant, with model-specific zero-output fallback when a lower-tension or nonnegativity condition fails. It is therefore not accurate to describe its actuation path as \emph{unconditionally direct injection of the requested torque}. The reported inferred cable-force log remains a numerical allocation-related signal, not a measured physical actuator input; zero entries can denote fallback rather than tension regulation. The workspace feasibility test evaluates raw signed force demand before the fallback. The separate bounded least-squares closed-loop experiments are reported only as supplemental simulations. Neither construction demonstrates a physical force servo, regulated pretension or successful physical cable-drive actuation.
\end{revisionblock}

A hardware-oriented extension could replace this logic with a constrained optimization layer, for example
\begin{equation}
\begin{split}
    \F^{\star}=\arg\min_{\F}\;&\|\Jc^T\F-\bm{\tau}^{\star}\|_2^2
    +\lambda\|\F-\F_{ref}\|_2^2,\\
    \mathrm{s.t.}\;&\F_{min}\leq\F\leq\F_{max},
\end{split}
\label{eq:QP}
\end{equation}
with limits obtained from motor torque, spool radius, cable strength, tension-sensor range, pulley friction, and application-specific safety requirements. In the square system, Eq.~\eqref{eq:QP} can also quantify torque residual when the requested command is infeasible. Because a physical $\F_{max}$ is not defined here, the manuscript reports nonnegative-tension checks and demanded force levels but does not claim actuator-saturation feasibility, optimal allocation, or guaranteed pretension.

\subsection{Rank, singular values, and conditioning}

The cable mapping is evaluated along the desired path using
\begin{equation}
    \sigma_{\min}(\Jc),\qquad
    \kappa_2(\Jc)=\frac{\sigma_{\max}(\Jc)}{\sigma_{\min}(\Jc)},\qquad
    \mathrm{rank}(\Jc).
\label{eq:conditioning}
\end{equation}
Rank three is required for a unique square mapping, but rank alone does not measure sensitivity. A small $\sigma_{\min}$ or large condition number indicates that small errors in torque, geometry, or numerical computation may produce comparatively large changes in the inferred tension. These diagnostics are therefore interpreted as trajectory-specific numerical conditioning indicators rather than as global controllability or safety certificates.

\FloatBarrier

%% file: sections/04_control_design.tex
\section{Control design}

The control architecture separates nominal nonlinear compensation from learned correction. CTC remains responsible for the main trajectory-tracking command, while DDPG is restricted to a bounded residual torque. This division is central to the method: the actor is not trained to replace the robot model or to generate the complete actuation command.

\subsection{Nominal computed torque control}

Define the joint-space position and velocity errors as
\begin{equation}
    \e=\q_d-\q,\qquad \ed=\dot{\q}_d-\qd.
\label{eq:tracking_errors}
\end{equation}
The nominal computed-torque command is
\begin{equation}
    \tauctc=\D_n(\q)\bm{v}+\C_n(\q,\qd)+\G_n(\q),
\label{eq:CTC}
\end{equation}
where the auxiliary acceleration is
\begin{equation}
    \bm{v}=\ddot{\q}_d+\bm{K}_d\ed+\bm{K}_p\e.
\label{eq:v}
\end{equation}
The fixed gains used for both the baseline and residual-controller branches are
\begin{equation}
    \bm{K}_p=60\bm{I}_3,\qquad \bm{K}_d=20\bm{I}_3.
\label{eq:gains}
\end{equation}
The subscript $n$ emphasizes that Eq.~\eqref{eq:CTC} uses nominal link parameters even when the simulated plant is uncertain. For an exact nominal plant and an unfiltered torque command, substitution into Eq.~\eqref{eq:dynamics} gives
\begin{equation}
    \ddot{\e}+\bm{K}_d\dot{\e}+\bm{K}_p\e=\bm{0},
\label{eq:error_dynamics}
\end{equation}
so positive-definite gains produce exponentially convergent ideal nominal error dynamics. This statement is limited to the exact-model idealization. The comparison branch includes first-order command filtering of the form $10/(s+10)$, and the uncertain/disturbed plant does not satisfy Eq.~\eqref{eq:error_dynamics} exactly. CTC is retained because it supplies interpretable model-based compensation and a strong baseline rather than because it is assumed to reject arbitrary mismatch.

\subsection{Bounded residual DDPG compensation}

The residual architecture forms the combined commanded joint torque
\begin{equation}
    \taucmd=\tauctc+\taurl.
\label{eq:total}
\end{equation}
\begin{revisionblock}
The originally reported comparisons are described as generalized-torque requests evaluated through the author-verified algebraic cable-force reconstruction and torque-backprojection signal path, which includes model-specific zero-force fallback. The applied generalized input may therefore differ from the requested controller torque. Neither the logged numerical cable demand nor the reported tracking improvement establishes physical cable-force actuation. Separate follow-up simulations explicitly use bounded force allocation to generate applied generalized torque, and these are reported under different experimental protocols in Section~\ref{sec:revision_methods}.
\end{revisionblock} The externally applied disturbance $\taud$ enters the simulated plant separately in Eq.~\eqref{eq:plant}. The learned component is bounded componentwise by
\begin{equation}
    -[5,3,2]^T\leq\taurl\leq[5,3,2]^T\quad \mathrm{N\,m}.
\label{eq:bounds}
\end{equation}
These limits constrain the maximum policy authority, preserve CTC as the nominal command source, and permit direct inspection of how frequently the actor approaches its joint-wise action bounds. Bounds alone do not prove that the residual is smaller than CTC at every instant; the relative contribution is therefore evaluated empirically using normalized action peaks and residual-to-CTC RMS torque ratios.

Under model mismatch, the residual role can be interpreted schematically as
\begin{equation}
    \ddot{\e}+\bm{K}_d\dot{\e}+\bm{K}_p\e
    =\bm{\Delta}(\q,\qd,t)-\D_p^{-1}(\q,t)\taurl,
\label{eq:residual_interpretation}
\end{equation}
where $\bm{\Delta}$ groups imperfect cancellation, filtering effects, and the external disturbance after nominal compensation. The actor is trained to reduce the tracking consequences of this aggregate term, and Eq.~\eqref{eq:residual_interpretation} provides the control-oriented interpretation used in the subsequent analysis.

\begin{figure}[!t]
    \centering
    \includegraphics[width=0.96\textwidth]{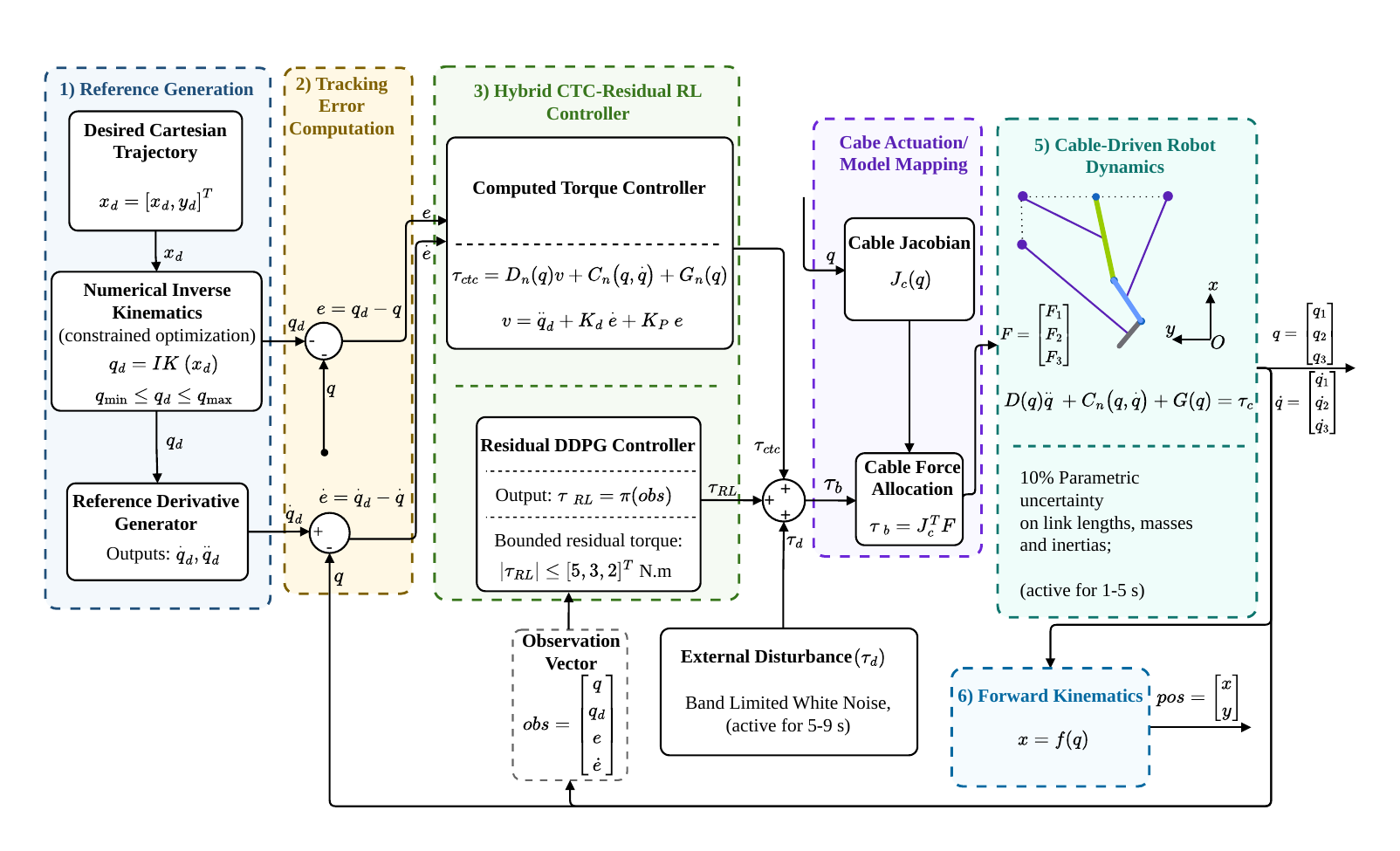}
    \caption{\revcaption{Hybrid CTC--residual DDPG control architecture and parallel cable-demand audit. Constrained reference generation supplies the desired joint states and derivatives; CTC produces the nominal model-based torque $\tauctc$, and the trained actor supplies the bounded correction $\taurl$. Their sum is the requested generalized torque. The authors verified the original numerical force-reconstruction and torque-backprojection signal path, including model-specific fallback; the later force-limited simulations use distinct configurations. A separate logged algebraic cable-demand signal supports feasibility diagnosis but must not be mistaken for measured physical cable tension. The external disturbance $\taud$ enters the plant separately. Plant-dynamics parameter mismatch is active from 1 to 5~s, while the calibrated kinematic geometry remains fixed; disturbance is active from 5 to 9~s.}}
    \label{fig:ctc_residual_loop}
\end{figure}

\begin{figure}[!htbp]
    \centering
    \includegraphics[width=0.96\textwidth]{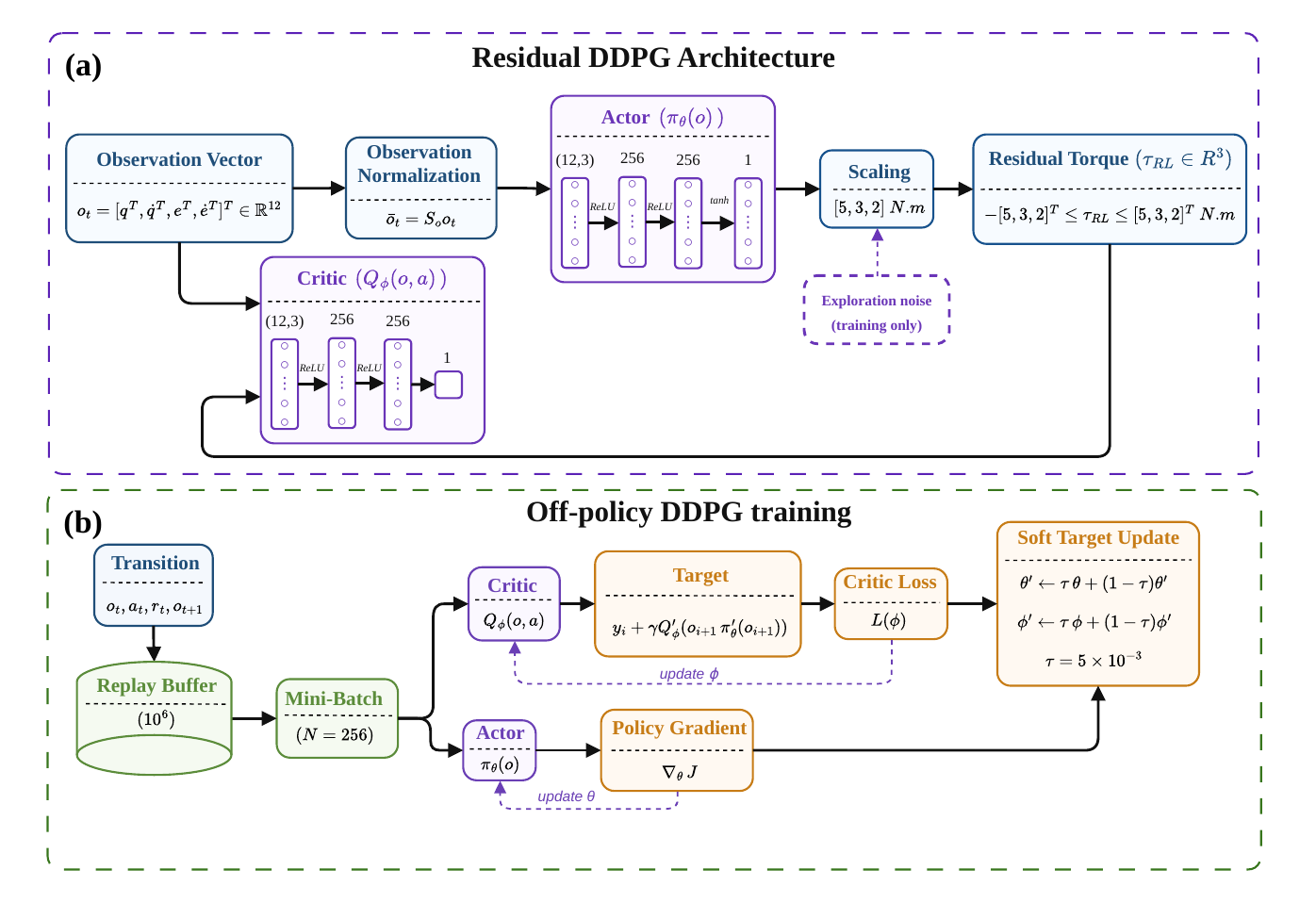}
    \caption{Residual DDPG actor--critic structure and off-policy training. The unnormalized 12-dimensional observation is mapped by the actor to three bounded residual torques. The critic estimates $Q(\bm{o},\bm{a})$ from separate observation and action feature paths. Replay-buffer sampling, gradient updates, and soft target-network updates are used during training; deterministic evaluation uses the saved actor without exploration noise or online learning.}
    \label{fig:ddpg_structure_clean}
\end{figure}

The actor observation is
\begin{equation}
    \bm{o}_t=[\q^T,\qd^T,\e^T,\ed^T]^T\in\mathbb{R}^{12},
\label{eq:obs}
\end{equation}
and the action is $\bm{a}_t=\taurl\in\mathbb{R}^{3}$. The archived network applies no automatic input normalization. The actor comprises a 12-element feature input, two fully connected layers of 256 neurons with ReLU activations, a three-neuron output, a hyperbolic tangent, and joint-wise scaling to Eq.~\eqref{eq:bounds}. The critic has separate observation and action paths: each is projected to 256 features, the paths are added, and two subsequent ReLU/fully connected stages produce the scalar action-value estimate $Q(\bm{o},\bm{a})$. Figure~\ref{fig:ctc_residual_loop} shows the closed-loop signal flow, and Fig.~\ref{fig:ddpg_structure_clean} shows the actor--critic training structure.

\begin{table}[!htbp]
\centering
\caption{Archived residual DDPG implementation and training settings.}
\label{tab:DDPG}
\begin{tabular}{ll}
\toprule
Parameter & Value \\
\midrule
Observation/action dimensions & 12 / 3 \\
Residual torque bounds & $\pm[5,3,2]^T$ N m \\
Actor hidden layers & 256 and 256 ReLU neurons \\
Critic feature width & 256 neurons per path/stage \\
Agent sample time & $T_s=0.01$ s \\
Discount factor & $\gamma=0.99$ \\
Minibatch size & 256 \\
Replay-buffer capacity & $10^6$ transitions \\
Target-network smoothing factor & $5\times10^{-3}$ \\
Actor/critic optimizer options & MATLAB R2022b defaults (not overridden) \\
Exploration-noise variance & 0.05 \\
Exploration-variance decay rate & $10^{-3}$ \\
Maximum episodes & 300 \\
Steps per episode & 1000 (10 s at $T_s=0.01$ s) \\
Score-averaging window & 20 episodes \\
Stop condition & average reward greater than $-0.05$ \\
Evaluated checkpoint & archived \texttt{Agent20\_3cables.mat} file \\
\bottomrule
\end{tabular}
\end{table}

\subsection{Reward and training objective}

At each agent step, the reward penalizes joint error, error rate, normalized residual magnitude, and changes in normalized residual torque:
\begin{equation}
\begin{split}
    r_t={}&-w_e\e^T\bm{Q}_e\e-w_{\dot e}\ed^T\bm{Q}_{\dot e}\ed
    -w_u\bar{\bm{u}}^T\bar{\bm{u}}
    -w_{\Delta u}\Delta\bar{\bm{u}}^T\Delta\bar{\bm{u}},
\end{split}
\label{eq:reward}
\end{equation}
where $\bar{\bm{u}}$ denotes the residual torque normalized by its joint-wise bound and $\Delta\bar{\bm{u}}=\bar{\bm{u}}_t-\bar{\bm{u}}_{t-1}$. The matrices and scalar weights are
\begin{equation}
    \bm{Q}_e=\mathrm{diag}(25,20,25),\qquad
    \bm{Q}_{\dot e}=\mathrm{diag}(6,5,6),
\end{equation}
\begin{equation}
    w_e=2.0,\quad w_{\dot e}=0.35,\quad w_u=0.05,\quad w_{\Delta u}=0.20.
\end{equation}
The last two terms discourage a large or rapidly varying learned correction. The archived training script uses the fixed 1000-step horizon and the average-reward stopping criterion in Table~\ref{tab:DDPG}; no separate state-based early-termination condition or episode-level reset/randomization function is documented. The script does not override the actor or critic optimizer options, so the MATLAB R2022b defaults apply. Evaluation uses the archived checkpoint designated \texttt{Agent20\_3cables.mat}. Genuine saved episode-by-episode training statistics and the original checkpoint-selection rationale were not retained, so no training-curve figure or claim of checkpoint optimality is presented.

The current reward emphasizes tracking and residual-action regularization. Cable-tension margin, joint-limit proximity, interaction force, comfort-related quantities, and actuator power are therefore identified as explicit objectives for the next constraint-aware training stage described in Section~7.

\subsection{Training and deterministic evaluation sequence}

Table~\ref{tab:algorithm} separates learning-time operations from the deployed evaluation loop. During training, exploration is added to the actor action, transitions are stored in the replay buffer, and minibatches update the critic and actor; target networks are moved toward the online networks using the smoothing factor. During evaluation, the saved trained actor is used deterministically and its parameters are not updated.

\begin{table}[!htbp]
\centering
\caption{Controller execution and DDPG training/evaluation sequence.}
\label{tab:algorithm}
\begin{tabular}{p{0.09\textwidth}p{0.82\textwidth}}
\toprule
Step & Operation \\
\midrule
1 & Read $\q,\qd$ and form $\e,\ed$ using the constrained reference. \\
2 & Evaluate $\D_n,\C_n,\G_n$ and compute $\tauctc$ from Eqs.~\eqref{eq:CTC}--\eqref{eq:v}. \\
3 & Form $\bm{o}_t$ and evaluate the actor; add exploration only during training. \\
4 & Scale the actor output to the bounds in Eq.~\eqref{eq:bounds} and form $\taucmd=\tauctc+\taurl$. \\
5 & Apply the numerically reconstructed generalized input, including any model-specific fallback, integrate the uncertain/disturbed plant, and compute the reward. \\
6 & During training, store $(\bm{o}_t,\bm{a}_t,r_t,\bm{o}_{t+1})$, sample a minibatch, update critic and actor, and softly update target networks. \\
7 & During evaluation, log tracking, residual usage, inferred tension, rank/conditioning, and joint-limit indicators without network updates. \\
\bottomrule
\end{tabular}
\end{table}

%% file: sections/05_simulation_method.tex
\section{Simulation model-consistency assessment and evaluation method}

The evaluation addresses three questions: whether the analytical implementation is consistent with independent multibody models, whether CTC and residual DDPG are compared under identical conditions, and whether the resulting trajectory and inferred cable demand satisfy the tested geometric, tension, and joint-limit constraints. The controlled simulation environment provides repeatable comparisons across all scenarios.

\subsection{Implementation and model-consistency checks}

The generalized-coordinate plant, nominal CTC, trained DDPG actor, uncertainty and disturbance gates, cable-Jacobian calculation, algebraic torque-to-tension mapping, and post-processing routines are implemented in MATLAB/Simulink R2022b. The archived main model runs for 10 s with a 0.01 s fixed step and an ode3 fixed-step integration setting. The DDPG action is updated at the same 0.01 s sample time. The nominal initial Cartesian position is $\x_0=[-0.90,-0.05]^T$ m, and the corresponding joint state is obtained from the constrained inverse-kinematics routine.

\begin{figure}[!htbp]
    \centering
    \includegraphics[width=0.88\textwidth]{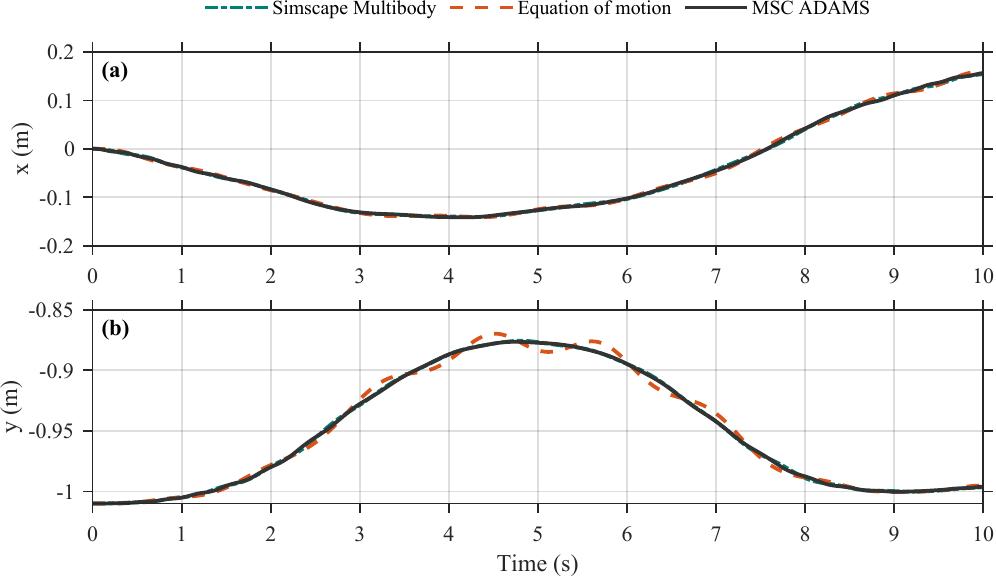}
    \caption{Cartesian model-consistency comparison for the analytical equation-of-motion, Simscape Multibody, and MSC ADAMS implementations under the same nominal reference. Panels (a) and (b) compare the end-effector $x$- and $y$-coordinate histories, respectively.}
    \label{fig:validation_xy}
\end{figure}

\begin{figure}[!htbp]
    \centering
    \includegraphics[width=0.88\textwidth]{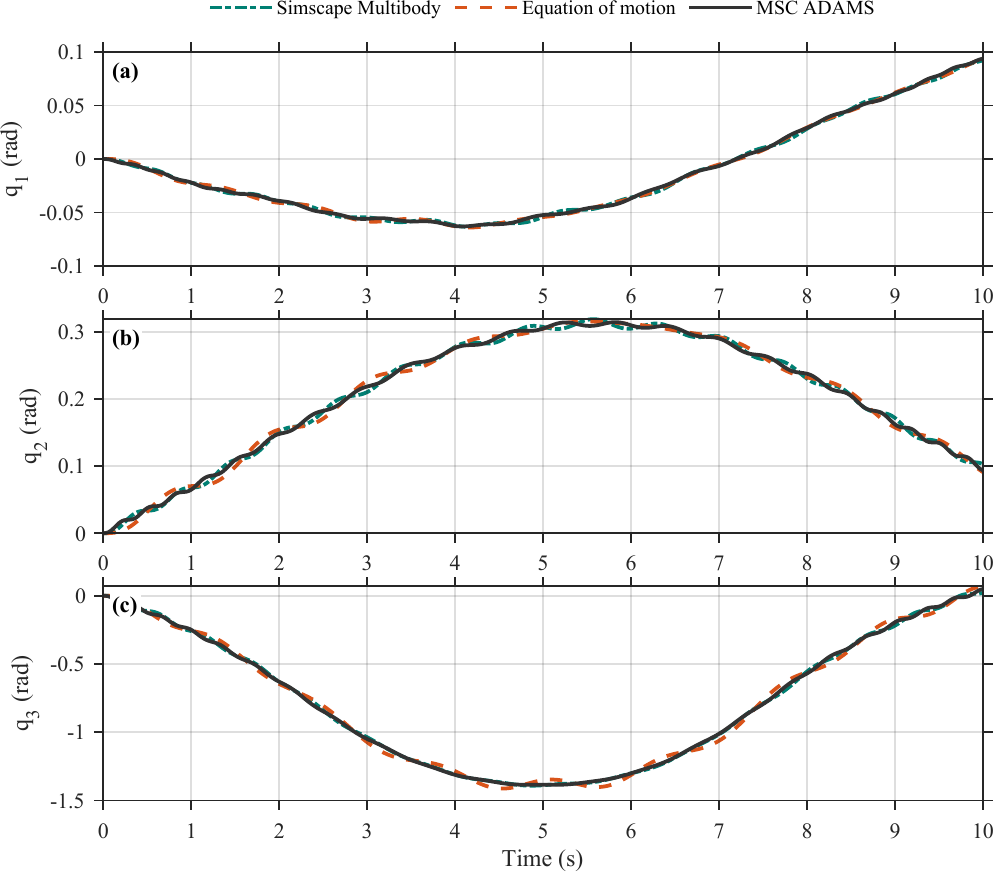}
    \caption{Joint-angle model-consistency comparison among the analytical equation-of-motion, Simscape Multibody, and MSC ADAMS implementations. Panels (a)--(c) report $q_1$, $q_2$, and $q_3$, respectively, under the same nominal reference.}
    \label{fig:validation_joint}
\end{figure}

\begin{figure}[!htbp]
    \centering
    \includegraphics[width=0.86\textwidth]{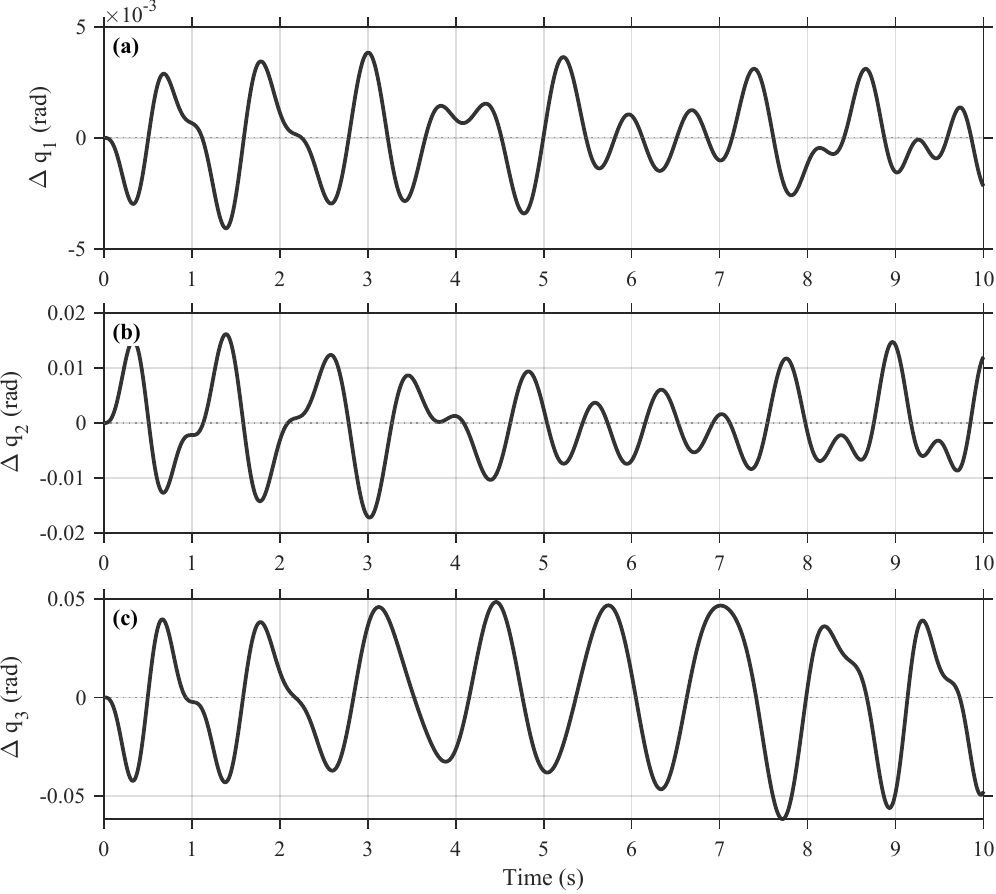}
    \caption{Simscape-minus-equation-of-motion joint-angle differences for $q_1$, $q_2$, and $q_3$ in panels (a)--(c), respectively. The difference histories localize the remaining implementation discrepancy, which is largest in the third joint channel.}
    \label{fig:validation_error}
\end{figure}

The simulated equation-of-motion implementation was compared with Simscape Multibody and MSC ADAMS trajectory records for nominal geometry and reference motion. Trajectory agreement is a numerical cross-model diagnostic rather than a kinetic-energy identity or hardware validation. Figures~\ref{fig:validation_xy}, \ref{fig:validation_joint}, and~\ref{fig:validation_error} compare the Cartesian coordinates, joint angles, and Simscape-minus-equation-of-motion joint differences. These comparisons are intended to detect sign, unit, joint-order, and implementation inconsistencies rather than to provide experimental validation.

Quantitative discrepancies were calculated by aligning the archived simulation histories over their common time interval. For the ADAMS comparison, the joint-angle histories were reconstructed from the archived angular-velocity signals using the same numerical integration and unit conversion employed in the corresponding figure-generation workflow. Table~\ref{tab:model_consistency_metrics} reports the root-mean-square and maximum absolute discrepancies relative to the analytical equation-of-motion implementation.

\begin{table}[!htbp]
\centering
\caption{Archived model-consistency discrepancies relative to the analytical equation-of-motion implementation. Each entry reports RMS~$/$~maximum absolute discrepancy.}
\label{tab:model_consistency_metrics}
\begin{tabular}{llll}
\toprule
Signal & Unit & Simscape--EOM & ADAMS--EOM \\
\midrule
$q_1$ & deg & $0.102\,/\,0.233$ & $0.091\,/\,0.224$ \\
$q_2$ & deg & $0.409\,/\,0.984$ & $0.362\,/\,0.832$ \\
$q_3$ & deg & $1.715\,/\,3.537$ & $1.684\,/\,3.023$ \\
$x$   & mm  & $3.050\,/\,7.877$ & $2.959\,/\,7.812$ \\
$y$   & mm  & $3.742\,/\,8.628$ & $3.692\,/\,8.609$ \\
\bottomrule
\end{tabular}
\end{table}

Across the archived comparisons, the RMS discrepancy ranges from approximately $0.98\%$ to $2.67\%$ of the corresponding analytical motion range, while the maximum absolute discrepancy ranges from approximately $2.47\%$ to $6.15\%$. The third joint has the largest absolute joint-angle discrepancy, and the $y$ coordinate has the largest range-normalized discrepancy. The similar Simscape and ADAMS histories show trajectory-level agreement for the evaluated motion; this comparison alone does not establish complete dynamic equivalence. These results remain simulation-to-simulation comparisons and do not constitute hardware, experimental, or clinical validation.

The model-consistency workflow and the interpretation assigned to each check are summarized in Table~\ref{tab:validation_scope}. Cable-force post-processing uses the same $\Jc$ sign convention as the Simulink allocation block. Workspace sampling and path-conditioning calculations are run independently of the tracking comparison, reducing the risk that a plotting or logging branch is mistaken for a controller result.

\begin{table}[!htbp]
\centering
\caption{Purpose and interpretive scope of the simulation checks.}
\label{tab:validation_scope}
\begin{tabularx}{\textwidth}{
>{\raggedright\arraybackslash}p{0.20\textwidth}
>{\raggedright\arraybackslash}X
>{\raggedright\arraybackslash}X}
\toprule
Check & Purpose & Interpretation \\
\midrule
Forward/inverse kinematics & Verify Cartesian-reference and joint-coordinate consistency & Kinematic consistency of the implemented reference-generation chain \\
Simscape Multibody & Compare the analytical dynamics with an independent multibody formulation & Nominal simulation-to-simulation consistency under matched parameters \\
MSC ADAMS & Provide an additional independent multibody implementation check & Cross-platform nominal dynamic consistency \\
Tension post-processing & Audit the pull-only cable-demand mapping & Generalized-torque demand mapped to inferred cable tension \\
Workspace and $\Jc$ analysis & Evaluate sampled reachability, static nonnegative-tension feasibility, rank, and conditioning & Sampled, trajectory-specific cable-feasibility evidence \\
\bottomrule
\end{tabularx}
\end{table}

\subsection{Primary comparison protocol}

In the archived experiment, baseline and residual-controller branches used the same desired Cartesian and joint trajectories, nominal model, CTC gains, simulation duration, integration step, initial state, uncertainty timing, and disturbance realization. The only controller difference is the addition of the deterministic saved actor output in Eq.~\eqref{eq:total}. The actor is not retrained, adapted, or updated between evaluation cases. This paired design prevents a favorable disturbance draw or different initial condition from being attributed to the residual policy. The primary trajectory is the reference used by the archived training environment; consequently, the main comparisons assess compensation on a known trajectory rather than broad trajectory generalization. The circular-path test in Section~\ref{sec:circular} is reported separately as an alternate-trajectory diagnostic.

Table~\ref{tab:simulation_settings} consolidates settings not already listed in Tables~\ref{tab:parameters} and~\ref{tab:DDPG}.

\begin{table}[!htbp]
\centering
\caption{Simulation and comparison settings for the reported numerical evaluation.}
\label{tab:simulation_settings}
\begin{tabularx}{\textwidth}{
>{\raggedright\arraybackslash}p{0.30\textwidth}
>{\raggedright\arraybackslash}X}
\toprule
Setting & Value \\
\midrule
Simulation duration / fixed step & 10 s / 0.01 s \\
Archived solver setting & Fixed-step ode3 under FixedStepAuto \\
Nominal Cartesian initial state & $[-0.90,-0.05]^T$ m \\
Primary trajectory & Eq.~\eqref{eq:main_path} \\
CTC gains & $K_p=60I_3$, $K_d=20I_3$ \\
Uncertainty & Simultaneous $+10\%$ perturbation of masses, COM inertias, and length-dependent plant-dynamics coefficients; calibrated kinematic geometry remains fixed \\
Uncertainty interval & 1--5 s \\
Disturbance source & Three band-limited white-noise torque channels \\
Disturbance interval & 5--9 s \\
Case-matrix noise seeds & 1001, 2001, 3001; the exact random-source state of the separately archived representative run was not retained \\
Primary controller comparison & CTC versus CTC + saved residual DDPG actor \\
\bottomrule
\end{tabularx}
\end{table}

The four primary cases are defined in Table~\ref{tab:scenarios}. ``Uncertainty only'' isolates model mismatch, ``disturbance only'' isolates the external torque input, and ``combined'' activates the two effects in consecutive intervals. The nominal case indicates whether the residual policy degrades or improves an already matched model.

\begin{table}[!htbp]
\centering
\caption{Four primary evaluation scenarios.}
\label{tab:scenarios}
\begin{tabular}{lll}
\toprule
Case & Parametric uncertainty & External disturbance \\
\midrule
C1: nominal & inactive & inactive \\
C2: uncertainty only & 10\%, 1--5 s & inactive \\
C3: disturbance only & inactive & 5--9 s \\
C4: combined & 10\%, 1--5 s & 5--9 s \\
\bottomrule
\end{tabular}
\end{table}

\subsection{Metrics and constraint indicators}

Let $\bm{e}_x(t)=\xd(t)-\x(t)\in\mathbb{R}^{2}$ and $e_x(t)=\|\bm{e}_x(t)\|_2$. For $N$ uniformly sampled values, the reported tracking metrics are
\begin{equation}
    e_{\mathrm{RMS}}
    =
    \sqrt{\frac{1}{N}\sum_{k=1}^{N}e_x^2(t_k)},
    \qquad
    e_{\mathrm{peak}}
    =
    \max_k e_x(t_k),
\label{eq:rms_peak}
\end{equation}
\begin{equation}
    \mathrm{IAE}
    =
    \int_0^T e_x(t)\,dt,
    \qquad
    \mathrm{ISE}
    =
    \int_0^T e_x^2(t)\,dt.
\label{eq:iae_ise}
\end{equation}
RMS characterizes the average error magnitude, peak error captures the worst logged deviation, IAE accumulates absolute displacement error, and ISE emphasizes larger excursions. Percentage improvement is computed as $100(M_{CTC}-M_{RL})/M_{CTC}$ for metrics in which a lower value is preferable.

Cable indicators include minimum and maximum values of the logged algebraic cable-demand audit output and whether that output reaches the zero-tension boundary. The allocation block replaces a rejected nonnegative-tension solution with a zero fallback; therefore, the logged signal is nonnegative by construction and a zero count of negative logged values is not independent evidence of dynamic cable feasibility. A channel-sample count sums violations over joints or cables and time; it is not the number of independent physical events. Joint-limit indicators use the active bounds in Eq.~\eqref{eq:limits}. Cable-mapping indicators are the rank, minimum singular value, and condition number of $\Jc$ along the desired path. Because $F_{max}$ is undefined, maximum demanded tension is reported but not classified as feasible or infeasible.

\subsection{Robustness and diagnostic extensions}

Four additional tests characterize sensitivity and transfer beyond the representative run. First, the disturbance-enabled cases are repeated over ten source seeds; deterministic nominal and uncertainty-only cases are consequently identical across those seed runs, while disturbance cases exhibit variability. Mean and standard deviation are reported descriptively; no inferential statistical test is claimed. Second, metrics are separated into the initial transient (0--1 s), uncertainty (1--5 s), disturbance (5--9 s), and recovery (9--10 s) intervals. Third, four nearby Cartesian initial positions are evaluated with the same actor and controller parameters. Fourth, a circular path is tested after confirming that the trajectory-mode switch changes the reference. This trajectory-transfer evaluation reports tracking and joint-limit behavior together so that Cartesian transfer and joint-coordinate coverage remain explicitly distinguished.

\FloatBarrier

%% file: sections/05b_revision_methods.tex
\section{Supplementary experimental methods introduced in the revision}\label{sec:revision_methods}
\begin{revisionblock}
The primary archived comparisons and the studies added during peer review are distinct experiments. Unless otherwise stated, the follow-up tests use a 10~s simulated trajectory sampled every 0.01~s, the same nominal CTC gains ($K_p=60 I_3$, $K_d=20 I_3$), and the frozen submitted checkpoint \texttt{Agent20\_3cables.mat}. Values reported for separate controller configurations or training campaigns are not pooled. ``Completion'' means that the recorded simulation reached its specified horizon without a detected terminating joint-limit crossing; it does not prove continuous-time constraint invariance. All force bounds and actuator time constants used below are assumed numerical parameters, not experimentally identified hardware ratings.
\end{revisionblock}

\subsection{Original checkpoint provenance and newly trained policies}
\begin{revisionblock}
The original evaluated policy is \texttt{Agent20\_3cables.mat}, whose SHA-256 digest begins \texttt{e4afcbc45943b3d3}. The archived training configuration used 12 observations, three outputs, two 256-neuron actor hidden layers, discount factor $0.99$, minibatch size 256, replay capacity $10^6$, target smoothing $0.005$, exploration variance $0.05$ (decay $0.001$), and sample time $0.01$~s. An upper training budget of 300 episodes, each of 1{,}000 steps, was configured. A subsequent read-only recovery yielded \emph{only} episodes 1--20, including recorded rewards and critic quantities. The original full learning curve, seed and random state, candidate-score records, and contemporaneous criterion for selecting \texttt{Agent20} did not survive. Historical convergence and original checkpoint optimality are not established.

For an explicitly separate reproducibility investigation, five DDPG agents were independently initialized with seeds 20261028--20261101 and trained for 80 episodes of 1{,}000 steps. The revised optimizer settings included actor and critic learning rates 0.01 and 0.001, respectively; therefore this campaign is \emph{not} a bitwise replication of the archived training. Episode-60 and episode-80 checkpoint tests were defined before evaluation. Eligibility required completion of the 10~s nominal reference, Cartesian RMSE at most 20~mm, zero sampled joint-limit violations, at most 5\% residual-action near-saturation (at 98\% of any native bound), and finite logged critic $Q_0$. Failure cases were retained. Additional matched three-seed, one-factor-at-a-time runs varied $\gamma$ from 0.99 to 0.95 or exploration-noise variance from 0.05 to 0.10, without changing both at once.

A separate matched-budget DDPG/TD3/SAC comparison used five training initializations per algorithm and 60 episodes per run. Each algorithm retained its native actor--critic and exploration mechanisms; equal episode budgets do not imply identical internal updates. Candidate checkpoints were screened on two synthetic disturbances (W1--W2) for completion without limit violations, then one selected policy per seed was frozen and evaluated on W11--W14. The original \texttt{Agent20} was a distinct, unchanged reference and was not counted as a sixth freshly trained DDPG seed. The four evaluation waveforms were excluded from this campaign's checkpoint-screening step but are not claimed to be universally unseen across earlier project investigations.
\end{revisionblock}

\subsection{Residual-action bounds and reward investigations}
\begin{revisionblock}
The original residual command satisfies $\bm{\tau}_{\rm RL}=\bm{b}\odot\tanh(\bm{a})$, with empirically chosen $\bm{b}=[5,3,2]^T$~N\,m. These are design bounds on the commanded \emph{correction}, not analytically derived maximum mismatch torques or calibrated motor capacities. With joint position and velocity errors $\bm e_q$, $\dot{\bm e}_q$, normalized residual $\bm u_N=\bm\tau_{\rm RL}\oslash\bm b$, and normalized action increment $\Delta\bm u_N$, the implemented per-step reward was
\begin{equation}\label{eq:revision_reward}
 r_k=-\left[2\bm e_q^T\!\operatorname{diag}(25,20,25)\bm e_q
 +0.35\dot{\bm e}_q^T\!\operatorname{diag}(6,5,6)\dot{\bm e}_q
 +0.05\|\bm u_N\|_2^2+0.20\|\Delta\bm u_N\|_2^2\right].
\end{equation}
The original reward has no direct negative-tension, cable-allocation-capacity or hard joint-limit term. A fixed-policy authority ablation scaled the frozen output by gains $\{0.50,0.75,1.00\}$ without retraining. A separate native-bound experiment trained one DDPG initialization each at $0.5\times$, $1.0\times$ and $1.5\times$ the original action limits for 60 episodes; checkpoint screening used W1--W2 and subsequent evaluation W7--W10. These bound experiments must not be combined with the five-seed algorithm benchmark.

For a reward-weight sensitivity study, one coefficient at a time was varied around $(w_e,w_{\dot e},w_u,w_{\Delta u})=(2,0.35,0.05,0.20)$. The tested levels were $w_e\in\{1,2,4\}$, $w_{\dot e}\in\{0.175,0.35,0.70\}$, $w_u\in\{0.025,0.05,0.10\}$ and $w_{\Delta u}\in\{0.10,0.20,0.40\}$. Reusing the shared baseline gave nine distinct 60-episode runs, \emph{not} an 81-configuration factorial experiment. For each run, the best within-run reward checkpoint, episode 30 and episode 60 were screened on two fixed disturbances. A pair was eligible only if both tests completed without joint-limit crossings; early-stopped runs were counted as failures, not assigned 10~s RMSE values. Pareto comparisons among eligible checkpoints used tracking RMSE, minimum sampled joint clearance and static torque-allocation mismatch. Raw training returns under different reward definitions were not treated as directly comparable.

A distinct single-seed ($20261030$), two-arm reward ablation studied a bounded-allocation proxy. It penalized the normalized \emph{pre-lag torque-allocation discrepancy}, $p_k=\|\operatorname{diag}(5,3,2)^{-1}(\bm\tau_{{\rm req},k}-\Jc^T\F_k^*)\|_2^2$, with $r_k^{\rm aug}=r_k-\lambda p_{k-1}$ for $\lambda\in\{0,30\}$. The one-step delay avoided an algebraic loop. Both agents were newly trained for 80 episodes; neither was the submitted \texttt{Agent20}. This is \emph{not} a penalty on negative raw pseudoinverse tensions and does not enforce all joint limits.
\end{revisionblock}

\subsection{Bounded numerical cable-force allocation and actuator response}
\begin{revisionblock}
The archived cable-demand mapping was principally presented as algebraic diagnostics. To assess its limitations, a \emph{separate} numerical closed-loop model was tested with force-based applied generalized torque. At each sample, the requested torque was mapped to bounded cable forces by
\begin{equation}\label{eq:bounded_allocator}
 \F^*=\arg\min_{\F}\|\Jc(\q)^T\F-\taucmd\|_2^2,
 \qquad 0.1\,{\rm N}\leq F_i\leq F_{\max},
\end{equation}
where assumed per-cable ceilings were $F_{\max}\in\{50,60,70,80,100\}$~N. The simulation used $\bm\tau_{\rm applied}=\Jc(\q)^T\F^*$; selected solver states were compared with MATLAB \texttt{quadprog}. A separate idealized first-order response obeyed $T_a\dot{\F}_{\rm actual}=\F^*-\F_{\rm actual}$, initialized at 0.1~N. Time constants of 30, 100 and 300~ms and an intermediate onset sweep were evaluated. The $0.1$-N lower bound is a numerical floor, \emph{not} evidence of regulated physical pretension. With three cables and three generalized coordinates, a nonsingular square cable Jacobian does not provide the free null space needed for arbitrary independent pretension. These models omit identified cable elasticity, pulley and spool inertia, motor drive electronics, transmission friction and force-sensor dynamics.

The authors additionally confirmed the original Simulink model's numerical force reconstruction and torque-backprojection path, including model-specific fallback handling. This verification establishes the reported model routing, not the physical validity of cable actuation, and does not make the supplemental force-bounded simulations equivalent to the original study. Consequently the representative original results and supplemental constrained-force tests remain labeled separately throughout.
\end{revisionblock}

\subsection{Observer, feedback-family and trajectory comparisons}
\begin{revisionblock}
For a representative disturbance-observer comparator, a reduced-order velocity-based extended-state observer (ESO) supplemented the same nominal CTC. Its per-channel nominal-torque correction was limited to $\pm[1,0.6,0.4]^T$~N\,m with observer bandwidth $\omega=8$~rad/s and sample time 0.01~s. An earlier pilot using the original DDPG's larger $\pm[5,3,2]^T$~N\,m bounds crossed a sampled joint limit; the smaller ESO limits were calibrated on the gait-like trajectory and subsequently held fixed. Thus the comparison tests \emph{unequal correction authority}, not an equal-budget algorithm ranking. Reported applied-torque effort is the simulated generalized torque after force allocation and assumed lag, not a motor power measurement.

Five additional controller families were adapted to supply bounded joint-torque corrections around the same CTC baseline: robust sigmoid PI--D, finite-memory softsign fractional-order PID, Gudermannian PID, CMAC-FOPID and BELBIC-PID. Their gains were selected using gait-like simulation configurations, then tested on an ellipse. These adaptations do not reproduce all original-controller plants or their theoretical guarantees; tuning effort and correction authority were not identically matched. Two other proposed families (robust fractional-order master--slave and multi-node neuroendocrine PID) were not implemented. The experiment is a limited two-reference comparison, not a universal algorithm benchmark.

The unchanged \texttt{Agent20} policy was tested on gait-like, elliptical, smooth point-to-point and one fixed stochastic-phase reference using a separate +10\% mass/inertia setting and replayed 5--9~s disturbance. A single-seed 80-episode warm-start experiment alternated all four references, with checkpoints at episodes 20, 40, 60 and 80; early-terminated evaluations were not assigned full-horizon RMSE. Four reference types do not mean four independent training seeds or represent a patient-motion distribution.
\end{revisionblock}

\subsection{Individual uncertainty, disturbance, loading and joint-limit protocols}
\begin{revisionblock}
One-factor mechanical mismatch tests scaled the three link masses, inertia parameters or link lengths together within one selected family by $\pm5$, $\pm10$ or $\pm20\%$, yielding 18 conditions per controller. The other parameter families, original frozen policy and reference generation were held fixed. Changing physical lengths without refitting the reference or initial inverse kinematics changes reachability as well as dynamics. Ten synthetic filtered-Gaussian disturbance shapes were tested at common RMS and at $1.5\times$ that RMS, always paired between CTC and residual DDPG. Distal point loads of 0, 0.5 and 1~kg were also tested. Reported conditional mean improvement uses only matched, fully completed 10~s pairs; terminations are shown separately.

Joint-limit violations were counted as joint-channel samples outside $\q_{\min}=[80,2,250]^T$ degrees or $\q_{\max}=[250,160,330]^T$ degrees at 0.01~s logging intervals; this is neither the count of continuous-time events nor a formal safety certificate. In the alternative circular-reference replay, an inference-time attenuation of the \emph{frozen} residual policy used $m_2=\min(q_2-2^\circ,160^\circ-q_2)$ and $g=\operatorname{clip}((m_2-4^\circ)/4^\circ,0,1)$, multiplying all three residual channels by $g$. The attenuation is a joint-2 heuristic and does not guarantee other joints or cable feasibility. Historical 44/101 counts and matched replay 47/102/47 counts use distinct logs and are never mixed when computing the gate's percentage effect.
\end{revisionblock}

\FloatBarrier

%% file: sections/06_results_and_discussion.tex
\section{Results and discussion}

\begin{revisionblock}
The archived performance metrics describe numerical controller/model outputs whose exact historical plant-input wiring is not fully attributable to a verified executable version. The figures distinguish logged requested generalized torque and algebraically inferred forces; neither is interpreted as measured physical actuation. Separate force-allocation experiments are presented in Section~\ref{sec:revision_results} and are not pooled with these archived results.
\end{revisionblock}
The performance metrics characterize the evaluated simulations; cable demand and geometric feasibility are assessed using the stated algebraic and sampled criteria. The results are presented in increasing order of controller dependence. The sampled workspace and cable-Jacobian checks first establish whether the selected reference is compatible with the modeled geometry. The representative combined case then provides detailed tracking, joint-limit, and algebraic cable-demand metrics. Four-case, seed, interval, initial-condition, residual-action, and alternate-trajectory analyses are subsequently used to determine which conclusions are repeatable and which remain trajectory- or dataset-specific. Independently generated result files are not merged when their constraint logs disagree.

\subsection{Sampled workspace, static tension feasibility, and cable-Jacobian conditioning}

The joint-limit domain in Eq.~\eqref{eq:limits} was sampled at 45 values per joint, giving $45^3=91{,}125$ configurations. For each sample, forward kinematics produced a geometric workspace point. The static tension-feasibility test then used the nominal gravity-compensation torque and the square mapping in Eq.~\eqref{eq:pinv}; a sample was counted as feasible when $\Jc$ was full rank, the reconstructed torque residual was negligible, and every inferred tension was nonnegative. This criterion directly evaluates pull-only static feasibility for the modeled cable routing.

\begin{figure}[!htbp]
    \centering
    \includegraphics[width=0.84\textwidth]{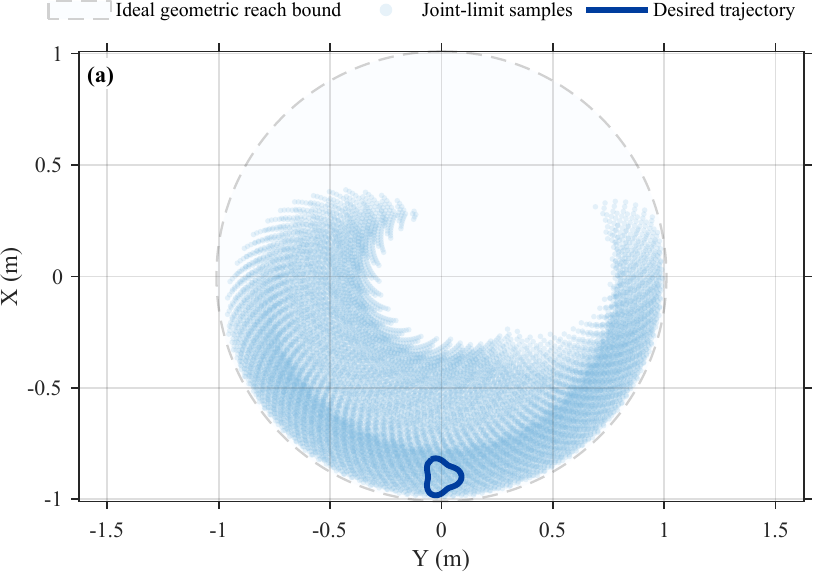}
    \caption{Sampled joint-limit workspace and desired Cartesian reference path for the planar 3-DOF mechanism. The light background denotes the ideal geometric reach bound, the sampled points denote configurations satisfying the active joint limits, and the high-contrast curve denotes the desired path; the path is closed for display by repeating its first plotted Cartesian point without modifying the evaluated trajectory data. The plotting orientation is horizontal $Y$ and vertical $X$, with 45 samples per joint.}
    \label{fig:workspace_joint_limits}
\end{figure}

\begin{figure}[!htbp]
    \centering
    \includegraphics[width=0.84\textwidth]{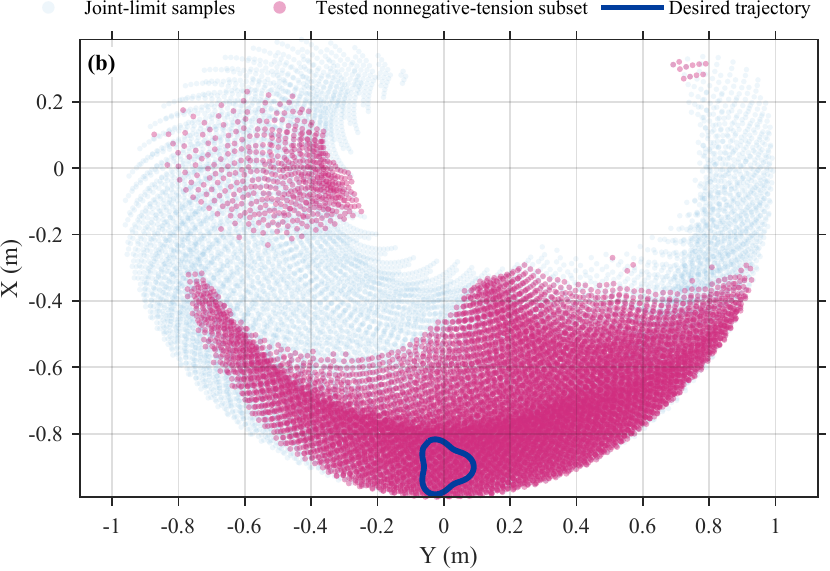}
    \caption{Sampled cable-feasibility map for the planar 3-DOF, three-cable mechanism. Light points denote joint-limited workspace samples, the emphasized subset satisfies the nominal static gravity-torque mapping with full-rank $\Jc$ and nonnegative inferred cable tension, and the high-contrast curve denotes the desired Cartesian path. The path is closed for display by repeating its first plotted point without modifying saved trajectory data or performance calculations; axes are horizontal $Y$ and vertical $X$.}
    \label{fig:workspace_tension}
\end{figure}

Figures~\ref{fig:workspace_joint_limits} and~\ref{fig:workspace_tension} show the geometric workspace and the statically tested nonnegative-tension subset. Of 91,125 samples, 33,921 satisfy the stated test, corresponding to 37.22\%. The desired path is contained in both sampled sets, supporting the selected trajectory and cable routing throughout the investigated operating region under the stated static criterion.

\begin{figure}[!htbp]
    \centering
    \includegraphics[width=0.86\textwidth]{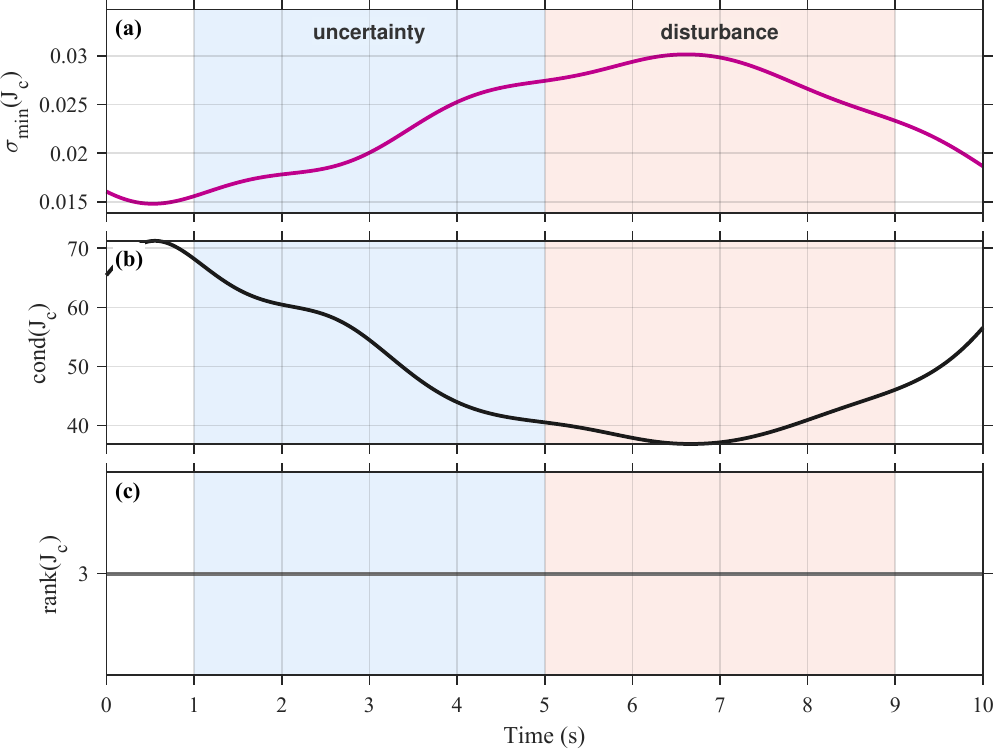}
    \caption{Cable-actuation Jacobian diagnostics along the desired trajectory: minimum singular value $\sigma_{\min}(\Jc)$, two-norm condition number $\kappa_2(\Jc)$, and matrix rank. Rank remains three over the logged path, while $\sigma_{\min}$ and $\kappa_2$ identify the least favorably conditioned intervals. Light-blue and light-peach shading denote the 1--5 s uncertainty and 5--9 s disturbance intervals, respectively.}
    \label{fig:workspace_jacobian}
\end{figure}

Figure~\ref{fig:workspace_jacobian} evaluates the mapping along the desired trajectory. Rank remains three, the minimum singular value is 0.01481, and the maximum two-norm condition number is 71.25. Full rank avoids an exact singularity in the logged path, but the condition-number peak indicates non-negligible numerical sensitivity: perturbations in geometry or torque can be amplified in the inferred tension near the least favorable posture. These values justify monitoring conditioning rather than relying only on determinant or rank. Table~\ref{tab:workspace} consolidates the sampled-workspace and trajectory-conditioning results used in this interpretation.

\begin{table}[!htbp]
\centering
\caption{Sampled workspace and cable-Jacobian summary for the desired trajectory.}
\label{tab:workspace}
\begin{tabular}{lc}
\toprule
Quantity & Value \\
\midrule
Samples per joint / total samples & 45 / 91,125 \\
Statically nonnegative-tension samples & 33,921 \\
Nonnegative-tension sample ratio & 37.22\% \\
Desired path inside geometric sample & 100\% \\
Desired path inside tested feasible sample & 100\% \\
Minimum $\sigma_{\min}(\Jc)$ along path & 0.01481 \\
Maximum $\kappa_2(\Jc)$ along path & 71.25 \\
Minimum rank of $\Jc$ along path & 3 \\
Minimum $|\det(\Jc)|$ along path & $1.7582\times10^{-3}$ \\
\bottomrule
\end{tabular}
\end{table}

\subsection{Representative combined uncertainty/disturbance case}

\begin{figure}[!htbp]
    \centering
    \begin{subfigure}[c]{0.49\textwidth}
        \centering
        \includegraphics[width=\textwidth]{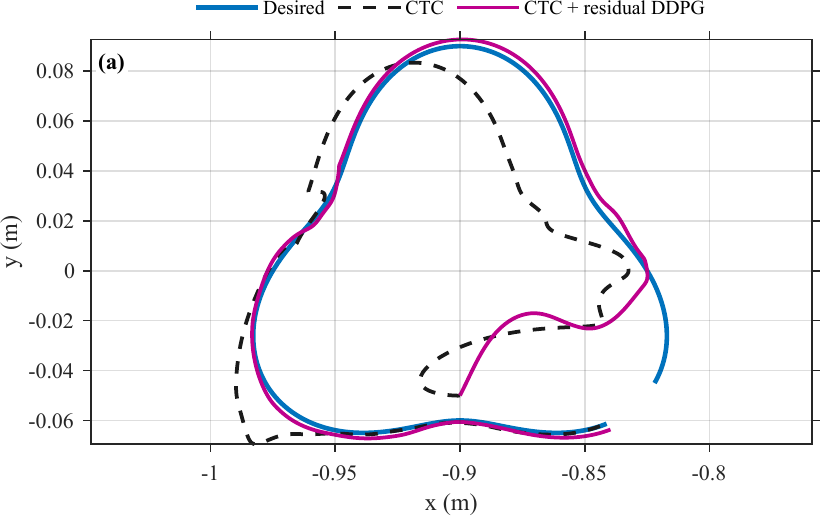}
        \caption{Cartesian path.}
    \end{subfigure}
    \hfill
    \begin{subfigure}[c]{0.49\textwidth}
        \centering
        \includegraphics[width=\textwidth]{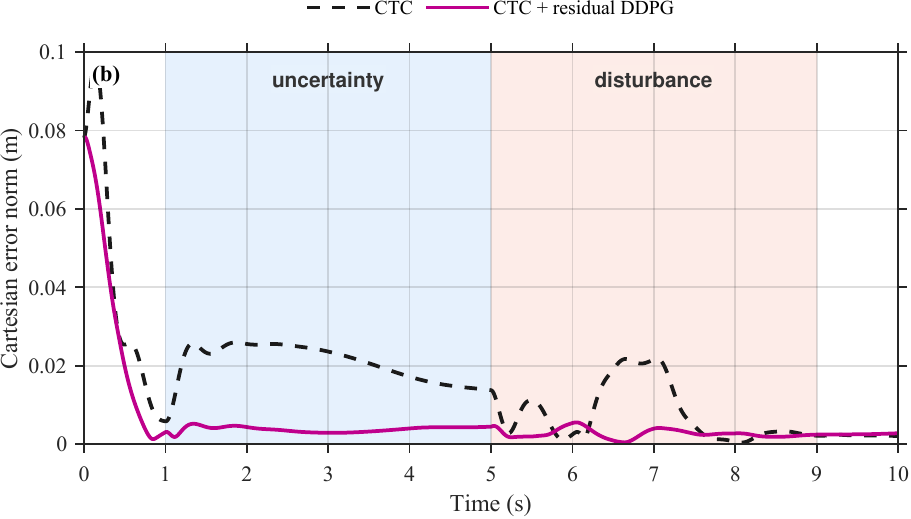}
        \caption{Cartesian error norm.}
    \end{subfigure}
    \caption{Cartesian tracking under combined parametric uncertainty and external disturbance. Panel (a) compares the desired path with the realized CTC and CTC + bounded residual DDPG trajectories; panel (b) reports the Euclidean Cartesian tracking-error norm. Light-blue shading marks the 1--5 s uncertainty interval and light-peach shading marks the 5--9 s disturbance interval.}
    \label{fig:combined_results}
\end{figure}

The representative stress test applies simultaneous 10\% plant-parameter increases from 1 to 5 s, followed by band-limited disturbance torque from 5 to 9 s. Figure~\ref{fig:combined_results} asks whether the residual policy reduces Cartesian deviation when the nominal inverse-dynamics cancellation is deliberately violated. Both controllers follow the closed path, but the CTC error rises more strongly during the non-nominal intervals. The residual branch maintains a visibly smaller error envelope, consistent with its role as a learned compensation signal rather than a trajectory generator.

Table~\ref{tab:main_results} is generated from the representative post-processing dataset. RMS Cartesian error decreases from 0.0221419 m to 0.0128462 m, a 41.9826\% reduction, reported as approximately 42\%. Peak error decreases by 19.4\%, IAE by 63.5\%, and ISE by 66.8\%. The substantially larger IAE and ISE reductions indicate that the residual policy reduces sustained error over the full run, not only an isolated peak. The peak values are identical across several independently generated cases because the largest excursion occurs in the shared initial transient; peak error is therefore less discriminating than interval RMS, IAE, or ISE for disturbance rejection in this dataset.

\begin{table}[!htbp]
\centering
\caption{\revcaption{Representative archived combined-case results from the detailed post-processing dataset; the original executable actuation-path attribution remains under verification.}}
\label{tab:main_results}
\resizebox{\linewidth}{!}{%
\begin{tabular}{lccc}
\toprule
Metric & CTC & CTC + residual DDPG & Change \\
\midrule
RMS Cartesian error (m) & 0.0221419 & 0.0128462 & 42.0\% lower \\
Peak Cartesian error (m) & 0.0970994 & 0.0782787 & 19.4\% lower \\
IAE (m s) & 0.156619 & 0.0572133 & 63.5\% lower \\
ISE (m$^2$ s) & 0.00487701 & 0.00162135 & 66.8\% lower \\
Recovery time after disturbance onset (s) & 0.08 & 0 & -- \\
Minimum logged cable-demand output (N) & 0 & 3.41445 & -- \\
Maximum logged cable-demand output (N) & 76.8738 & 86.2373 & 12.2\% higher \\
Negative logged allocation values & 0$^{\ast}$ & 0$^{\ast}$ & -- \\
Joint-limit channel-samples & 61 & 0 & -- \\
Joint-limit violation duration (s) & 0.61 & 0 & -- \\
\midrule
\multicolumn{4}{l}{\footnotesize $^{\ast}$The allocation output is nonnegative by construction; see Section~\ref{sec:allocation_scope}.}\\
\bottomrule
\end{tabular}
}
\end{table}

\begin{figure}[!htbp]
    \centering
    \includegraphics[width=0.86\textwidth]{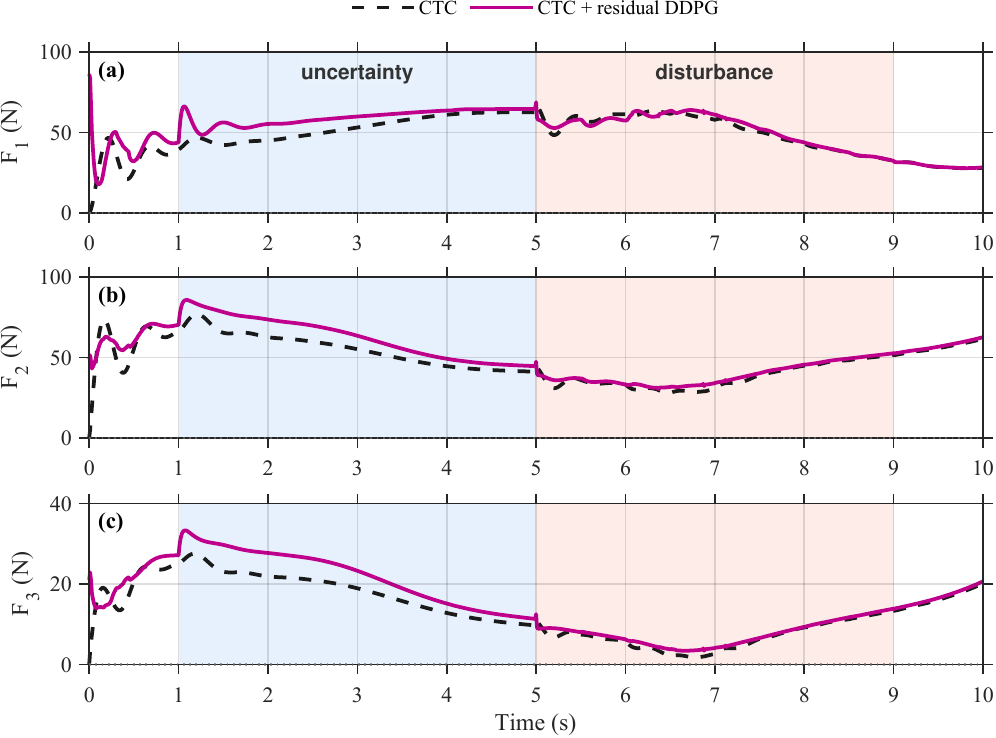}
    \caption{Algebraically inferred cable-tension demands $F_1$, $F_2$, and $F_3$ in the representative combined case. Dashed curves correspond to CTC and solid curves to CTC + bounded residual DDPG. Light-blue and light-peach shading identify the uncertainty and disturbance intervals, respectively.}
    \label{fig:cable_tensions_main}
\end{figure}

\begin{figure}[!htbp]
    \centering
    \includegraphics[width=0.82\textwidth]{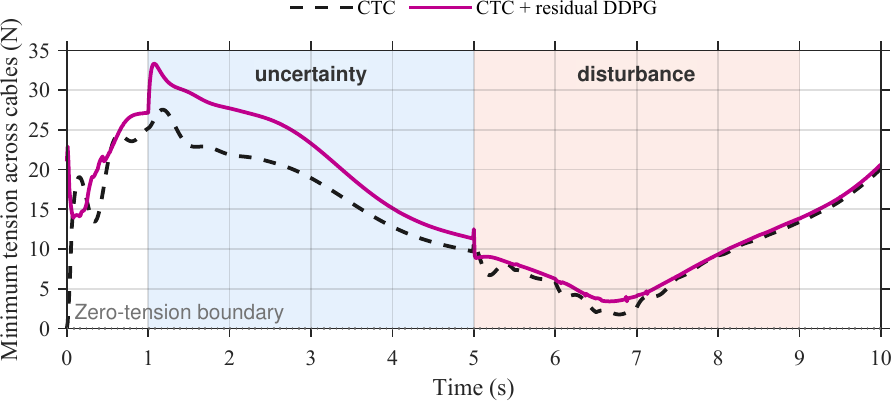}
    \caption{\revcaption{Minimum logged algebraic cable-demand audit output, $\min_i F_i(t)$, in the representative combined case. The audit logic returns zero when its nonnegative-tension test rejects a solution, so the logged signal is nonnegative by construction. CTC reaches this zero fallback boundary, whereas the CTC + bounded residual DDPG branch retains a minimum of 3.41445~N. This logged audit output is not a measured physical cable force and does not independently establish the complete historical plant-input path. Light-blue and light-peach shading denote the uncertainty and disturbance intervals.}}
    \label{fig:tension_constraint_check}
\end{figure}

\begin{figure}[!htbp]
    \centering
    \includegraphics[width=0.84\textwidth]{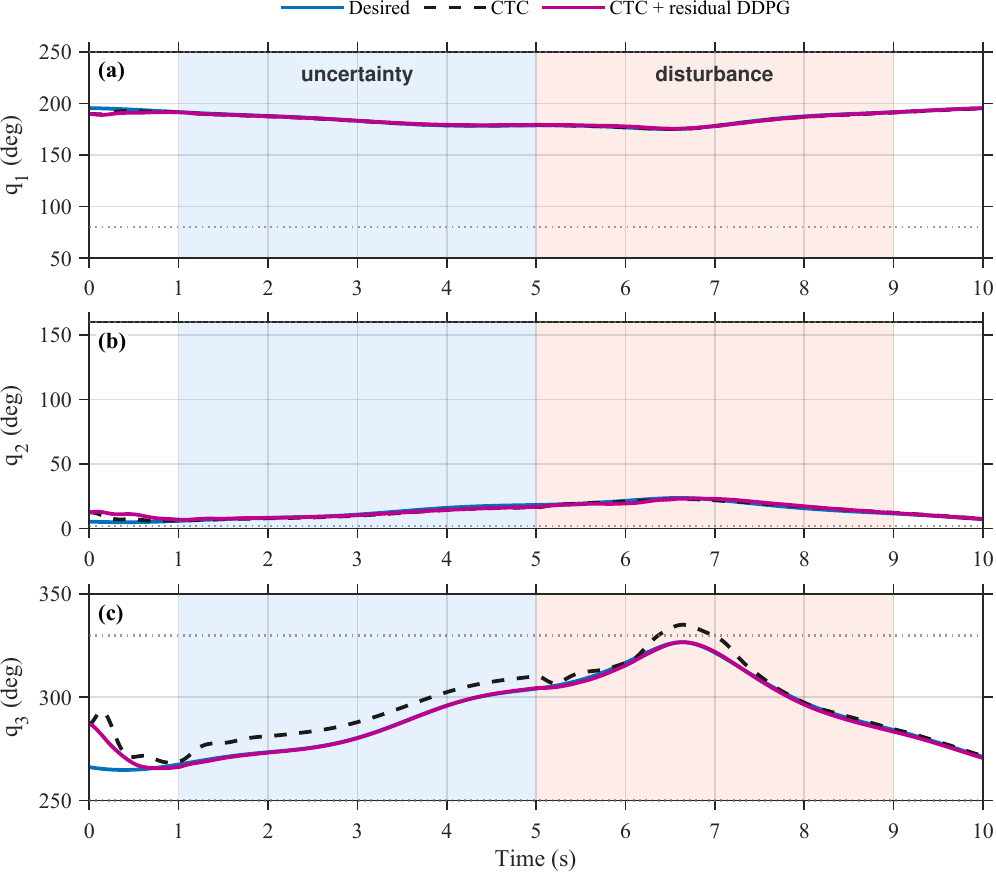}
    \caption{Joint-coordinate responses $q_1$, $q_2$, and $q_3$ with their active lower and upper limits in the representative combined case. The CTC trajectory contains 61 out-of-limit channel-samples, whereas the CTC + bounded residual DDPG trajectory contains none under the same sampling and tolerance convention.}
    \label{fig:joint_limit_constraint_check}
\end{figure}

For the residual branch, the saved threshold-and-hold recovery metric remains below its excursion criterion after disturbance onset and therefore reports zero recovery delay. The count of 61 joint-limit channel-samples for CTC corresponds to 0.61 s in the representative file, while the residual branch has zero such samples. Figures~\ref{fig:cable_tensions_main} and~\ref{fig:tension_constraint_check} report the logged algebraic cable-demand audit outputs and their channel-wise minimum, and Fig.~\ref{fig:joint_limit_constraint_check} reports the joint-coordinate response relative to the active limits.

\begin{revisionblock}
The residual controller raises the maximum logged algebraic cable demand from 76.87 N to 86.24 N while reducing tracking error. This quantifies the associated tracking--demand tradeoff and motivates incorporating cable-force margin into the reward and allocation layers once physical motor, spool, cable, pretension, and $F_{max}$ specifications are defined. Because historical requested torque may pass through model-specific force backprojection and rejected allocation solutions may be replaced by a fallback, the logged demand cannot be interpreted as physically realized dynamic cable tension.
\end{revisionblock}

\subsection{Four-case comparison and dataset provenance}

Table~\ref{tab:case_matrix} compares the four primary scenarios using an independently generated case-matrix file. Residual DDPG reduces RMS error by 21.1\% in the nominal case, 39.4\% with uncertainty only, 27.7\% with disturbance only, and 41.6\% in the combined case. The larger gains in the uncertainty-containing cases support the interpretation that the actor compensates for nominal-model mismatch. The nonzero nominal improvement can arise from effects not canceled by the ideal CTC equation, including command filtering, numerical differentiation, discretization, and the policy's learned correction of the repeatable trajectory transient.

\begin{table}[!htbp]
\centering
\caption{\revcaption{Archived four-case numerical comparison from the independent case-matrix dataset; the archived algebraic force-backprojection actuation path was confirmed by the authors.}}
\label{tab:case_matrix}
\resizebox{\linewidth}{!}{%
\begin{tabular}{llcccc}
\toprule
Case & Controller & RMS (m) & Peak (m) & Min. tension (N) & RMS reduction \\
\midrule
Nominal & CTC & 0.016479 & 0.097099 & 0 & -- \\
Nominal & CTC + residual DDPG & 0.013001 & 0.078279 & 3.7196 & 21.1\% \\
Uncertainty only & CTC & 0.021239 & 0.097099 & 0 & -- \\
Uncertainty only & CTC + residual DDPG & 0.012875 & 0.078279 & 3.7197 & 39.4\% \\
Disturbance only & CTC & 0.018034 & 0.097099 & 0 & -- \\
Disturbance only & CTC + residual DDPG & 0.013034 & 0.078279 & 3.9467 & 27.7\% \\
Combined & CTC & 0.022032 & 0.097099 & 0 & -- \\
Combined & CTC + residual DDPG & 0.012863 & 0.078279 & 3.7129 & 41.6\% \\
\bottomrule
\end{tabular}
}
\end{table}

\begin{figure}[!htbp]
    \centering
    \begin{subfigure}[b]{0.48\textwidth}
        \centering
        \includegraphics[width=\textwidth]{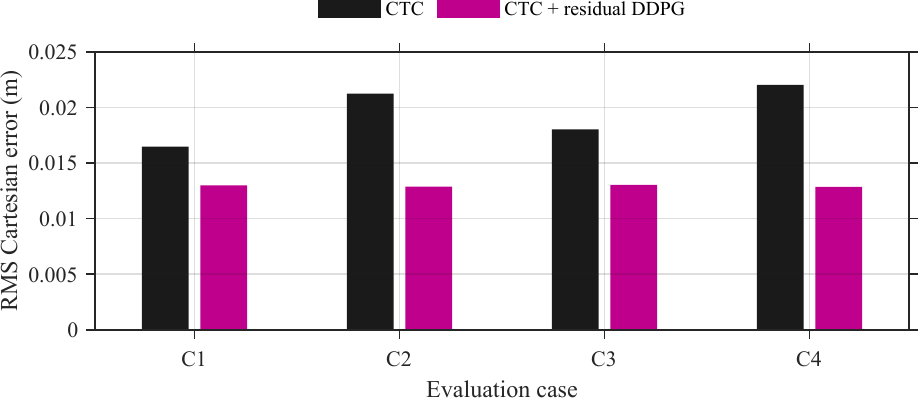}
        \caption{RMS error.}
    \end{subfigure}
    \hfill
    \begin{subfigure}[b]{0.48\textwidth}
        \centering
        \includegraphics[width=\textwidth]{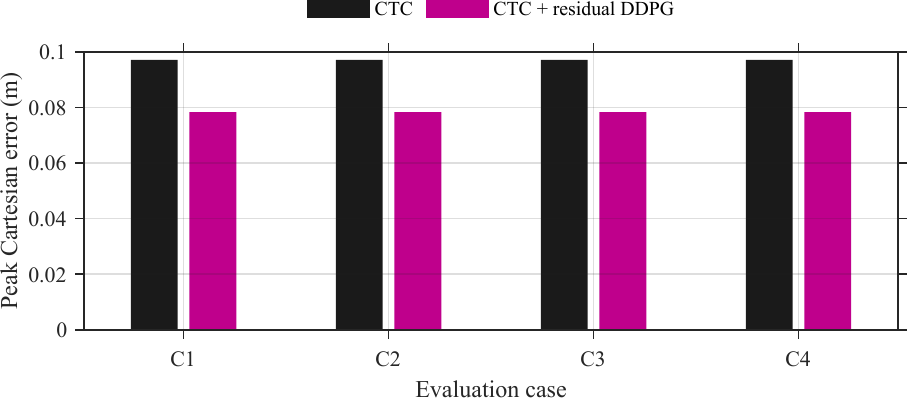}
        \caption{Peak error.}
    \end{subfigure}
    \caption{Cartesian tracking metrics for the nominal, uncertainty-only, disturbance-only, and combined scenarios: panel (a) reports RMS error and panel (b) reports peak error for CTC and CTC + bounded residual DDPG.}
    \label{fig:case_summary}
\end{figure}

\begin{figure}[!htbp]
    \centering
    \includegraphics[width=0.86\textwidth]{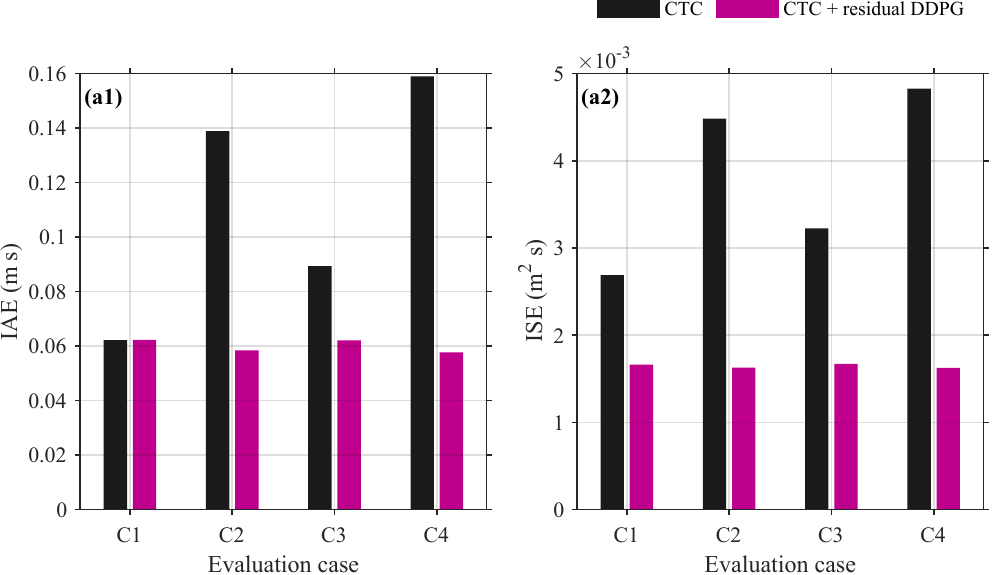}
    \caption{Integral Cartesian tracking metrics for the four scenarios. The grouped values compare IAE and ISE for CTC and CTC + bounded residual DDPG using the independently generated case-matrix dataset.}
    \label{fig:case_integral_metrics}
\end{figure}

\begin{figure}[!htbp]
    \centering
    \begin{subfigure}[c]{0.82\textwidth}
        \centering
        \includegraphics[width=0.98\textwidth]{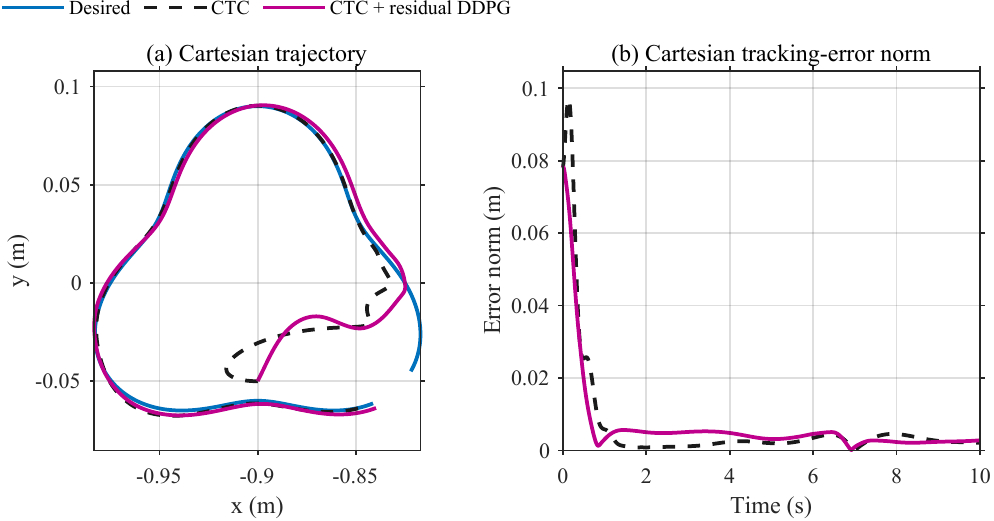}
        \caption{Nominal.}
    \end{subfigure}

    \vspace{2pt}
    \begin{subfigure}[c]{0.82\textwidth}
        \centering
        \includegraphics[width=0.98\textwidth]{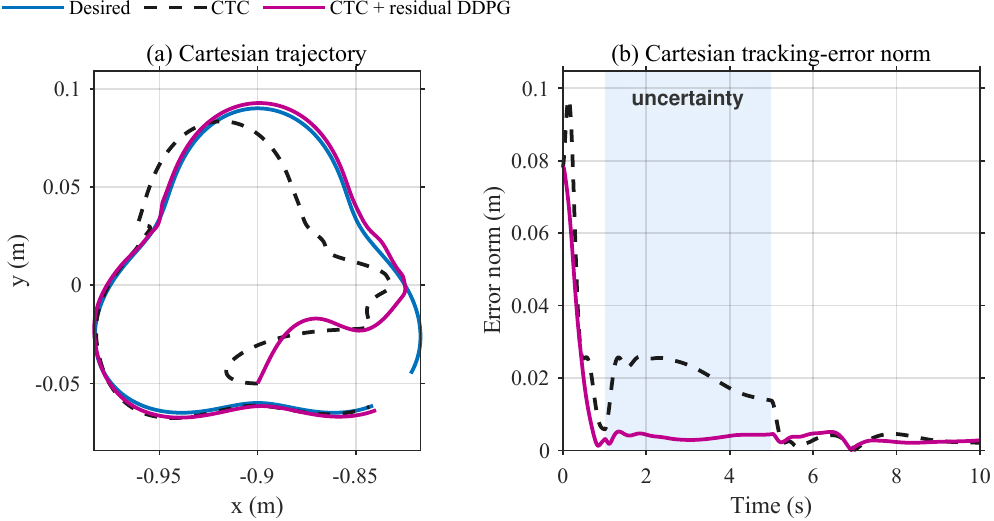}
        \caption{Uncertainty only.}
    \end{subfigure}
    \caption{Cartesian path and Euclidean tracking-error histories for C1 (nominal) and C2 (10\% parametric uncertainty active from 1 to 5 s). Each case compares CTC with CTC + bounded residual DDPG under the same desired path and initial condition.}
    \label{fig:case_timehistories_1}
\end{figure}

\begin{figure}[!t]
    \centering
    \begin{subfigure}[c]{0.82\textwidth}
        \centering
        \includegraphics[width=0.98\textwidth]{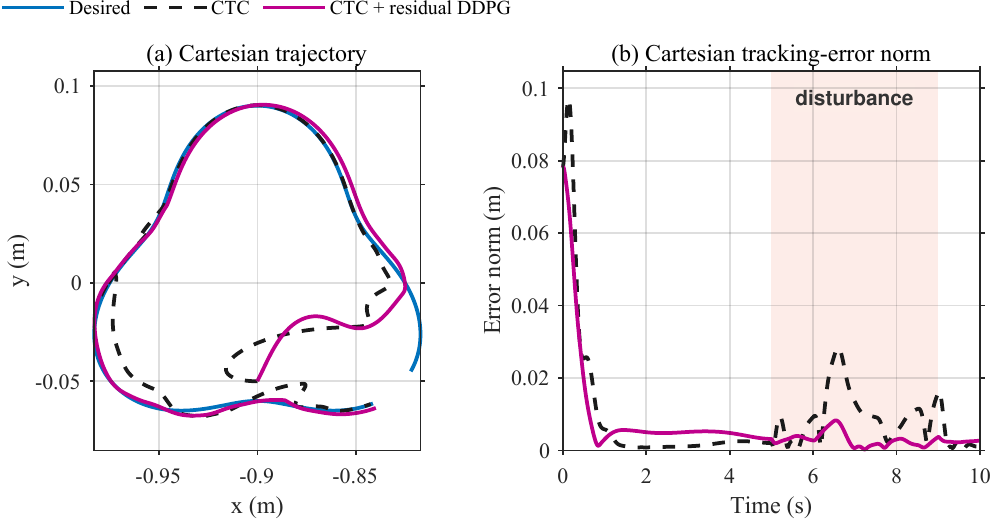}
        \caption{Disturbance only.}
    \end{subfigure}

    \vspace{2pt}
    \begin{subfigure}[c]{0.82\textwidth}
        \centering
        \includegraphics[width=0.98\textwidth]{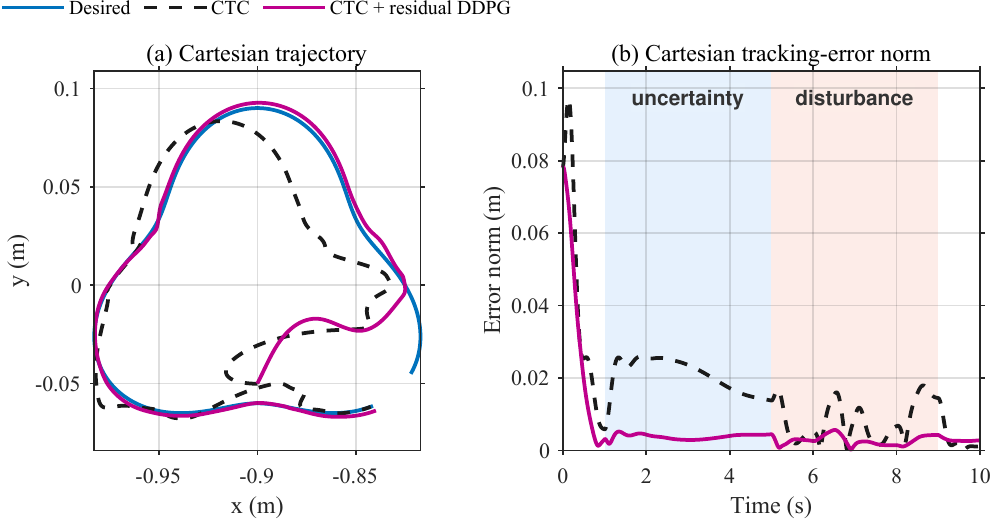}
        \caption{Combined uncertainty and disturbance.}
    \end{subfigure}
    \caption{Cartesian path and Euclidean tracking-error histories for C3 (external disturbance active from 5 to 9 s) and C4 (10\% uncertainty from 1 to 5 s followed by disturbance from 5 to 9 s). Each case compares CTC with CTC + bounded residual DDPG.}
    \label{fig:case_timehistories_2}
\end{figure}

In the nominal case, IAE changes from 0.0621909 to 0.0622811 m s, a difference of about 0.15\%, while RMS and ISE decrease. Reporting the full metric set therefore distinguishes sustained-error improvements in the non-nominal cases from this near-equal nominal IAE result. Figures~\ref{fig:case_summary} and~\ref{fig:case_integral_metrics} visualize the aggregate metrics, whereas Figs.~\ref{fig:case_timehistories_1} and~\ref{fig:case_timehistories_2} locate the changes within the nominal, uncertainty, disturbance, and combined time histories.

The combined RMS reductions differ slightly between the independently saved datasets: 41.9826\% in the representative post-processing file and 41.6161\% in the case matrix. The manuscript therefore uses ``approximately 42\%'' for the general combined tracking statement and ties exact values to their source table. Detailed joint-limit counts are reported only from the representative post-processing file, whose channel-sample accounting was used for the associated constraint figures; constraint logs from independently generated files are not pooled.

\subsection{Ten-seed robustness and interval-wise behavior}

\begin{figure}[!htbp]
    \centering
    \begin{subfigure}[b]{0.48\textwidth}
        \centering
        \includegraphics[width=\textwidth]{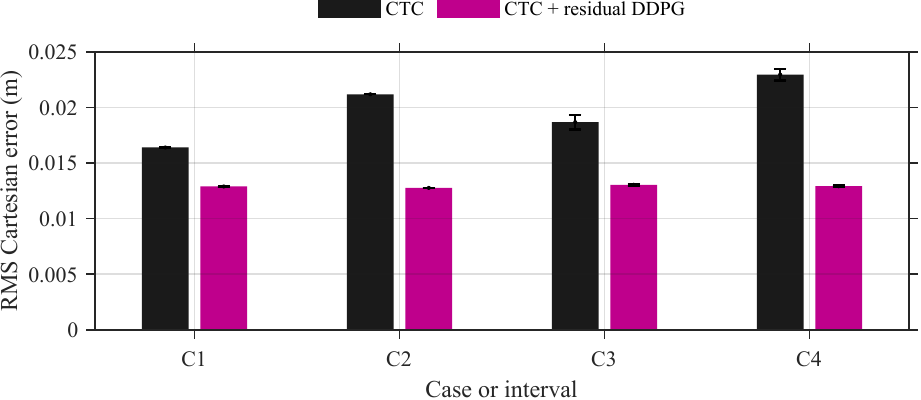}
        \caption{Seed-level RMS error.}
    \end{subfigure}
    \hfill
    \begin{subfigure}[b]{0.48\textwidth}
        \centering
        \includegraphics[width=\textwidth]{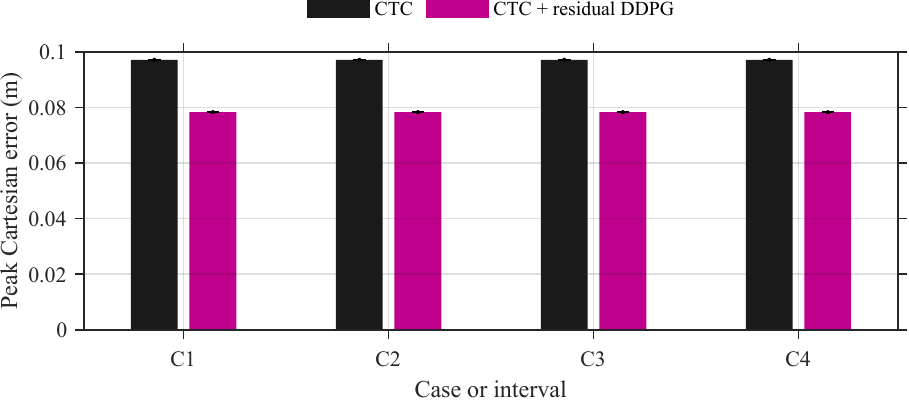}
        \caption{Seed-level peak error.}
    \end{subfigure}
    \caption{Ten-seed Cartesian-error summary for the four scenarios. Bars report the mean and error bars report one standard deviation for panel (a) RMS error and panel (b) peak error. Seed-dependent variability appears in the disturbance-enabled scenarios; the nominal and uncertainty-only scenarios are deterministic in this evaluation.}
    \label{fig:robustness_seed}
\end{figure}

Disturbance-enabled cases were repeated with ten seeds. For the combined case, mean RMS error is $0.022945\pm0.000533$ m for CTC and $0.012916\pm0.000082$ m for CTC + residual DDPG, giving a mean reduction of $43.69\pm1.17$\%. Figure~\ref{fig:robustness_seed} summarizes the seed-level RMS and peak statistics. The smaller residual-controller standard deviation indicates reduced sensitivity across the tested disturbance realizations.

\begin{figure}[!htbp]
    \centering
    \includegraphics[width=0.86\textwidth]{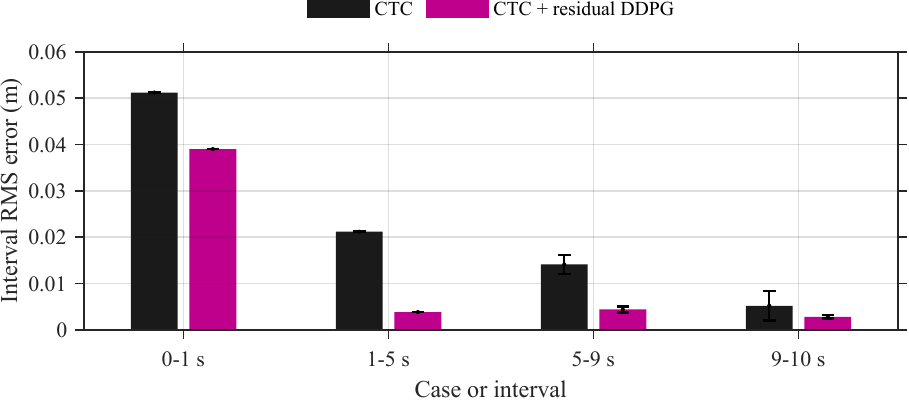}
    \caption{Interval-wise Cartesian RMS error averaged over ten seed runs for 0--1 s (initial transient), 1--5 s (parametric uncertainty), 5--9 s (external disturbance), and 9--10 s (post-disturbance recovery).}
    \label{fig:robustness_interval}
\end{figure}

The interval analysis in Fig.~\ref{fig:robustness_interval} provides stronger temporal interpretation than the full-run peak. Mean RMS reduction is 23.82\% during the initial transient, 81.84\% during the deterministic uncertainty interval, 68.45\% during the disturbance interval, and 22.64\% during recovery. Recovery variability is comparatively large because the CTC recovery error depends strongly on the preceding random disturbance. The largest and most consistent gains occur while model mismatch or disturbance is active, which is aligned with the residual-compensation design.

\subsection{Sensitivity to nearby initial configurations}
\label{sec:initial_condition_sensitivity}

\begin{figure}[!htbp]
    \centering
    \includegraphics[width=0.65\textwidth]{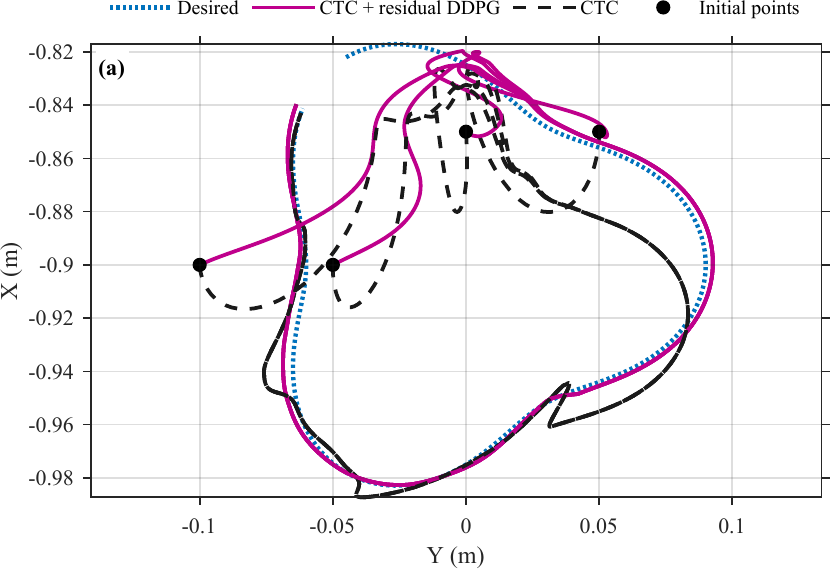}
    \caption{Cartesian trajectory-transfer evaluation from four nearby initial configurations under the combined uncertainty/disturbance scenario. The dotted curve is the desired trajectory, solid magenta curves are the CTC + bounded residual DDPG responses, dashed black curves are the CTC responses, and filled markers denote the four initial Cartesian positions. The plotting orientation is horizontal $Y$ and vertical $X$.}
    \label{fig:multi_initial_condition}
\end{figure}

Four nearby Cartesian initial conditions were evaluated without retraining the actor, with all other settings unchanged. Table~\ref{tab:initial_condition_scenarios} and Fig.~\ref{fig:multi_initial_condition} show convergence toward the planned path from each local offset, providing a focused sensitivity assessment around the nominal initial configuration.

\begin{table}[!htbp]
\centering
\caption{Cartesian initial conditions used in the local sensitivity test.}
\label{tab:initial_condition_scenarios}
\begin{tabular}{cccc}
\toprule
Case & $x_0$ (m) & $y_0$ (m) & Role \\
\midrule
IC1 & $-0.85$ & $0.05$ & Upper-side offset \\
IC2 & $-0.85$ & $0.00$ & Mild offset \\
IC3 & $-0.90$ & $-0.05$ & Nominal main-run initial state \\
IC4 & $-0.90$ & $-0.10$ & Lower-side offset \\
\bottomrule
\end{tabular}
\end{table}

\subsection{Residual authority and alternate-trajectory diagnostic}
\label{sec:circular}

\begin{figure}[!htbp]
    \centering
    \includegraphics[width=0.90\textwidth]{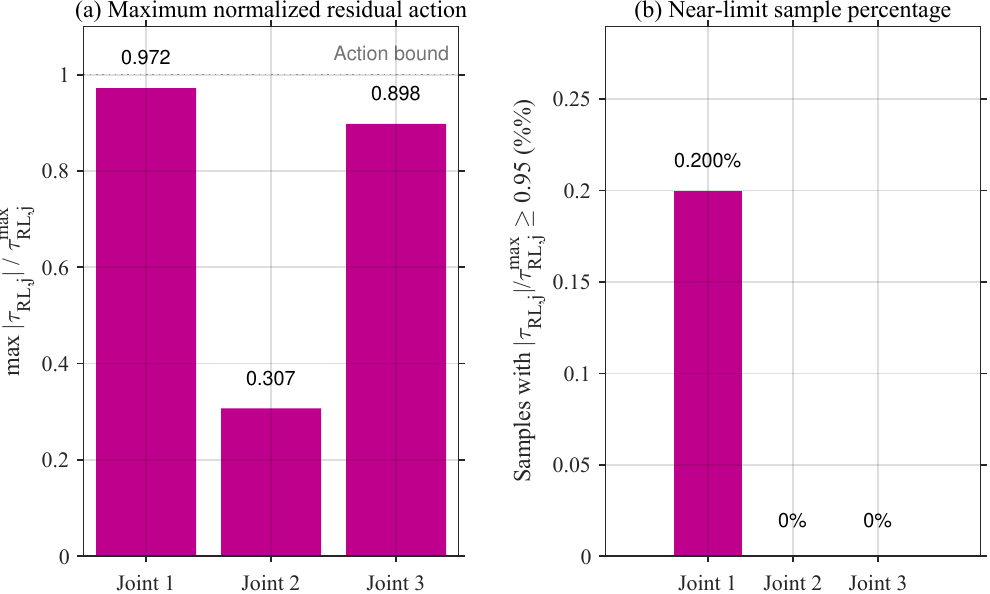}
    \caption{Residual-action usage relative to the joint-wise limits. Panel (a) reports $\rho_j^{\max}=\max_t|\tau_{\mathrm{RL},j}|/\tau_{\mathrm{RL},j}^{\max}$. Panel (b) reports the percentage of samples satisfying $|\tau_{\mathrm{RL},j}|/\tau_{\mathrm{RL},j}^{\max}\geq0.95$. Joint 1 reaches 0.972 and is near the bound for 0.200\% of samples; joints 2 and 3 have no near-limit samples.}
    \label{fig:residual_action_usage}
\end{figure}

The residual-action audit determines whether the actor behaves as a bounded correction or frequently approaches its allowed authority. For joint $j$, the peak normalized action is defined as
\begin{equation}
    \rho_j^{\max}=\max_t\frac{|\tau_{\mathrm{RL},j}(t)|}{\tau_{\mathrm{RL},j}^{\max}},
\end{equation}
and a saved sample is classified as near-limit when $|\tau_{\mathrm{RL},j}|/\tau_{\mathrm{RL},j}^{\max}\geq0.95$. Maximum absolute residual torques are 4.8598, 0.92096, and 1.7953 N m for joints 1--3, corresponding to normalized peaks of 0.972, 0.307, and 0.898. Only joint 1 meets the near-limit criterion, for 0.02 s or 0.200\% of the trajectory; joints 2 and 3 have 0\% near-limit samples. Residual RMS torque is 11.85\%, 7.79\%, and 20.79\% of the corresponding CTC RMS torque. Figure~\ref{fig:residual_action_usage} therefore supports the intended compensation role while identifying the first-joint peak for additional action-margin and force-demand analysis.

\begin{figure}[!htbp]
    \centering
    \includegraphics[width=0.90\textwidth]{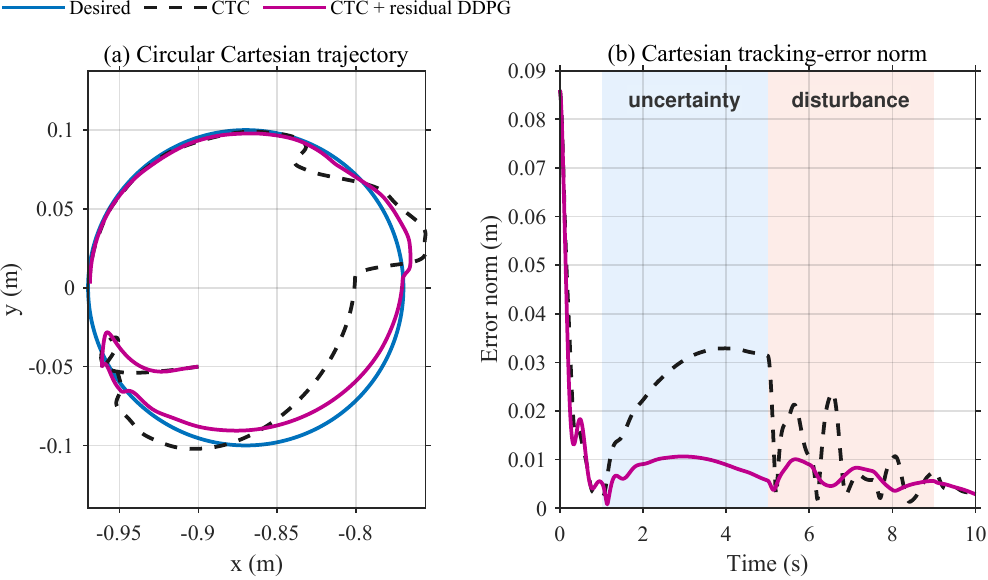}
    \caption{Circular-trajectory transfer evaluation under the combined uncertainty/disturbance scenario. Panel (a) compares the desired Cartesian path with the realized CTC and CTC + bounded residual DDPG trajectories; panel (b) reports the Euclidean Cartesian tracking-error norm. Light-blue and light-peach shading denote the 1--5 s uncertainty and 5--9 s disturbance intervals. RMS error decreases from 0.021738 to 0.012052 m; the associated source dataset contains 44 and 101 joint-limit channel-samples, respectively, identifying trajectory-dependent joint-coordinate coverage for subsequent constraint-aware training.}
    \label{fig:circular_generalization}
\end{figure}

The alternate circular path evaluates trajectory transfer and the associated joint-coordinate coverage. Residual DDPG reduces RMS Cartesian error from 0.021738 to 0.012052 m (a trajectory-specific 44.56\% reduction), while joint-limit channel-samples are 44 for CTC and 101 for the residual branch. Minimum logged tension is zero for both controllers, and maximum inferred tension is 230.42 N for CTC and 162.58 N for the residual branch. Figure~\ref{fig:circular_generalization} demonstrates transfer of Cartesian tracking performance and identifies the expanded joint-coordinate range that should be incorporated into subsequent constraint-aware training.

\begin{figure}[!htbp]
    \centering
    \begin{subfigure}[c]{0.485\textwidth}
        \centering
        \includegraphics[width=\textwidth]{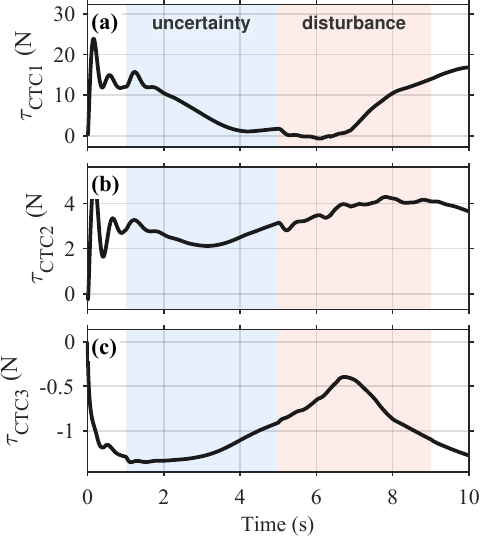}
        \caption{Nominal computed-torque component.}
    \end{subfigure}
    \hfill
    \begin{subfigure}[c]{0.485\textwidth}
        \centering
        \includegraphics[width=\textwidth]{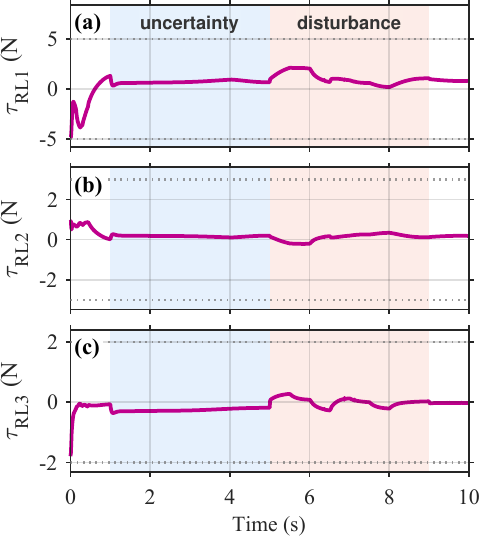}
        \caption{Bounded residual DDPG component.}
    \end{subfigure}
    \caption{Joint-torque components in the representative combined case. Panel (a) shows the nominal computed-torque command $\tau_{\mathrm{CTC},j}$ and panel (b) shows the bounded residual action $\tau_{\mathrm{RL},j}$ for joints $j=1,2,3$.}
    \label{fig:torque_components}
\end{figure}

\begin{figure}[!htbp]
    \centering
    \begin{subfigure}[c]{0.485\textwidth}
        \centering
        \includegraphics[width=\textwidth]{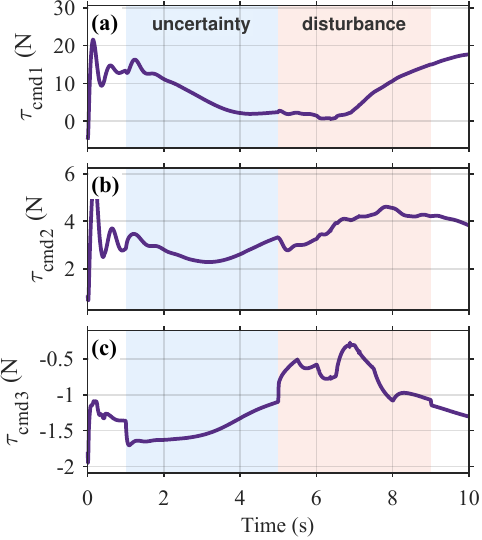}
        \caption{Combined command $\taucmd$.}
    \end{subfigure}
    \hfill
    \begin{subfigure}[c]{0.485\textwidth}
        \centering
        \includegraphics[width=\textwidth]{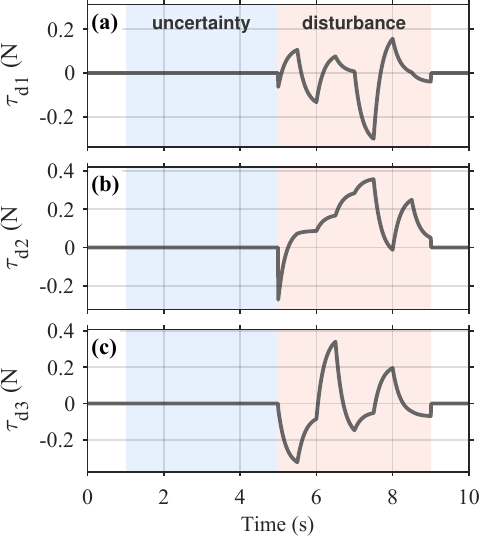}
        \caption{External disturbance $\taud$.}
    \end{subfigure}
    \caption{Controller command and plant disturbance in the representative combined case. Panel (a) reports the combined commanded joint torque $\taucmd=\tauctc+\taurl$, and panel (b) reports the externally applied disturbance torque $\taud$. The disturbance enters the simulated plant separately from the cable-allocation command.}
    \label{fig:torque_total_disturbance}
\end{figure}

\begin{revisionblock}
Figures~\ref{fig:torque_components} and~\ref{fig:torque_total_disturbance} separate the nominal model-based request, learned residual request, their combined requested torque $\taucmd=\tauctc+\taurl$, and externally injected disturbance. These logs describe \emph{requested} generalized torque; they cannot establish equality with applied generalized plant torque at each sample, because numerical force reconstruction and fallback can modify the applied input. The confirmed archived model includes algebraic force backprojection on the plant-input path; a different later bounded allocator is assessed separately. The disturbance $\taud$ is not part of the logged controller-request sum.
\end{revisionblock}

\FloatBarrier

\subsection{Overall interpretation, relation to prior work, and scope}

Across the nominal and non-nominal scenarios, the results exhibit the intended residual-control structure: CTC supplies the dominant model-based command, while the bounded DDPG action compensates repeatable mismatch and disturbance-induced error \cite{johannink2019residual}. Unlike end-to-end learned cable-robot control, the actor does not replace the full joint-torque command, and its utilization can be evaluated directly through normalized action peaks, near-limit duration, and RMS torque ratios \cite{xiong2019hybrid,joyo2025drl}. The largest interval-wise improvements occur during active parameter uncertainty and disturbance, aligning the learned action with its designated compensation role.

The cable analysis complements the tracking comparison by quantifying pull-only feasibility and mapping quality. The desired path remains within the sampled joint-limited and static nonnegative-tension-feasible regions; $\Jc$ retains rank three; and the minimum singular value and condition number identify the least favorably conditioned postures. The higher peak inferred tension in the residual branch makes the tracking--actuation tradeoff explicit. A constrained quadratic-programming allocator with defined pretension and physical $F_{min}$/$F_{max}$ bounds is the appropriate extension for hardware-oriented force allocation.

Relative to rehabilitation-control studies employing robust/adaptive, impedance, admittance, or assist-as-needed strategies, direct numerical ranking is not meaningful because the mechanisms, reference motions, patient-interaction models, and reported metrics differ \cite{seyfi2021robust,li2021hybrid,wang2023bedside}. The contribution here is a traceable integration of interpretable CTC, bounded residual DDPG, trajectory-tracking metrics, joint-limit accounting, cable-tension demand, sampled workspace feasibility, and cable-Jacobian conditioning within one controlled comparison.

%% file: sections/06b_revision_results.tex
\FloatBarrier
\section{Additional comparisons and sensitivity results}\label{sec:revision_results}
\begin{revisionblock}
This section reports new analyses conducted in response to the reviewers; they do not replace the archived primary-case metrics presented above. Original-policy transfer evaluations, newly trained policies, different force-limited model configurations, and corrected retrospective analyses are labeled independently. Experimental means refer to the explicit unit of replication stated in each protocol. Results after a recorded joint-limit crossing are not treated as admissible full-horizon tracking.
\end{revisionblock}

\subsection{Training transparency, reproducibility and DDPG alternatives}
\begin{revisionblock}
The surviving original \texttt{Agent20} history contains 20 episode rewards, with the largest recorded value $-122.919$ at episode 10 and episode-20 value $-151.058$. These are partial training records, not evidence that the originally selected policy converged or was optimal. The five new 80-episode training runs had mean episode-80 nominal RMSE $14.683\pm6.019$~mm (sample mean and SD over independently trained seeds). All five policies passed the predetermined episode-60 check, while one failed the episode-80 check at 25.365~mm RMSE and 99.80\% near-saturation. The original checkpoint was not replaced. The loss of the original seed and complete training history prevents exact reproduction of its original selection process.
\end{revisionblock}
\input{tables/training_gate}
\begin{revisionblock}
Supplementary Table~S2 details the matched three-seed optimizer-hyperparameter experiment, and Supplementary Fig.~S1 displays the corresponding newly generated return curves.
\end{revisionblock}
\begin{figure}[!htbp]\centering
\includegraphics[width=0.82\linewidth,height=0.27\textheight,keepaspectratio]{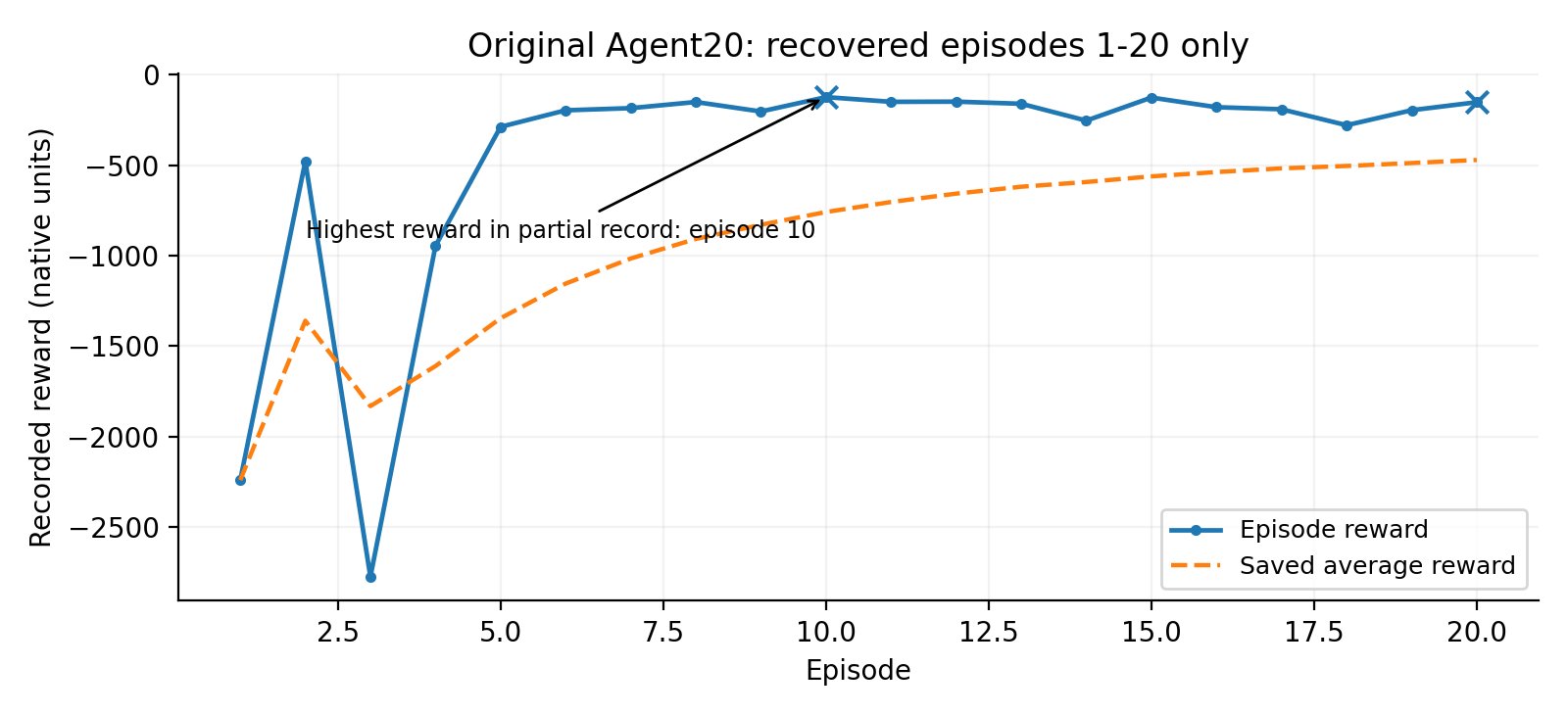}
\caption{\revcaption{Authentically recovered episode-level reward history for the first 20 episodes of the archived evaluated policy. No later original training records are inferred.}}\label{fig:revision_partial_train}
\end{figure}
\begin{figure}[!htbp]\centering
\includegraphics[width=0.85\linewidth,height=0.27\textheight,keepaspectratio]{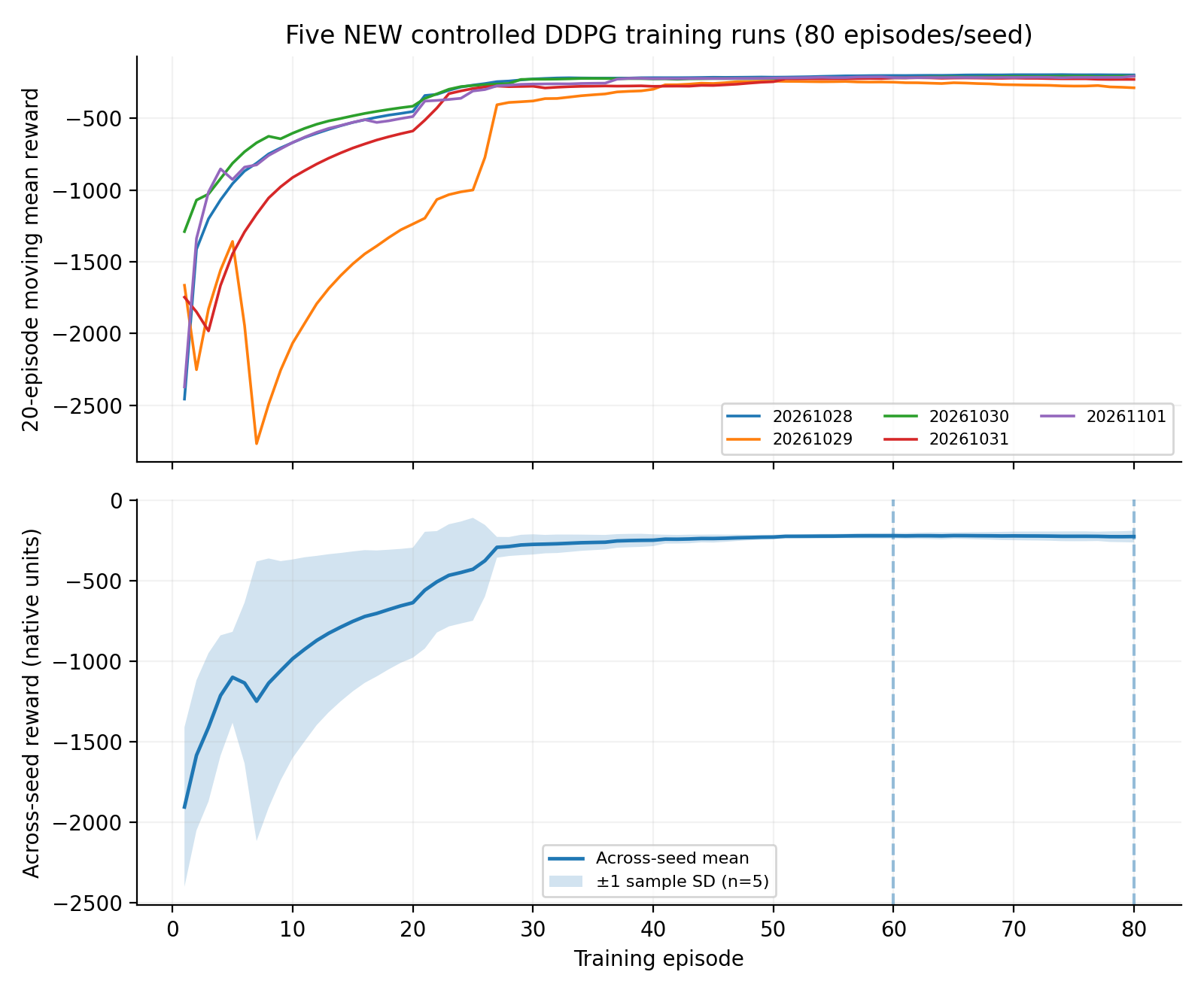}
\caption{\revcaption{Five independently initialized \emph{new} 80-episode DDPG runs. Their across-seed variation is not the original \texttt{Agent20} convergence history.}}\label{fig:revision_new_train}
\end{figure}
\begin{revisionblock}
In the separate three-seed hyperparameter test, mean nominal episode-80 RMSE was 16.316~mm with the reference settings, 13.319~mm with $\gamma=0.95$ and 11.738~mm with doubled exploration-noise variance. These descriptive means do not imply uniform paired improvement: the lower discount factor helped only one of three matched seeds, and the experiment changed one parameter at a time. A separate single-seed-per-bound retraining test on W7--W10 produced mean RMSE values 12.270, 13.183 and 15.359~mm with $0.5\times$, $1.0\times$ and $1.5\times$ native action limits; the frozen submitted policy gave 12.999~mm under that protocol. The $0.5\times$ policy had a smaller minimum limit clearance (2.653$^\circ$ versus 4.228$^\circ$ for the frozen reference). No analytical bound-optimality result follows: an ideal mismatch envelope would require $B_i=\sup_{z\in\Omega}|\bm e_i^T\Delta\bm\tau(z)|$ over a specified uncertainty domain, which is not established here.

Under the independent five-seed DDPG/TD3/SAC study, selected-policy mean W11--W14 RMSE was $15.323\pm3.417$~mm (DDPG), $18.785\pm16.329$~mm (TD3) and $12.295\pm2.003$~mm (SAC). TD3's unfavorable second seed (47.989~mm) remains in the mean and dispersion; SAC had the lowest mean but TD3 a slightly lower median. The submitted \texttt{Agent20} gave 12.997~mm separately. An exploratory unadjusted exact permutation comparison for SAC versus new DDPG yielded $p=0.111$; no statistically established or universal algorithm superiority is claimed. A complete return history for the fresh DDPG seed S01 was not retained in the original response evidence, so missing points are not reconstructed.
\end{revisionblock}
\input{tables/algorithms}
\begin{revisionblock}
Supplementary Table~S4 compares the three one-seed-per-native-bound policies with the original frozen checkpoint; it is separate from the five-seed DDPG/TD3/SAC benchmark.
\end{revisionblock}
\begin{figure}[!htbp]\centering
\includegraphics[width=0.84\linewidth,height=0.27\textheight,keepaspectratio]{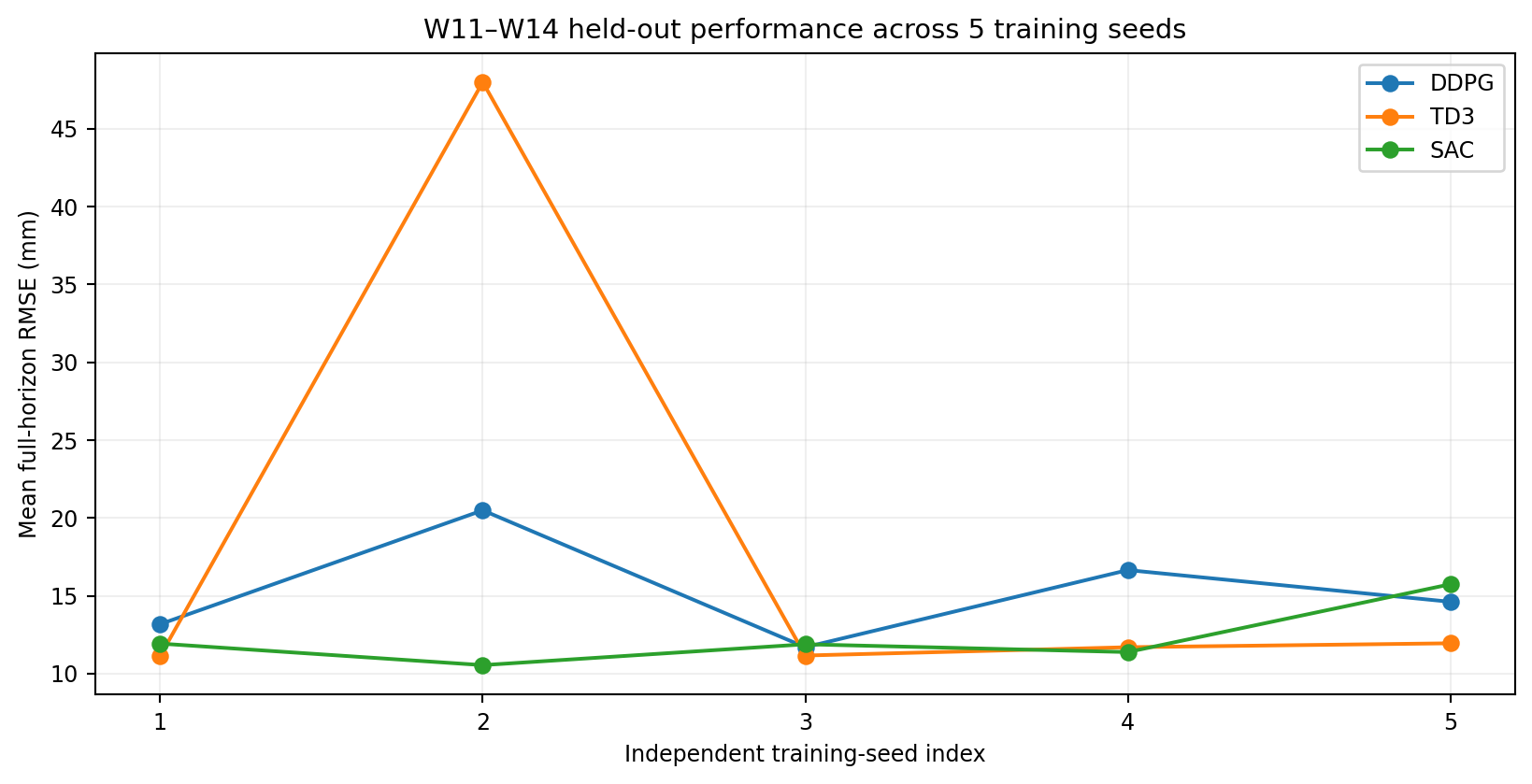}
\caption{\revcaption{Matched-budget five-training-seed DDPG/TD3/SAC test. The fixed originally submitted DDPG checkpoint is a separate reference.}}\label{fig:revision_algorithms}
\end{figure}

\subsection{Reward and correction-authority trade-offs}
\begin{revisionblock}
The original fixed-policy authority test gave RMSE 7.708, 7.174 and 7.192~mm at gains 0.50, 0.75 and 1.00, with zero sampled limit violations but decreasing minimum clearances of 7.319$^\circ$, 6.793$^\circ$ and 6.340$^\circ$. This is a gain ablation, not policy retraining or identification of the physical torque envelope.

The single-seed allocation-gap reward ablation reduced the episode-80 nominal mean pre-lag static torque-gap from 0.210862 to approximately $3.42\times10^{-9}$~N\,m, but worsened RMSE from 15.188 to 25.324~mm (66.74\% higher) and increased violating \emph{time samples} from zero to 160 of 1{,}001. The disturbance-case RMSE similarly worsened (15.463 to 24.674~mm). The allocator clipped both arms to the assumed force bounds, and the reward penalized requested-versus-allocated torque rather than negative raw inferred tension. Near-zero static mismatch thus did not establish safe, feasible or better tracking.
\end{revisionblock}
\input{tables/reward_penalty}
\begin{figure}[!htbp]\centering
\includegraphics[width=0.82\linewidth,height=0.27\textheight,keepaspectratio]{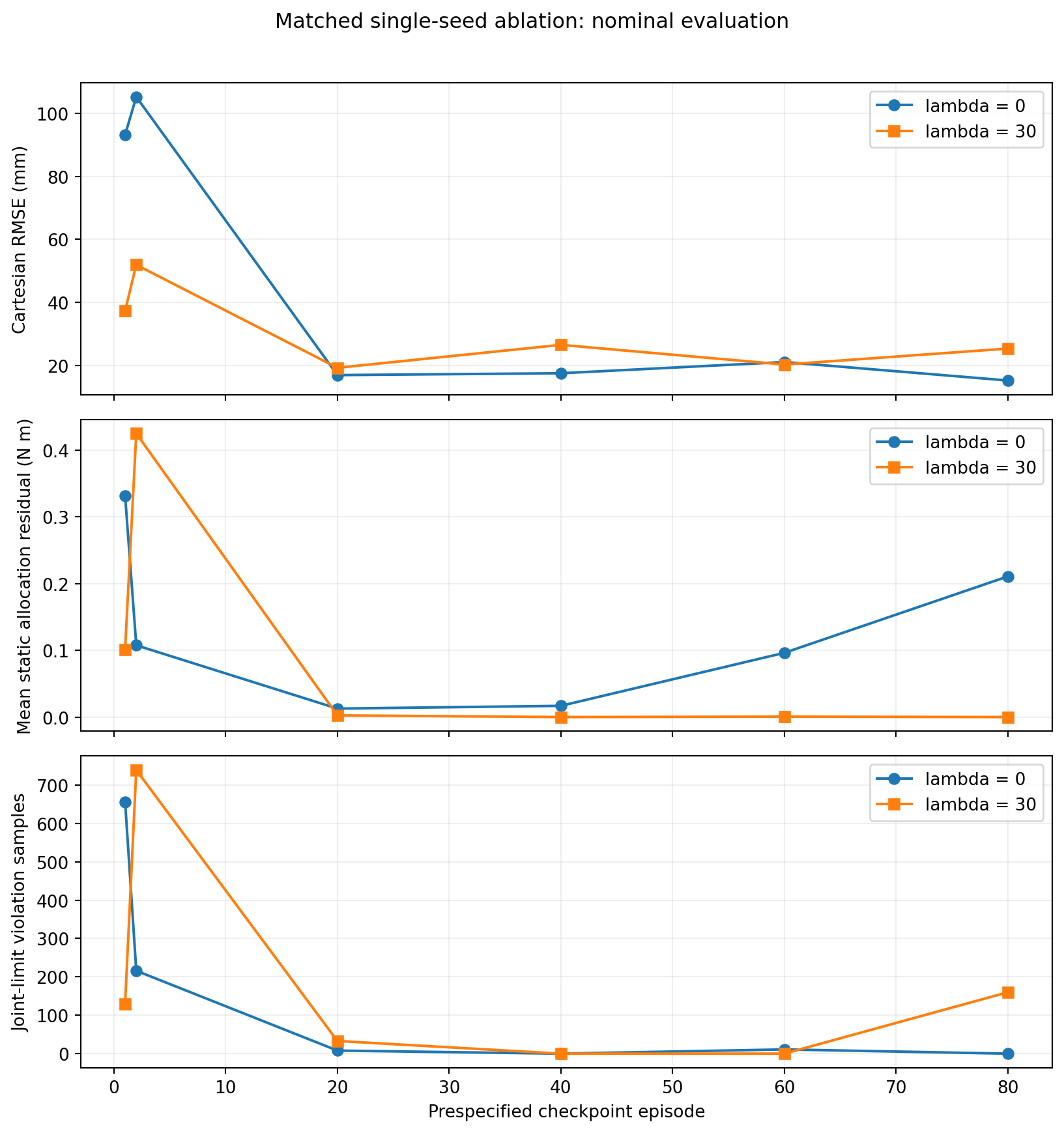}
\caption{\revcaption{Single-seed modified-reward comparison: reducing static allocation mismatch can worsen tracking and sampled joint-limit compliance.}}\label{fig:revision_gapreward}
\end{figure}
\begin{revisionblock}
The four-coefficient three-level one-factor reward study produced 27 selected checkpoints and 54 fixed-noise evaluations. Of those, 39 runs terminated at joint limits, 15 reached 10~s, and only seven checkpoints completed \emph{both} waveforms. The 81-combination full factorial requested by the reviewer was not performed. Under its two-wave selection criterion, the best eligible $w_{\Delta u}=0.100$ checkpoint achieved 13.695~mm mean RMSE, 4.344$^\circ$ minimum clearance and 0.07093~N\,m static gap. The eligible $w_u=0.100$ checkpoint gave 14.530~mm, 2.689$^\circ$ and 0.02468~N\,m. Four of the seven eligible checkpoint objective triples were nondominated; no unique global optimum was demonstrated. The same two evaluation waveforms supported screening and ranking, so the resulting Pareto set is selection-conditioned, not an independent hold-out finding.
\end{revisionblock}
\input{tables/reward_sweep}
\begin{revisionblock}
Supplementary Table~S5 lists the seven eligible reward-study checkpoints and their Pareto classification.
\end{revisionblock}
\begin{figure}[!htbp]\centering
\includegraphics[width=0.84\linewidth,height=0.27\textheight,keepaspectratio]{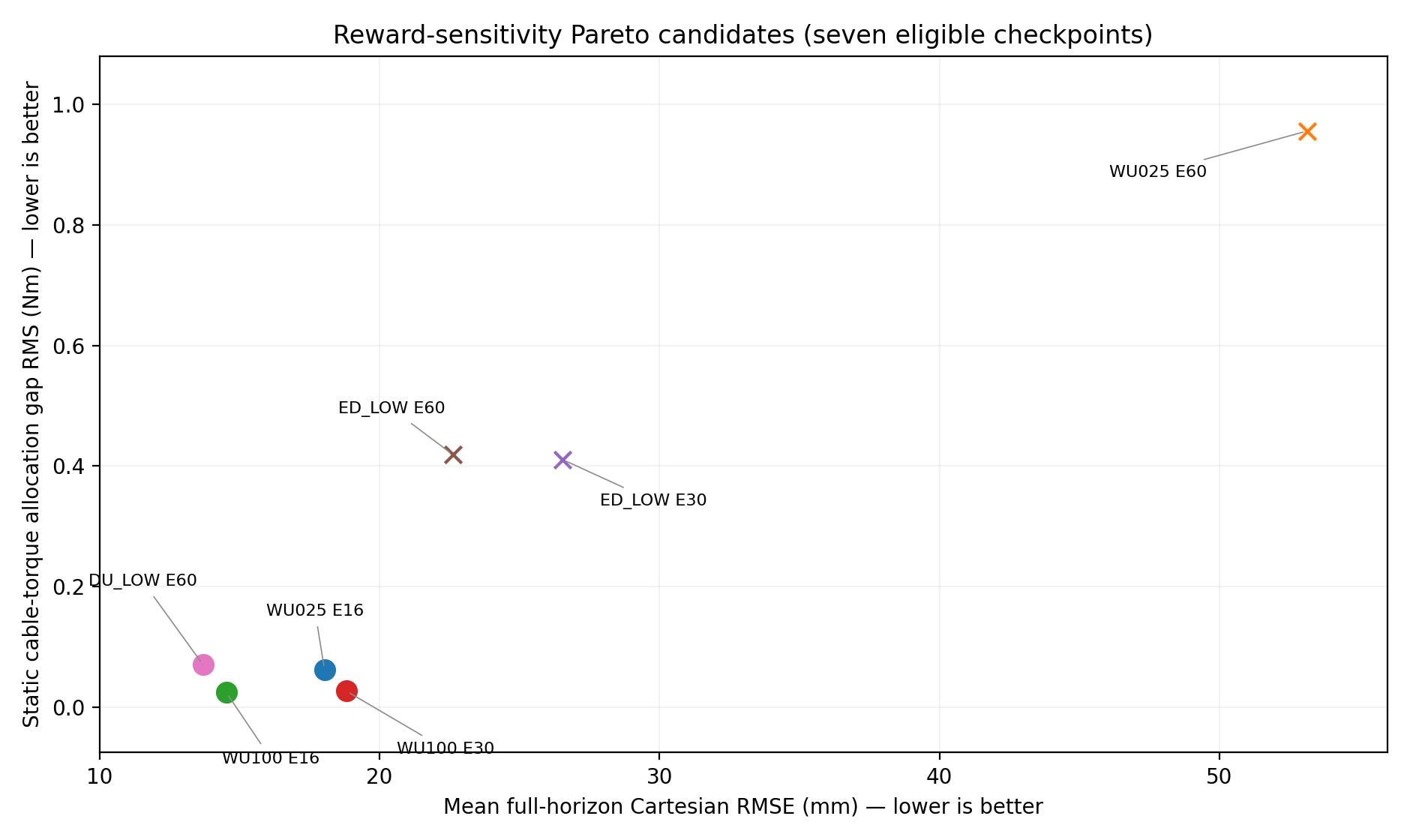}
\caption{\revcaption{Descriptive tracking--clearance--allocation Pareto comparison among the seven eligible reward-sweep checkpoints; four are nondominated.}}\label{fig:revision_pareto}
\end{figure}

\subsection{Constrained cable-force sensitivity and response lag}
\begin{revisionblock}
In the force-actuated supplemental model, the assumed cap of 50~N yielded 38.551~mm CTC versus 74.382~mm residual-DDPG RMSE and 114 versus 248 violating joint-channel samples. At 60~N, RMSEs were 22.516/22.794~mm and the CTC branch had 48 violating samples while DDPG had zero. At 70--100~N, both completed without sampled violations and the residual branch was more accurate under that selected configuration. At a 100~N cap, peak numerical tensions were 66.001~N (CTC) and 86.237~N (DDPG), leaving 33.999/13.763~N upper margins; this is not evidence of motor saturation limits or physical pretension. The 0.1~N lower floor was active at one and 36 cable-channel samples, respectively.
\end{revisionblock}
\input{tables/force_caps}
\begin{revisionblock}
Supplementary Table~S1 lists individual force-cap activations and limit-violation counts; caps are assumed, not measured motor ratings.
\end{revisionblock}
\begin{figure}[!htbp]\centering
\includegraphics[width=0.80\linewidth,height=0.27\textheight,keepaspectratio]{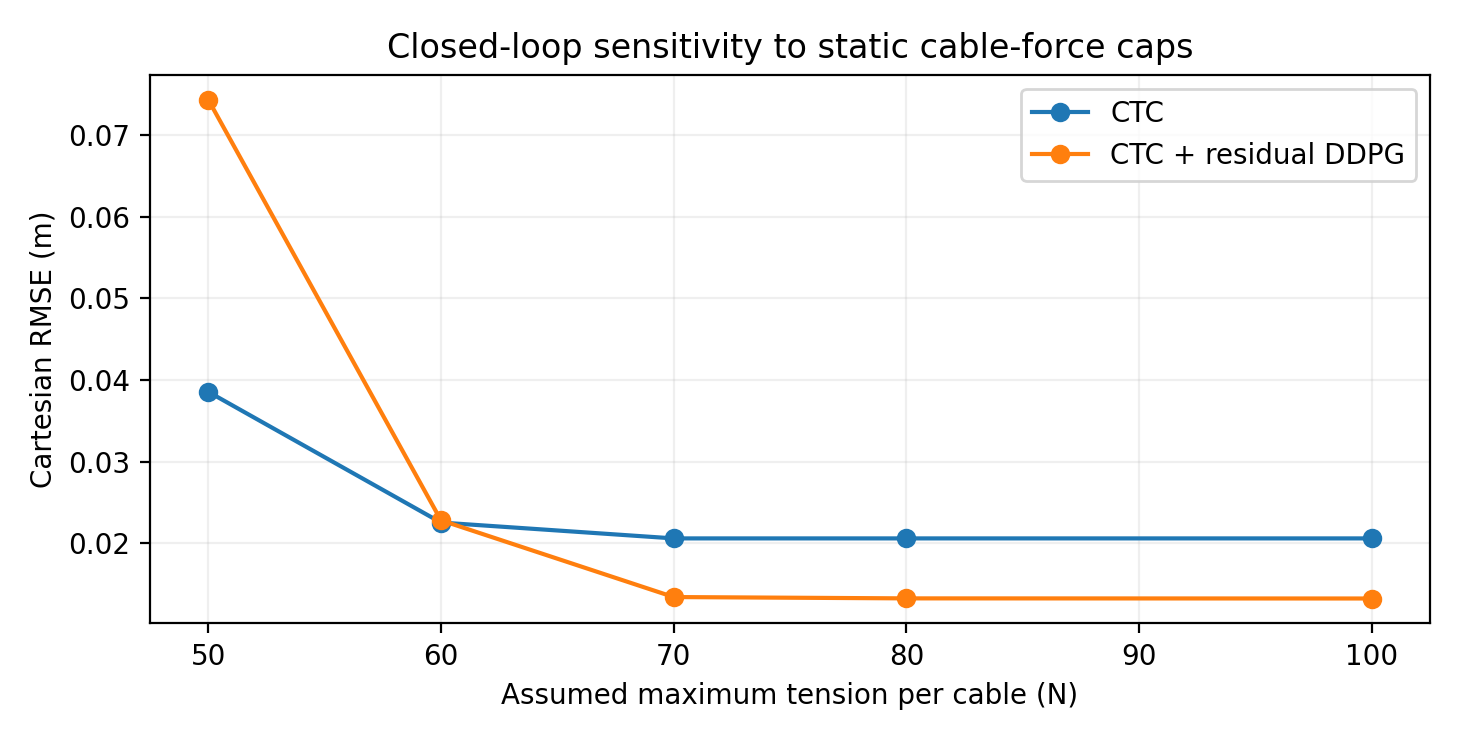}
\caption{\revcaption{Tracking sensitivity to \emph{assumed} per-cable force caps in the supplemental constrained numerical actuator model.}}\label{fig:revision_caps}
\end{figure}
\begin{revisionblock}
With an assumed 30-ms first-order response and 100-N cap, the CTC/residual RMSE values were 24.494/15.126~mm with no sampled violations; at 70~N, they were 25.475/30.541~mm, reversing the ranking. At a 100-N cap and 40-ms response, the residual branch incurred 68 violating channel samples; with 100-ms lag both branches crossed joint limits. Reported full-horizon numerical continuation after a limit crossing is \emph{not} a valid physically admissible outcome. The ideal response $G(s)=1/(T_as+1)$ attenuates rapid force requests and adds phase lag, but does not model actual cable elasticity, pulley inertia, friction or motor-controller bandwidth.
\end{revisionblock}
\input{tables/force_lag}
\begin{figure}[!htbp]\centering
\includegraphics[width=0.81\linewidth,height=0.27\textheight,keepaspectratio]{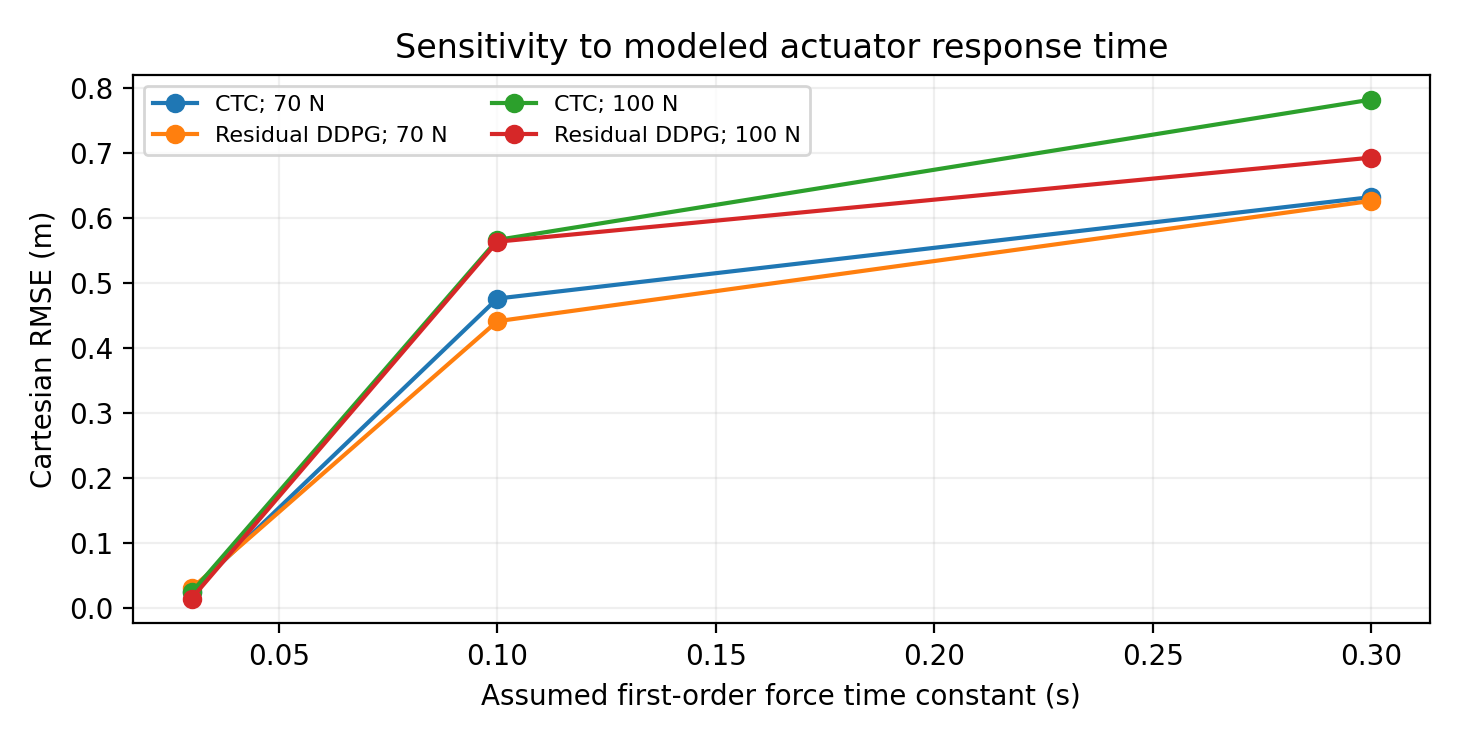}
\caption{\revcaption{Tracking and termination sensitivity to idealized numerical cable-force response times. Curves are not hardware frequency-response measurements.}}\label{fig:revision_lag}
\end{figure}

\subsection{Comparative feedback compensation and control effort}
\begin{revisionblock}
The reduced-order ESO, calibrated with lower correction limits than DDPG, achieved a smaller RMSE than CTC across four synthetic trajectories but not the lowest recorded error. The descriptive cross-reference means were 24.63, 20.77 and 7.17~mm for CTC, CTC+ESO and CTC+frozen-DDPG, respectively. For the ellipse, applied generalized-torque RMS was 15.731, 16.213 and 16.056~N\,m, respectively; the corresponding integrated squared torques were 2476.05, 2629.86 and 2579.39~(N\,m)$^2$s. These are modeled applied torques, not physical motor energy. Because the ESO was limited to $\pm[1,0.6,0.4]$~N\,m and DDPG to $\pm[5,3,2]$~N\,m, no equal-authority or algorithm-independent superiority inference is justified.
\end{revisionblock}
\input{tables/eso_paths}
\begin{revisionblock}
Supplementary Table~S3 reports the elliptical applied generalized-torque effort, and Supplementary Fig.~S2 shows the associated additional torque comparison.
\end{revisionblock}
\begin{revisionblock}
In a different feedback-family comparison, 162 gait-tuning candidates were examined across six grids, 32 were excluded for sampled joint-limit violations, and 130 met the gait eligibility criterion. The original frozen residual policy had the lowest \emph{observed} RMSE on the gait-like and elliptical references (7.192 and 7.291~mm). Among five adapted families, CMAC-FOPID had the lowest selected gait-like RMSE (15.047~mm), while BELBIC-PID had the lowest ellipse RMSE (14.994~mm). These are tuned application-specific approximations, with differing configurations, only two tested reference types, and no repeated-trial uncertainty bands for all families. Two requested controller families were not evaluated. This comparison does not establish the method's universal superiority.
\end{revisionblock}
\input{tables/advanced_controllers}
\begin{figure}[!htbp]\centering
\includegraphics[width=0.83\linewidth,height=0.27\textheight,keepaspectratio]{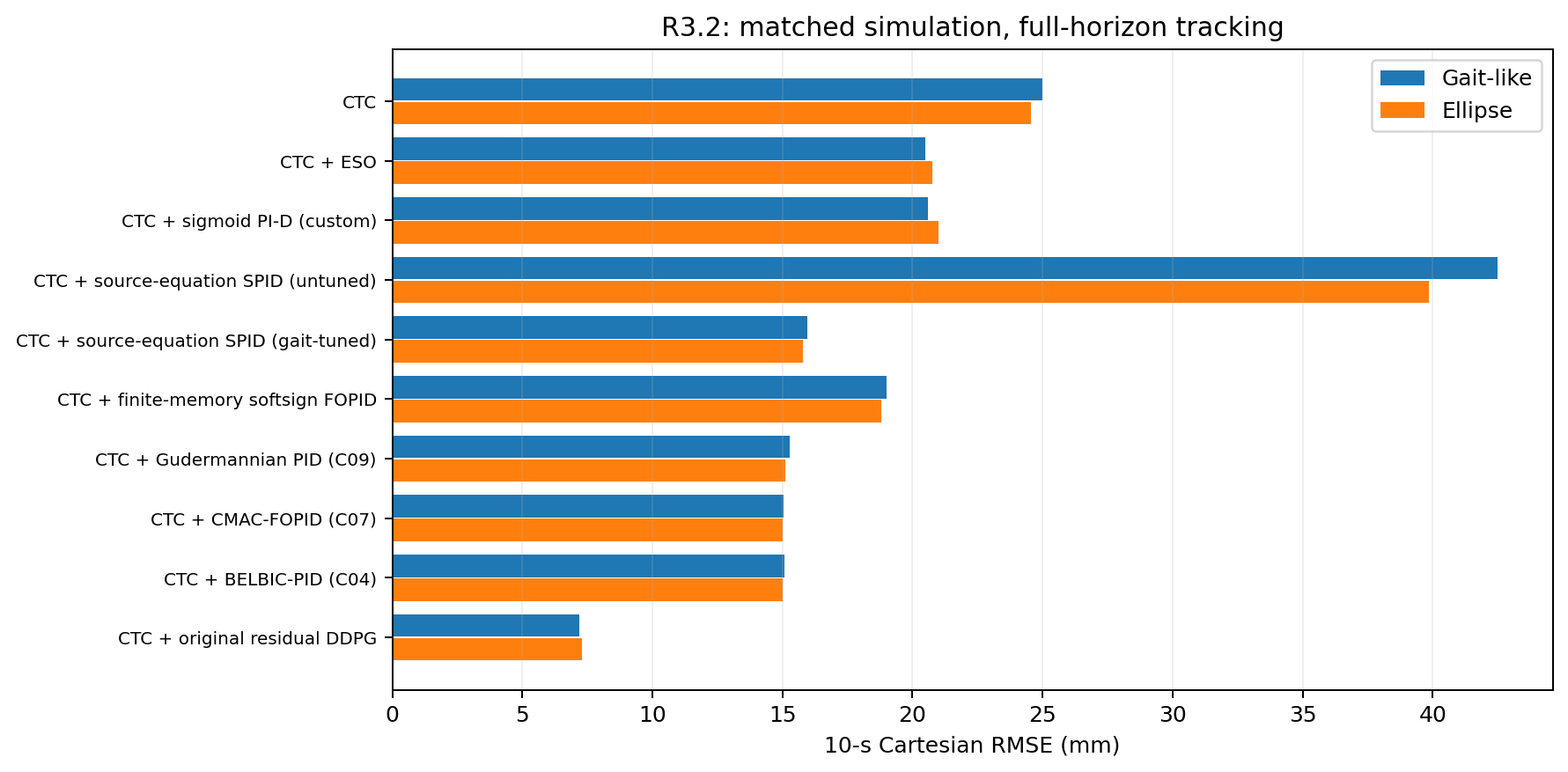}
\caption{\revcaption{Robot-adapted feedback-family comparisons on gait-like and elliptical trajectories. The tuning scenario was not an independent hold-out reference.}}\label{fig:revision_controllers}
\end{figure}

\subsection{Robustness, trajectory transfer and early termination}
\begin{revisionblock}
Eighteen mechanical mismatch cases yielded 13/18 completed CTC and 12/18 completed DDPG runs, with eight mutual completions. On those eight paired full-horizon runs, residual-DDPG RMSE reduction ranged from 23.20\% to 43.37\%. Both methods completed all six inertia cases, with CTC RMSE 24.65--24.80~mm and DDPG 15.14--15.25~mm. Negative mass/length changes frequently terminated the residual controller before disturbance onset. At shortened link lengths of $-20$, $-10$ and $-5\%$, the reference radius exceeded the necessary unconstrained maximum reach at 1001, 496 and 191 of 1001 samples. Hence these tests confound physical reference reachability, initial-state mismatch and robustness; they do not isolate a universal parameter-sensitivity ranking.
\end{revisionblock}
\input{tables/parameter_completion}
\begin{revisionblock}
Supplementary Table~S8 provides every mass-, inertia- and length-family case, including stopping times; Supplementary Figs.~S3 and S4 visualize terminations and necessary reachability conditions.
\end{revisionblock}
\begin{figure}[!htbp]\centering
\includegraphics[width=0.83\linewidth,height=0.27\textheight,keepaspectratio]{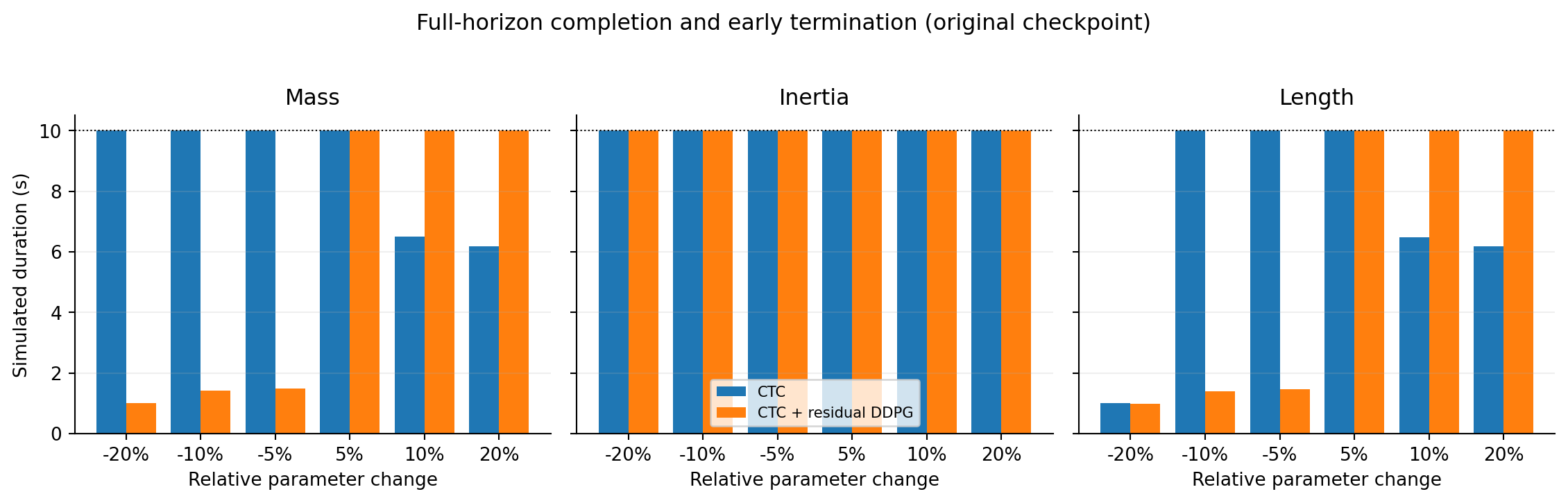}
\caption{\revcaption{Full-horizon completion outcomes across independent mechanical-parameter-family perturbations; stopped trials are not assigned full-horizon RMSE.}}\label{fig:revision_perturb}
\end{figure}
\begin{revisionblock}
Across ten paired synthetic disturbance shapes at nominal RMS all ten controller pairs completed, with mean paired RMSE reduction $38.740\pm0.279$ percentage points. At $1.5\times$ RMS, nine of ten pairs completed and the conditional mean was $39.098\pm0.418$ points; the incomplete CTC run stopped at 6.78~s. These independent waveform shapes are \emph{not} independently trained policies. In a separate distal point-load test, the 0.5-kg CTC run stopped at 0.20~s while the DDPG run completed at 29.554~mm RMSE; at 1~kg both stopped (CTC 0.17~s, DDPG 3.48~s). Failures constrain the tested loading envelope.
\end{revisionblock}
\input{tables/noise_levels}
\begin{revisionblock}
On four programmed references under one additional fixed mismatch/noise protocol, the frozen \texttt{Agent20} policy completed every case without sampled violations, with respective gait/ellipse/point-to-point/stochastic-phase RMSE values of 7.192/7.291/6.937/7.251~mm versus 24.989/24.552/24.833/24.165~mm under CTC. In a separate bounded-force version, commanded and post-lag tensions remained in the assumed 0.1--100-N range by construction, while raw unconstrained tension demand exceeded that range in 8 of 1001 samples per Agent20 reference and one of 1001 CTC samples. The four-class mean delivered-torque gap was 1.244 versus 0.988~N\,m RMS. Feasible numerical clipped commands do not mean exact generalized-torque realization.

An 80-episode, one-seed mixed-reference warm-start experiment terminated 70 training episodes early; checkpoints E20 and E40 also stopped early in all four fixed evaluations. E60 and E80 completed, but mean full-horizon RMSE was 30.535 and 25.509~mm, respectively, worse than the original frozen policy's 7.168~mm. The E80 raw force-demand infeasibility fraction averaged 25.32\% across four references (0.80\% for Agent20), and its mean delivered torque gap was 2.720 versus 1.244~N\,m RMS. Thus the added fine-tuning failed in this test and does not establish improved multi-trajectory training.
\end{revisionblock}
\input{tables/trajectory_transfer}
\begin{revisionblock}
Supplementary Tables~S6 and S7 report the aggregate and per-reference force-demand discrepancies for the transfer and fine-tuning tests.
\end{revisionblock}
\begin{figure}[!htbp]\centering
\includegraphics[width=0.84\linewidth,height=0.27\textheight,keepaspectratio]{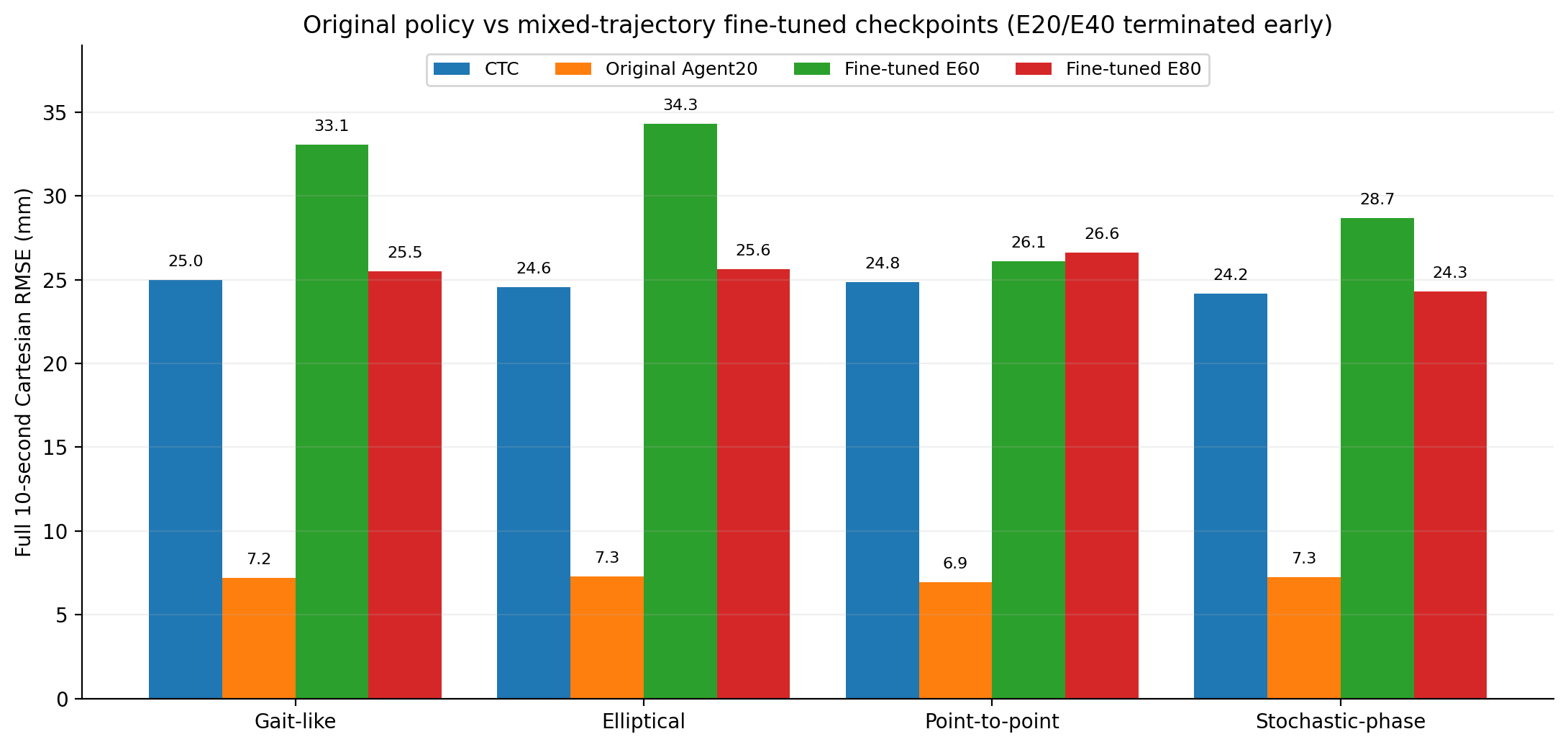}
\caption{\revcaption{Four synthetic reference classes for frozen-policy transfer and separately evaluated single-seed mixed-reference fine-tuning.}}\label{fig:revision_transfer}
\end{figure}

\subsection{Joint-limit trade-off and historical checkpoint audit}
\begin{revisionblock}
The original primary-case joint-limit log had 61 violating CTC joint-channel samples and none for residual DDPG, while the original alternative circular evaluation had 44 under CTC and 101 under DDPG. In the circular case the reference itself touched the joint-2 lower limit for 166 sampled times. A \emph{separate matched replay} showed 47 versus 102 samples for CTC and unguarded DDPG. The heuristic joint-2 attenuation reduced total DDPG violations from 102 to 47 (53.92\%) and joint-2 violations from 102 to 10, but caused 37 joint-3 lower-limit violations and increased RMSE from 12.052 to 17.497~mm. These replay values must not be substituted for the historical 44/101 counts in the original paper. Neither the original policy nor the gate provides a formally invariant all-joint or cable-tension safe set.
\end{revisionblock}
\input{tables/joint_gate}
\begin{figure}[!htbp]\centering
\includegraphics[width=0.79\linewidth,height=0.27\textheight,keepaspectratio]{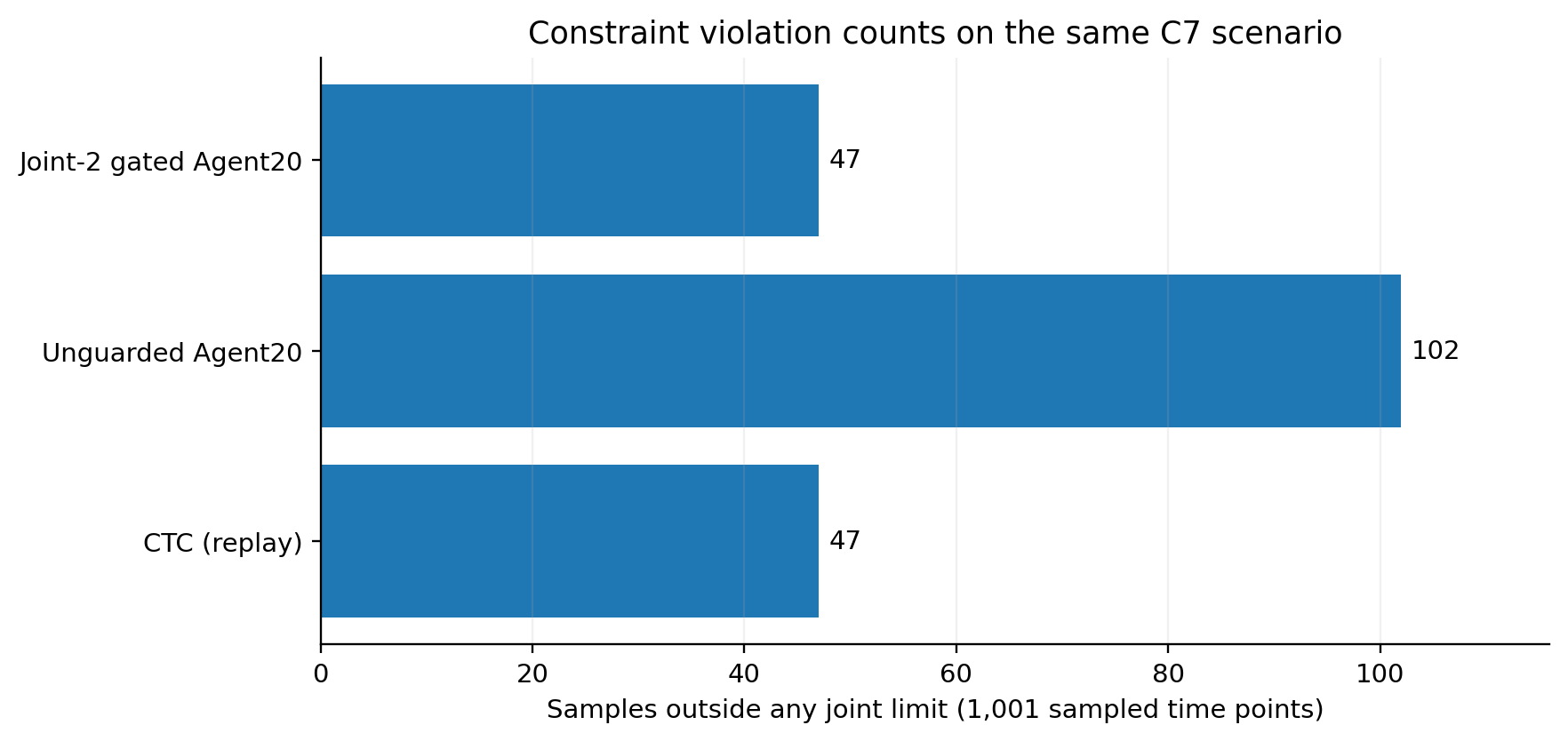}
\caption{\revcaption{Matched circular-reference joint-limit sample counts (CTC 47, ungated DDPG 102, heuristic-gated DDPG 47), a separate protocol from the original 44/101 log.}}\label{fig:revision_gate}
\end{figure}
\begin{revisionblock}
A distinct retrospective evaluation compared 23 SHA-256-distinct surviving DDPG policies, each on primary and circular references under five externally seeded disturbance realizations. The five waveforms were repeated \emph{evaluations}, not five training seeds. The submitted \texttt{Agent20} had the second-lowest mean RMSE among the reported policies (12.168~mm), but 1031 violating joint-channel samples across the tests; \texttt{Agent13\_3cables.mat} gave 12.019~mm but 97 violating samples, while \texttt{Agent13.mat} gave 12.739~mm with zero violating samples. This retrospectively reveals the tracking-versus-limits trade-off and does not reconstruct how the authors originally selected \texttt{Agent20}. A previously invalid repeated-noise audit was excluded from the corrected comparison.
\end{revisionblock}
\input{tables/checkpoint_audit}
\begin{revisionblock}
Supplementary Fig.~S5 presents the corrected retrospective ranking of surviving checkpoint policies; its five external disturbance trials are not new training seeds.
\end{revisionblock}
\begin{figure}[!htbp]\centering
\includegraphics[width=0.83\linewidth,height=0.27\textheight,keepaspectratio]{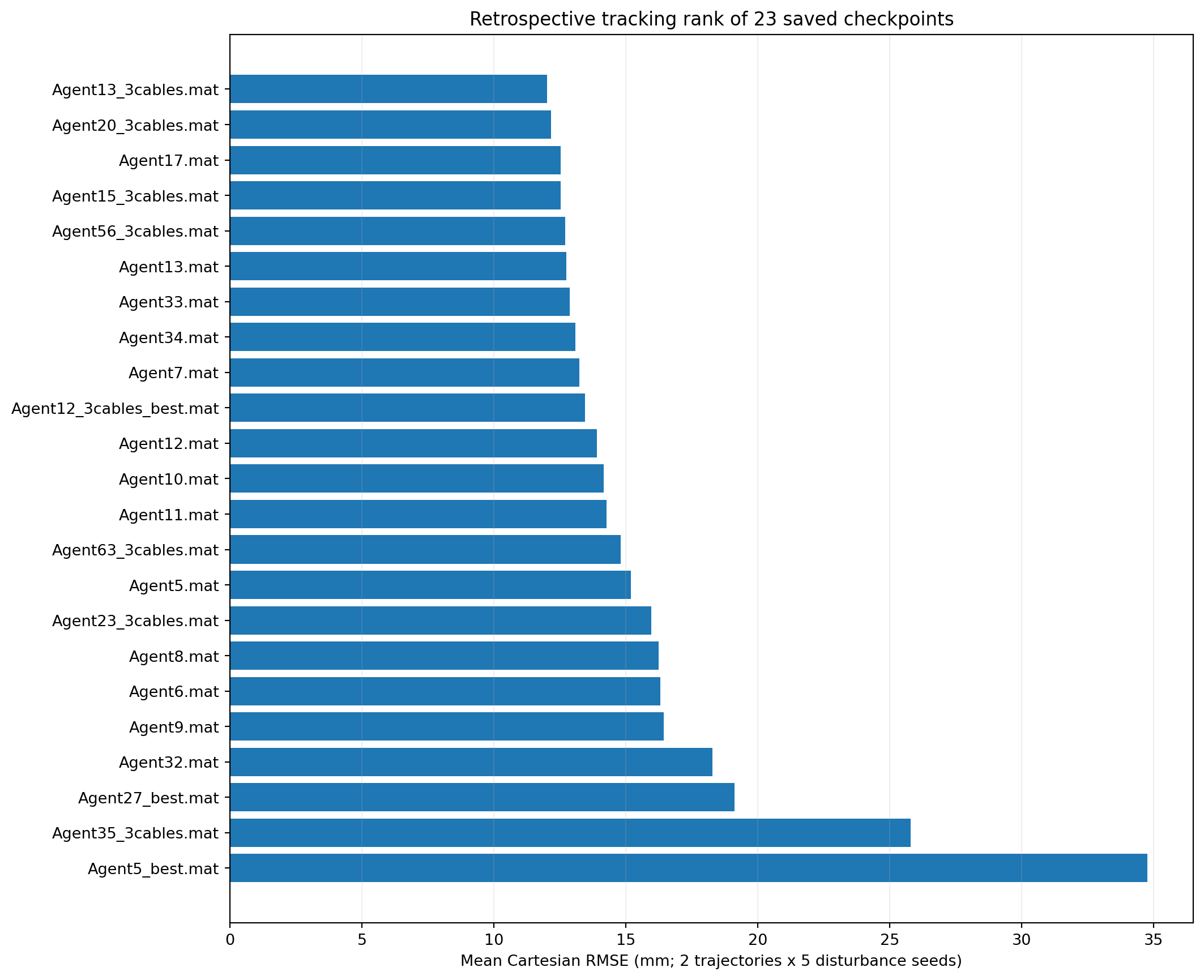}
\caption{\revcaption{Corrected retrospective performance comparison across historical checkpoints. Disturbance-waveform repetitions are separate from training-seed replicates.}}\label{fig:revision_retro}
\end{figure}

\FloatBarrier

%% file: tables/training_gate.tex
\begin{table}[H]
\centering
\scriptsize
\setlength{\tabcolsep}{3pt}
\begin{tabular}{@{}llll@{}}
\toprule
Seed & EP60 RMSE (mm) & EP80 RMSE (mm) & Both pass \\
\midrule
20261028 & 11.502 & 12.458 & PASS \\
20261029 & 19.416 & 25.365 & FAIL \\
20261030 & 11.230 & 11.125 & PASS \\
20261031 & 11.694 & 11.458 & PASS \\
20261101 & 10.904 & 13.008 & PASS \\
\bottomrule
\end{tabular}
\caption{\revcaption{Independent 80-episode DDPG training: predeclared episode-60 and episode-80 checkpoint tests; all failures retained.}}
\label{tab:revised_training}
\end{table}

%% file: tables/algorithms.tex
\begin{table}[H]
\centering
\scriptsize
\setlength{\tabcolsep}{3pt}
\begin{tabular}{@{}lllll@{}}
\toprule
Method & Training seeds & Held-out RMSE (mm) & SD (mm) & Median (mm) \\
\midrule
Fresh DDPG & 5 & 15.323 & 3.417 & 14.613 \\
TD3 & 5 & 18.785 & 16.329 & 11.694 \\
SAC & 5 & 12.295 & 2.003 & 11.883 \\
Submitted DDPG & Fixed reference & 12.997 & -- & -- \\
\bottomrule
\end{tabular}
\caption{\revcaption{Selected-policy full-horizon Cartesian RMSE on the four held-out synthetic disturbances W11--W14 (five independent training seeds per method).}}
\label{tab:revised_algorithms}
\end{table}

%% file: tables/reward_penalty.tex
\begin{table}[H]
\centering
\scriptsize
\setlength{\tabcolsep}{3pt}
\begin{tabular}{@{}llll@{}}
\toprule
Scenario & Metric & lambda = 0 & lambda = 30 \\
\midrule
Nominal & Tracking RMSE (mm) & 15.188 & 25.324 \\
Nominal & Mean static allocation residual (N m) & 0.210862 & 3.42e-09 \\
Nominal & Joint-limit time samples & 0 & 160 \\
Disturbance 5-9 s & Tracking RMSE (mm) & 15.463 & 24.674 \\
Disturbance 5-9 s & Mean static allocation residual (N m) & 0.189536 & 3.45e-09 \\
Disturbance 5-9 s & Joint-limit time samples & 0 & 160 \\
\bottomrule
\end{tabular}
\caption{\revcaption{Single-seed allocation-gap reward ablation: competing tracking, static torque-allocation gap and joint-limit outcomes.}}
\label{tab:revised_gapreward}
\end{table}

%% file: tables/reward_sweep.tex
\begin{table}[H]
\centering
\scriptsize
\setlength{\tabcolsep}{3pt}
\resizebox{\linewidth}{!}{%
\begin{tabular}{@{}llllll@{}}
\toprule
Reward coefficient & Level & Weight & Eligible & Violation runs & Best eligible mean RMSE (mm) \\
\midrule
Position error & Low & 1 & 0/3 & 5/6 (83.3\%) & -- \\
Position error & Baseline & 2 & 0/3 & 6/6 (100.0\%) & -- \\
Position error & High & 4 & 0/3 & 6/6 (100.0\%) & -- \\
Velocity error & Low & 0.175 & 2/3 & 2/6 (33.3\%) & 22.632 \\
Velocity error & Baseline & 0.35 & 0/3 & 6/6 (100.0\%) & -- \\
Velocity error & High & 0.7 & 0/3 & 6/6 (100.0\%) & -- \\
Residual effort & Low & 0.025 & 2/3 & 2/6 (33.3\%) & 18.038 \\
Residual effort & Baseline & 0.05 & 0/3 & 6/6 (100.0\%) & -- \\
Residual effort & High & 0.1 & 2/3 & 2/6 (33.3\%) & 14.530 \\
Action change & Low & 0.1 & 1/3 & 4/6 (66.7\%) & 13.695 \\
Action change & Baseline & 0.2 & 0/3 & 6/6 (100.0\%) & -- \\
Action change & High & 0.4 & 0/3 & 6/6 (100.0\%) & -- \\
\bottomrule
\end{tabular}}
\caption{\revcaption{Three-level one-factor-at-a-time reward study: checkpoint eligibility and joint-limit termination outcomes.}}
\label{tab:revised_reward_sweep}
\end{table}

%% file: tables/force_caps.tex
\begin{table}[H]
\centering
\scriptsize
\setlength{\tabcolsep}{3pt}
\begin{tabular}{@{}lllll@{}}
\toprule
Force cap (N) & CTC RMSE (mm) & DDPG RMSE (mm) & CTC violations & DDPG violations \\
\midrule
50 & 38.551 & 74.382 & 114 & 248 \\
60 & 22.516 & 22.794 & 48 & 0 \\
70 & 20.572 & 13.380 & 0 & 0 \\
80 & 20.572 & 13.217 & 0 & 0 \\
100 & 20.572 & 13.203 & 0 & 0 \\
\bottomrule
\end{tabular}
\caption{\revcaption{Force-capacity sensitivity in the \emph{separate} force-allocation numerical model.}}
\label{tab:revised_forcecaps}
\end{table}

%% file: tables/force_lag.tex
\begin{table}[H]
\centering
\scriptsize
\setlength{\tabcolsep}{3pt}
\begin{tabular}{@{}lllll@{}}
\toprule
Cap & Lag & CTC RMSE (mm) & DDPG RMSE (mm) & Limit samples \\
\midrule
70 N & 30 ms & 25.475 & 30.541 & 0 / 0 \\
100 N & 30 ms & 24.494 & 15.126 & 0 / 0 \\
100 N & 100 ms & 565.970* & 563.264* & 1228 / 1183 \\
100 N & 300 ms & 781.545* & 692.417* & 1798 / 1665 \\
\bottomrule
\end{tabular}
\caption{\revcaption{Illustrative actuator-lag sensitivity (starred post-violation RMSE is numerical continuation only).}}
\label{tab:revised_lag}
\end{table}

%% file: tables/eso_paths.tex
\begin{table}[H]
\centering
\scriptsize
\setlength{\tabcolsep}{3pt}
\resizebox{\linewidth}{!}{%
\begin{tabular}{@{}lllllll@{}}
\toprule
Reference & CTC full & ESO full & DDPG full & CTC 5--9s & ESO 5--9s & DDPG 5--9s \\
\midrule
Gait Like & 24.99 & 20.49 & 7.19 & 19.80 & 6.69 & 5.70 \\
Ellipse & 24.55 & 20.75 & 7.29 & 18.54 & 6.87 & 6.27 \\
Line P2P & 24.83 & 20.84 & 6.94 & 20.30 & 7.44 & 4.79 \\
Stochastic Phases & 24.16 & 21.00 & 7.25 & 19.29 & 7.80 & 5.50 \\
\bottomrule
\end{tabular}}
\caption{\revcaption{Matched four-trajectory comparison of CTC, conservatively bounded CTC+ESO and original CTC+residual DDPG (all RMSE values in mm).}}
\label{tab:revised_eso}
\end{table}

%% file: tables/advanced_controllers.tex
\begin{table}[H]
\centering
\scriptsize
\setlength{\tabcolsep}{3pt}
\begin{tabular}{@{}lll@{}}
\toprule
Controller & Gait RMSE (mm) & Ellipse RMSE (mm) \\
\midrule
CTC & 24.989 & 24.552 \\
CTC + ESO & 20.487 & 20.747 \\
CTC + sigmoid PI-D (custom) & 20.591 & 20.988 \\
CTC + robust sigmoid PID (untuned) & 42.494 & 39.846 \\
CTC + robust sigmoid PID (gait-tuned) & 15.954 & 15.781 \\
CTC + softsign FOPID & 19.012 & 18.796 \\
CTC + Gudermannian PID & 15.271 & 15.116 \\
CTC + CMAC-FOPID & 15.047 & 15.006 \\
CTC + BELBIC-PID & 15.067 & 14.994 \\
CTC + original residual DDPG & 7.192 & 7.291 \\
\bottomrule
\end{tabular}
\caption{\revcaption{Illustrative comparison against five robot-adapted feedback-controller families on gait-like and elliptical paths (all RMSE values in mm).}}
\label{tab:revised_adapted}
\end{table}

%% file: tables/parameter_completion.tex
\begin{table}[H]
\centering
\scriptsize
\setlength{\tabcolsep}{3pt}
\begin{tabular}{@{}llll@{}}
\toprule
Parameter family & CTC 10 s & DDPG 10 s & Both 10 s \\
\midrule
Mass & 4/6 & 3/6 & 1 \\
Inertia & 6/6 & 6/6 & 6 \\
Length & 3/6 & 3/6 & 1 \\
Total & 13/18 & 12/18 & 8 \\
\bottomrule
\end{tabular}
\caption{\revcaption{Completion of the 10-s simulated horizon for three one-factor mechanical-parameter families.}}
\label{tab:revised_perturb_completed}
\end{table}

%% file: tables/noise_levels.tex
\begin{table}[H]
\centering
\scriptsize
\setlength{\tabcolsep}{3pt}
\begin{tabular}{@{}lllll@{}}
\toprule
RMS & CTC & DDPG & Paired n & Paired improvement  $\pm$  SD \\
\midrule
1.0$\times$ & 10/10 & 10/10 & 10 & 38.740  $\pm$  0.279 pp \\
1.5$\times$ & 9/10 & 10/10 & 9 & 39.098  $\pm$  0.418 pp \\
\bottomrule
\end{tabular}
\caption{\revcaption{Completed synthetic disturbance-waveform pairs and conditional sample mean improvement.}}
\label{tab:revised_noise}
\end{table}

%% file: tables/trajectory_transfer.tex
\begin{table}[H]
\centering
\scriptsize
\setlength{\tabcolsep}{3pt}
\begin{tabular}{@{}lllll@{}}
\toprule
Trajectory & CTC & Agent20 & E60 & E80 \\
\midrule
Gait-like & 24.989 & 7.192 & 33.059 & 25.493 \\
Elliptical & 24.552 & 7.291 & 34.315 & 25.626 \\
Straight-line P2P & 24.833 & 6.937 & 26.084 & 26.619 \\
Stochastic-phase & 24.165 & 7.251 & 28.684 & 24.297 \\
Four-trajectory mean & 24.635 & 7.168 & 30.535 & 25.509 \\
\bottomrule
\end{tabular}
\caption{\revcaption{Four-reference tracking RMSE (mm) with original frozen policy and later mixed-trajectory fine-tuned checkpoints.}}
\label{tab:revised_trajectory}
\end{table}

%% file: tables/joint_gate.tex
\begin{table}[H]
\centering
\scriptsize
\setlength{\tabcolsep}{3pt}
\resizebox{\linewidth}{!}{%
\begin{tabular}{@{}lllllll@{}}
\toprule
Controller & Total & J2 & J3 & Peak ($^\circ$) & RMSE (mm) & IAE (m\,s) \\
\midrule
CTC replay & 47 & 10 & 37 & 0.657 & 21.742 & 0.1730 \\
Unguarded DDPG & 102 & 102 & 0 & 1.568 & 12.052 & 0.08519 \\
Joint-2 gated DDPG & 47 & 10 & 37 & 0.666 & 17.497 & 0.13145 \\
\bottomrule
\end{tabular}}
\caption{\revcaption{Historical-replay-matched circular-reference joint-limit counts and error; distinct from the archived 44/101 run.}}
\label{tab:revised_jointgate}
\end{table}

%% file: tables/checkpoint_audit.tex
\begin{table}[H]
\centering
\scriptsize
\setlength{\tabcolsep}{3pt}
\begin{tabular}{@{}lllll@{}}
\toprule
Saved policy & Mean RMSE (mm) & Mean IAE (mm s) & Violation samples & Violating tests \\
\midrule
Agent13\_3cables.mat & 12.019 & 62.798 & 97 & 2 \\
Agent20\_3cables.mat & 12.168 & 65.197 & 1031 & 5 \\
Agent13.mat & 12.739 & 74.844 & 0 & 0 \\
\bottomrule
\end{tabular}
\caption{\revcaption{Retrospective 23-policy comparison (three representative checkpoints shown); five disturbance seeds are evaluation repetitions, not independent policy trainings.}}
\label{tab:revised_retro}
\end{table}

%% file: sections/06c_revision_limitations.tex
\section{Limitations and pathway toward physical validation}\label{sec:revised_limitations}
\begin{revisionblock}
The work is a foundational \emph{numerical} evaluation, not a physical rehabilitation demonstration. The archived controller results do not constitute measured actuator commands, independently verified physical tensions, or a validated patient--robot interaction. The authors confirmed the original numerical force-reconstruction and backprojection path, including its model-specific fallback logic; supplemental bounded-force results were generated in distinct later model configurations and must not be interpreted as experiments on physical actuators. Cross-model checks using Simscape Multibody and MSC ADAMS are numerical consistency diagnostics only. The assumed 0.1-N tension floor, 50--100-N ceilings and first-order lag constants are not calibrated actuator properties, and the mechanisms of cable elasticity, spool inertia, pulley friction, force sensing and motor-drive dynamics remain unmodeled.

The surviving historical training records cover only 20 episodes and cannot support a complete convergence or original checkpoint-selection narrative. New five-seed runs, TD3/SAC comparisons, single-seed bounds/reward experiments, one-factor uncertainty sweeps and multi-reference ablations have different experimental units, settings and selection histories. In particular, training on two waveform cases and evaluating the selected checkpoint on the same cases is not independent validation. Some tests terminate early on joint-limit crossings and some shortened-link references violate a necessary geometric reach bound; these outcomes are failures or feasibility confounders rather than missing-at-random statistical observations. Neither the trial counts nor zero sampled violations justify probabilistic patient-safety statements.

The heuristic joint-2 attenuation leaves joint-3 violations and worsens tracking; current reward terms do not enforce cable feasibility or joint-limit invariance. More comprehensive future reward or safety-filter formulations should account jointly for sampled posture clearance, nonnegative and capacity-bounded cable force, realized rather than merely requested torque, and actuator dynamics, with independent verification of all-joint constraint satisfaction. A prospective differentiable penalty could use $-\lambda_q\sum_i\{[q_{\min,i}+\delta_i-q_i]_+^2+[q_i-(q_{\max,i}-\delta_i)]_+^2\}$, but this is a \emph{proposed} training modification and has not been evaluated in the submitted study.

A staged hardware-validation program would begin with measurement of actual link and payload inertias, attachment/cable geometry, cable stiffness and preload, transmission losses, load-cell characteristics and force-loop bandwidth. The model should then be independently checked under bounded cable actuation and verified emergency-stop and joint-limit protection, followed by real-time or hardware-in-the-loop tests, benchtop robotic experiments, and only subsequently ethically approved human-subject evaluation. No claim of patient safety, clinical rehabilitation efficacy, hardware readiness or formal closed-loop safety follows from the present simulations.
\end{revisionblock}

%% file: sections/07_conclusion.tex
\section{Conclusion}
\begin{revisionblock}
For the studied planar three-joint, three-cable numerical model, a bounded DDPG residual around computed torque control improved Cartesian tracking in the archived representative combined uncertainty/disturbance case: RMS error decreased from 0.0221419 to 0.0128462~m (approximately 42\%). Tracking performance must nonetheless be interpreted jointly with geometric feasibility, joint-limit samples, static cable-demand diagnostics and model-dependent actuation assumptions. An alternate circular reference increased the original residual controller's sampled joint-limit violations relative to CTC (101 versus 44).

Additional post-review simulations show repeatability limits and nonuniform outcomes. Five newly trained DDPG seeds varied in performance; a separate five-seed DDPG/TD3/SAC study did not establish universal DDPG superiority. Exploratory action/reward studies revealed tracking--constraint--allocation trade-offs, including a reward penalty that reduced static allocation error while worsening tracking and joint-limit outcomes. Distinct force-constrained and first-order actuator-response simulations showed sensitivity to assumed force ratings and lag, while independent uncertainty, loading and trajectory-transfer tests produced both successful tracking and early joint-limit terminations. Single-seed mixed-trajectory fine-tuning failed to improve the original saved policy. A heuristic joint-2 gate reduced matched circular-path violation counts but worsened RMSE and did not protect the other joints.

These are simulation outcomes, not experimental demonstrations of pretension, motor capability, patient safety or rehabilitation efficacy. Next steps are independent runtime validation with identified cable and motor dynamics, constrained multi-joint control and reward design, validation across patient-relevant parameters, and staged real-time, hardware-in-the-loop and physical tests.
\end{revisionblock}

%% file: sections/08_declarations_competing.tex
\section*{Declaration of competing interest}

The authors declare that they have no known competing financial interests or personal relationships that could have appeared to influence the work reported in this paper.

%% file: sections/09_declarations_funding.tex
\section*{Funding}

This research did not receive any specific grant from funding agencies in the public, commercial, or not-for-profit sectors.

%% file: sections/10_data_code_availability.tex
\section*{Data availability}
\begin{revisionblock}
The numerical results and descriptive summaries supporting the analyses are reported in the article and Supplementary Information. The original training seed, complete random-source state and contemporaneous checkpoint-selection records were not retained; the surviving 20-episode history and subsequently trained independent seeds are identified separately. Some supplemental runs terminated at joint limits and were not assigned full-horizon metrics. The associated computational-code archive is identified in the Code availability statement. The contents of that software archive should not be construed as recovery of missing original records.
\end{revisionblock}

\section*{Code availability}
\begin{revisionblock}
The computational code associated with this study has been deposited by the authors in Zenodo at \href{https://doi.org/\ZenodoDOI}{\texttt{\ZenodoDOI}}. The published deposit identifies the released software version; MATLAB/Simulink and any third-party dependencies may require separate licenses. The original complete training history, source-random state and contemporaneous checkpoint-selection records were not preserved, and cannot be reconstructed from the deposited code.
\end{revisionblock}

%% file: sections/11_declaration_ai.tex
\section*{Declaration of generative AI and AI-assisted technologies in the manuscript preparation process}

During the preparation of this work, the authors used OpenAI ChatGPT as an assistive tool. The authors reviewed and revised the AI-assisted content and take full responsibility for the manuscript. No AI-generated simulation results or numerical data are reported.